\documentclass[10pt,journal,compsoc]{IEEEtran}
\usepackage[table,xcdraw]{xcolor}
\usepackage{color}
\usepackage{amsmath,amsfonts}
\usepackage{algorithmic}
\usepackage{array}
\usepackage{textcomp}
\usepackage{stfloats}
\usepackage{url}
\usepackage{verbatim}
\usepackage{graphicx}
\usepackage{multirow}
\usepackage{amssymb}
\usepackage{bm}
\usepackage{tikz}
\usepackage{comment}
\usepackage{mathrsfs}
\usepackage{indentfirst}
\usepackage[ruled,linesnumbered]{algorithm2e}
\usepackage{grffile}
\usepackage{float} 
\usepackage{hyperref}
\hypersetup{hidelinks,
	colorlinks=true,
	allcolors=black,
	pdfstartview=Fit,
	breaklinks=true}

\usepackage{subfig}
\usepackage{longtable}
\usepackage{threeparttable}
\usepackage{booktabs}
\usepackage{caption}
\hypersetup{
        colorlinks=true,
        linkcolor=blue,
        anchorcolor=blue,
        citecolor=blue}           
\usepackage{cite}
\usepackage{enumitem}
\usepackage{bigstrut}
\usepackage[table]{xcolor}

\usepackage[capitalize]{cleveref}
\crefname{section}{Sec.}{Secs.}
\Crefname{section}{Section}{Sections}
\Crefname{table}{Table}{Tables}

\ifCLASSINFOpdf

\else

\fi

\usepackage{amsmath}

\makeatletter
\newcommand*\smallsum{\mathpalette\smallsum@{.6}}
\newcommand*\smallsum@[2]{\mathop{\vcenter{\hbox{\scalebox{#2}{$\m@th#1\sum$}}}}}
\makeatother

\makeatletter
\newcommand*\smallprod{\mathpalette\smallprod@{.8}}
\newcommand*\smallprod@[2]{\mathop{\vcenter{\hbox{\scalebox{#2}{$\m@th#1\prod$}}}}}
\makeatother

\begin{document}
\title{Diffusion Models in Low-Level Vision: A Survey}
\author{Chunming~He,~\IEEEmembership{}
        Yuqi~Shen,~\IEEEmembership{}
        Chengyu~Fang,~\IEEEmembership{}
        Fengyang~Xiao,~\IEEEmembership{}
      Longxiang~Tang,~\IEEEmembership{} \\
Yulun~Zhang, ~%\IEEEmembership{Member,~IEEE,}
        Wangmeng~Zuo,~\IEEEmembership{Senior Member,~IEEE,}
        Zhenhua~Guo,~\IEEEmembership{}
        Xiu~Li~%\IEEEmembership{Member,~IEEE,}% <-this % stops a space
\thanks{This work was supported by the STI 2030-Major Projects under Grant 2021ZD0201404.}

\thanks{\textit{Corresponding author: Xiu Li (e-mail: li.xiu@sz.tsinghua.edu.cn).}}

\thanks{\textit{Chunming He, Yuqi Shen, and Chenyu Fang contributed equally. }
}
\thanks{Chunming He, Yuqi Shen, Chengyu Fang, Longxiang Tang, and Xiu Li are with Tsinghua Shenzhen International Graduate School, Tsinghua University, Shenzhen 518055, China (e-mail:chunminghe19990224@gmail.com; ericsyq\_buaa@163.com; chengyufang.thu@gmail.com; lloong.x@gmail.com).} 
\thanks{Chunming He and Fengyang Xiao are with the Department of Biomedical Engineering, Duke University, Durham, NC 27708 USA (e-mail: chunming.he@duke.edu; xiaofy5@mail2.sysu.edu.cn).}
\thanks{Yulun Zhang is with MoE Key Lab of Artificial Intelligence, AI Institute, Shanghai Jiao Tong University, Shanghai, China (e-mail: yulun100@gmail.com).}
\thanks{Wangmeng Zuo is with the School of Computer Science and Technology,
Harbin Institute of Technology, Harbin 150006, China (e-mail: wmzuo@hit.edu.cn).}
\thanks{Zhenhua Guo is with Tianyijiaotong Technology Ltd., Suzhou 215131, China (e-mail: cszguo@gmail.com).}}

\markboth{IEEE TPAMI}%
{He \MakeLowercase{\textit{et al.}}: Diffusion Models in Low-Level Vision Tasks: A Survey}

\IEEEtitleabstractindextext{%
\begin{abstract}
Deep generative models have gained considerable attention in low-level vision tasks due to their powerful generative capabilities. Among these, diffusion model-based approaches, which employ a forward diffusion process to degrade an image and a reverse denoising process for image generation, have become particularly prominent for producing high-quality, diverse samples with intricate texture details. Despite their widespread success in low-level vision, there remains a lack of a comprehensive, insightful survey that synthesizes and organizes the advances in diffusion model-based techniques. 
To address this gap, this paper presents the first comprehensive review focused on denoising diffusion models applied to low-level vision tasks, covering both theoretical and practical contributions. We outline three general diffusion modeling frameworks and explore their connections with other popular deep generative models, establishing a solid theoretical foundation for subsequent analysis. We then categorize diffusion models used in low-level vision tasks from multiple perspectives, considering both the underlying framework and the target application. Beyond natural image processing, we also summarize diffusion models applied to other low-level vision domains, including medical imaging, remote sensing, and video processing.
Additionally, we provide an overview of widely used benchmarks and evaluation metrics in low-level vision tasks. Our review includes an extensive evaluation of diffusion model-based techniques across six representative tasks, with both quantitative and qualitative analysis. Finally, we highlight the limitations of current diffusion models and propose four promising directions for future research.
This comprehensive review aims to foster a deeper understanding of the role of denoising diffusion models in low-level vision. For those interested, a curated list of diffusion model-based techniques, datasets, and related information across over 20 low-level vision tasks is available at \url{https://github.com/ChunmingHe/awesome-diffusion-models-in-low-level-vision}.

\end{abstract}

}

\maketitle

\IEEEdisplaynontitleabstractindextext

\IEEEpeerreviewmaketitle

%% narrow the gap between equations and sentences
\setlength{\abovedisplayskip}{2pt}
\setlength{\belowdisplayskip}{2pt}

\begin{figure*}[ht]
    \subfloat[Image Super-resolution]{
    {\includegraphics[height=78pt,width=0.15\textwidth]{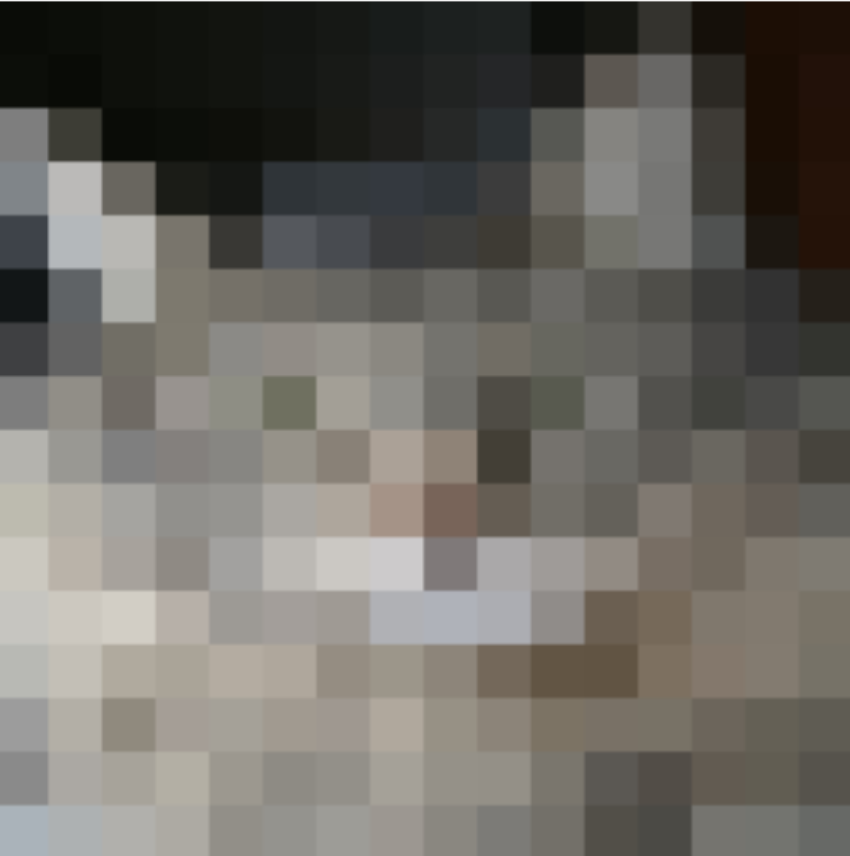}}
    {\includegraphics[height=78pt,width=0.15\textwidth]{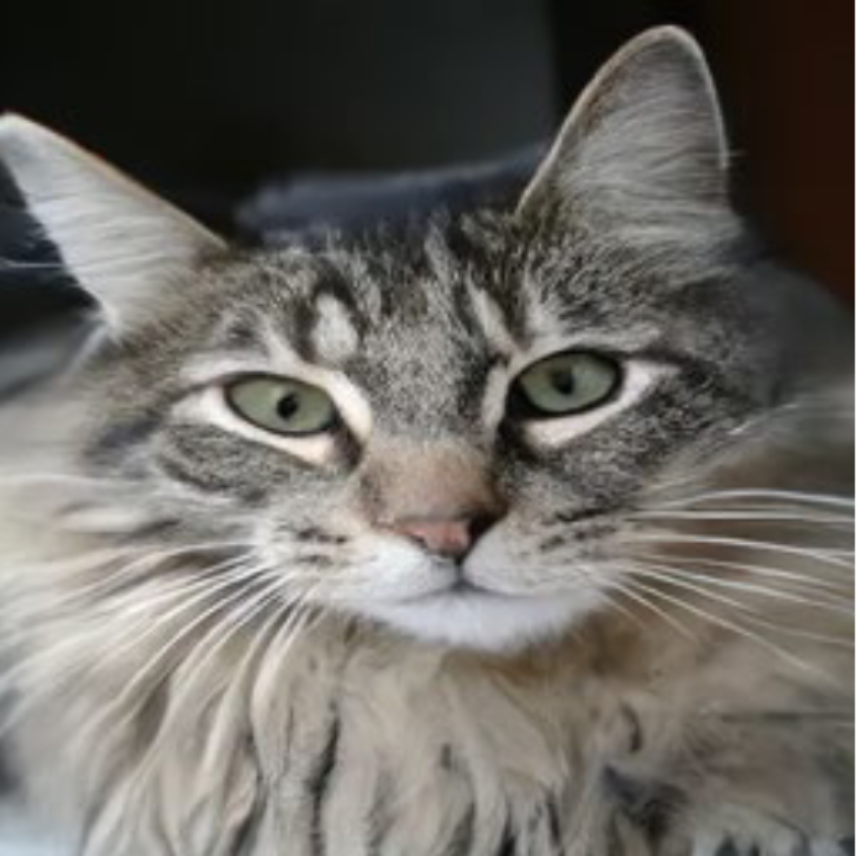}}}
 \hfill 	
  \subfloat[Image Deblurring]{
  {\includegraphics[height=78pt,width=0.15\textwidth]{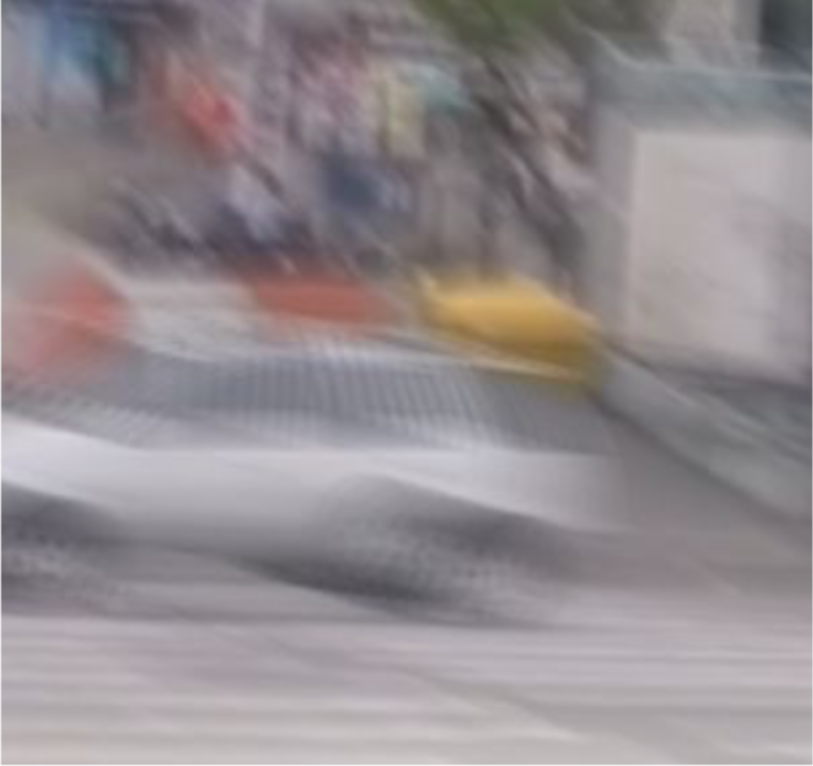}}
  {\includegraphics[height=78pt,width=0.15\textwidth]{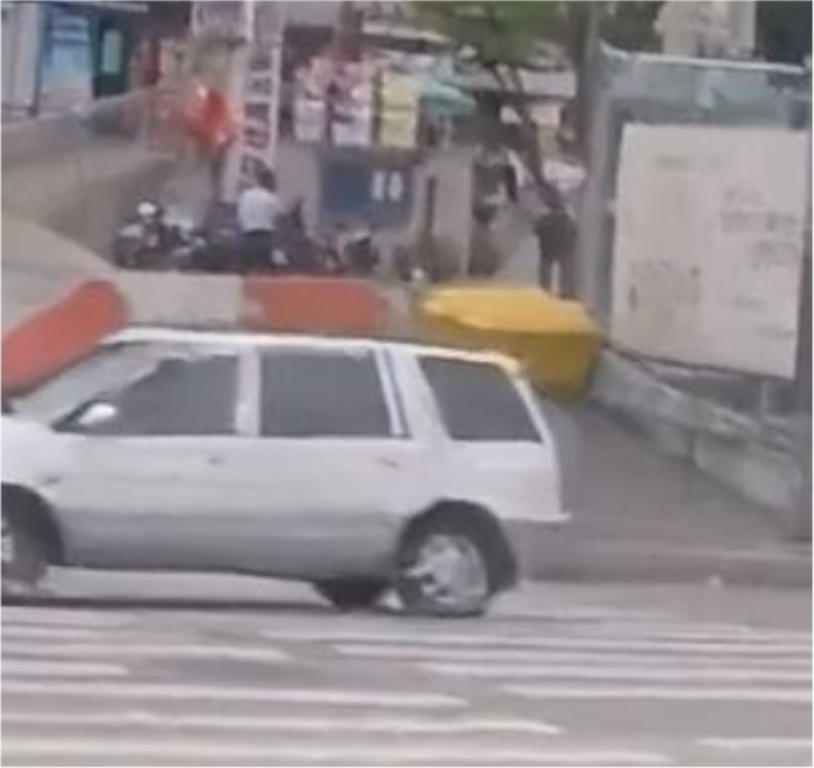}}}
 \hfill	
 %  \subfloat[Image Deraining]{{\includegraphics[height=78pt,width=0.15\textwidth]{images/examples/WeatherDiffusion-input.png}}
 %  {\includegraphics[height=78pt,width=0.15\textwidth]{images/examples/WeatherDiffusion-output.png}}}
 %  % \newline
 %  \subfloat[Image Dehazing]{
 %  {\includegraphics[height=78pt,width=0.15\textwidth]
 %  {images/examples/Refusion-dehaze-input.png}}
 %  {\includegraphics[height=78pt,width=0.15\textwidth]
 %  {images/examples/Refusion-dehaze-output.png}}}
 % \hfill 	
  \subfloat[Image Inpainting]{  {\includegraphics[height=78pt,width=0.15\textwidth]
  {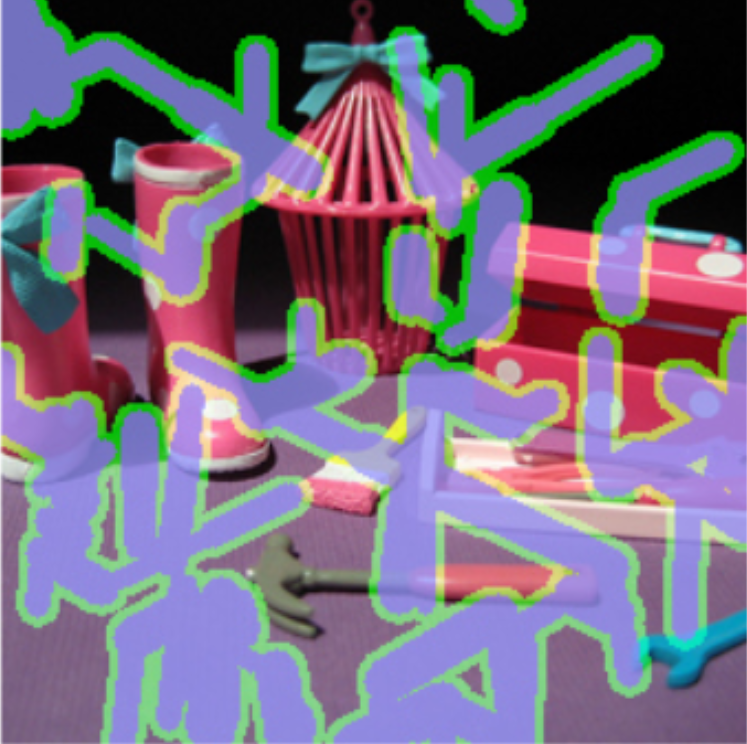}}
  {\includegraphics[height=78pt,width=0.15\textwidth]
  {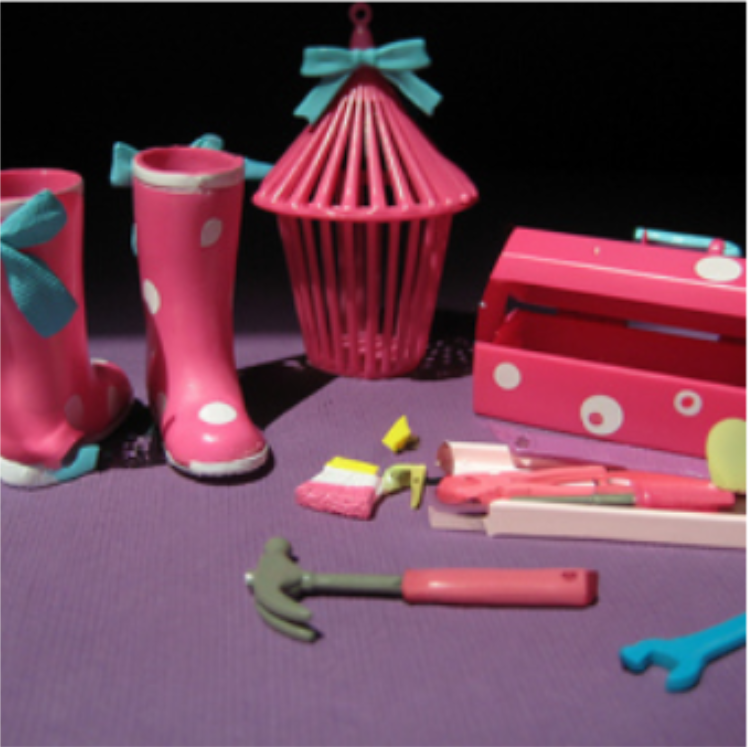}}}
 \\ \vspace{-2mm}
  \subfloat[Low-light Image Enhancement]{  {\includegraphics[height=78pt,width=0.15\textwidth]
  {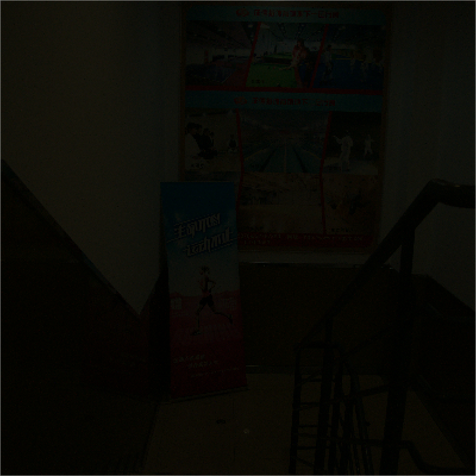}}
  {\includegraphics[height=78pt,width=0.15\textwidth]
  {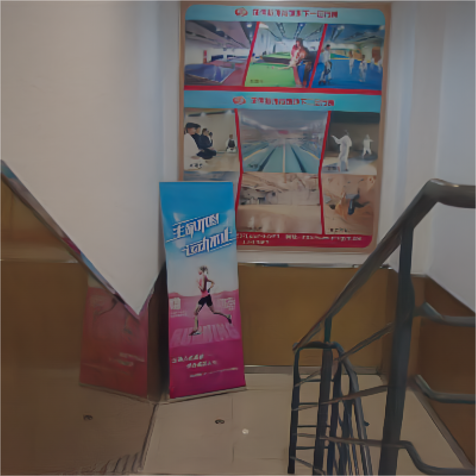}}}
   \hfill
  \subfloat[Limited-angle CT Reconstruction]{
  {\includegraphics[height=78pt,width=0.15\textwidth]
  {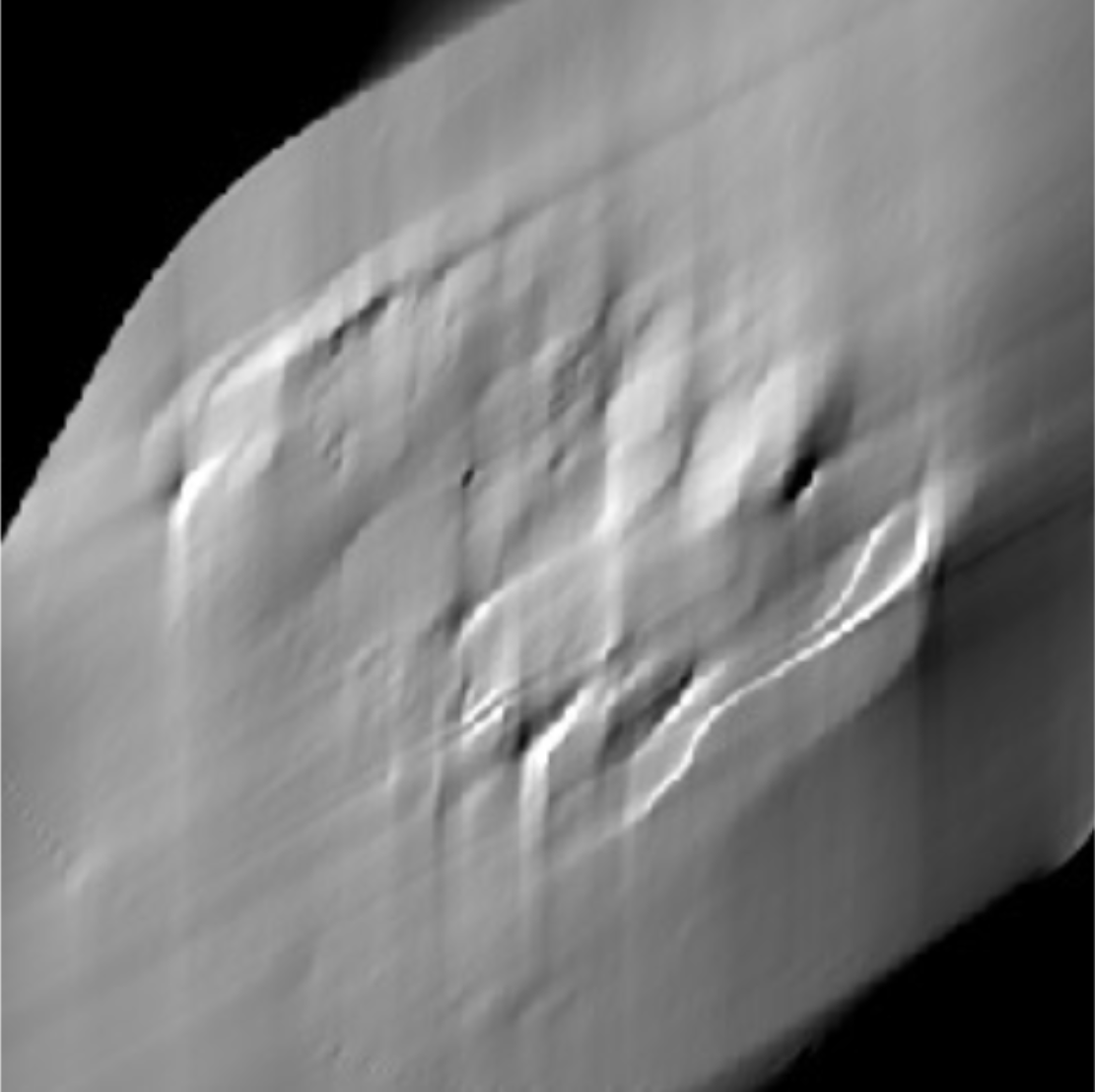}}
  {\includegraphics[height=78pt,width=0.15\textwidth]
  {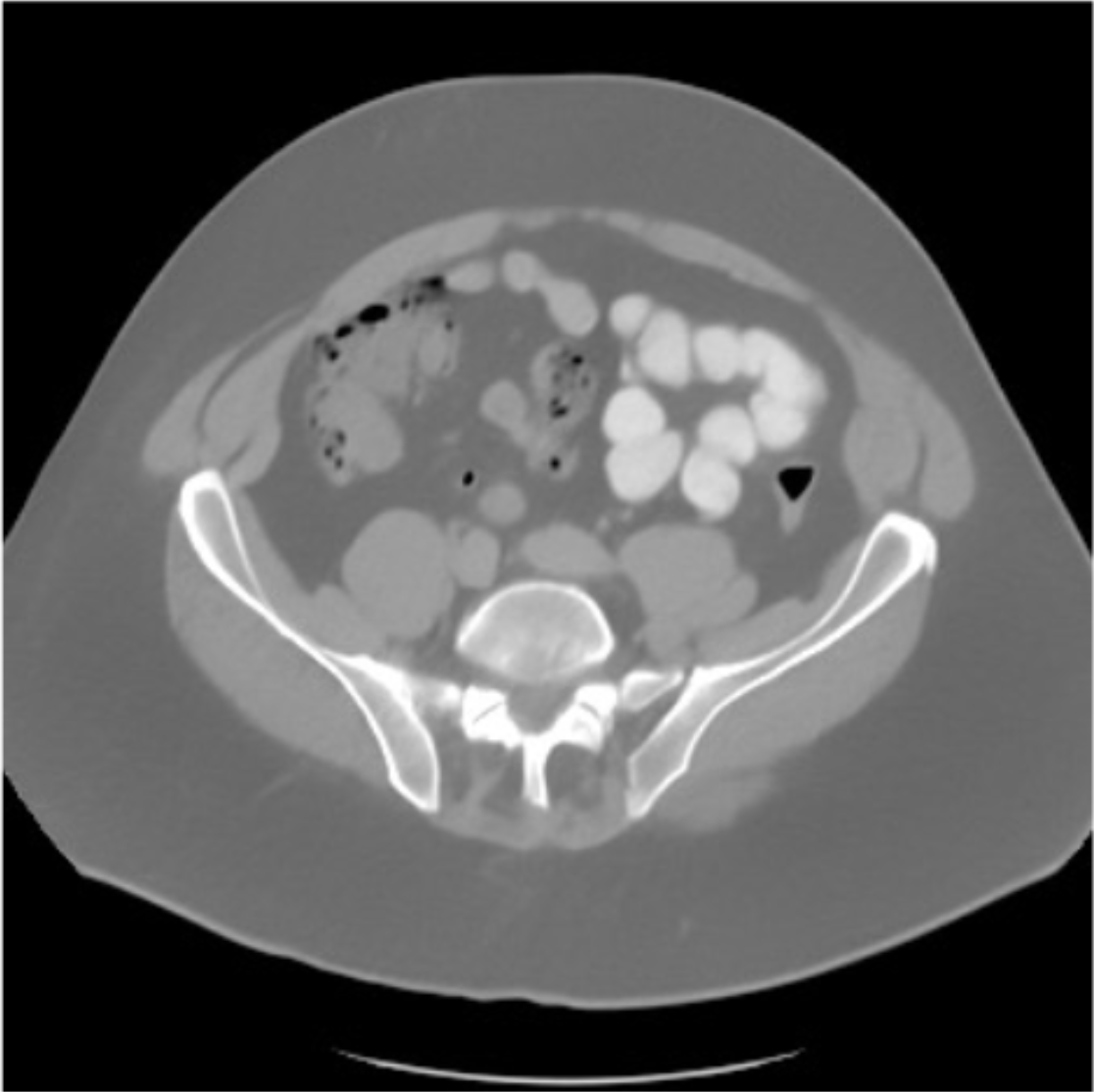}}}
 % \hfill 	
 %  \subfloat[Accelerated MRI Reconstruction]{  {\includegraphics[height=78pt,width=0.15\textwidth]
 %  {images/examples/ScoreMRI_input.png}}
 %  {\includegraphics[height=78pt,width=0.15\textwidth]
 %  {images/examples/ScoreMRI_output.png}}}
 \hfill	
  \subfloat[Cloud Removal]{  {\includegraphics[height=78pt,width=0.15\textwidth]
  {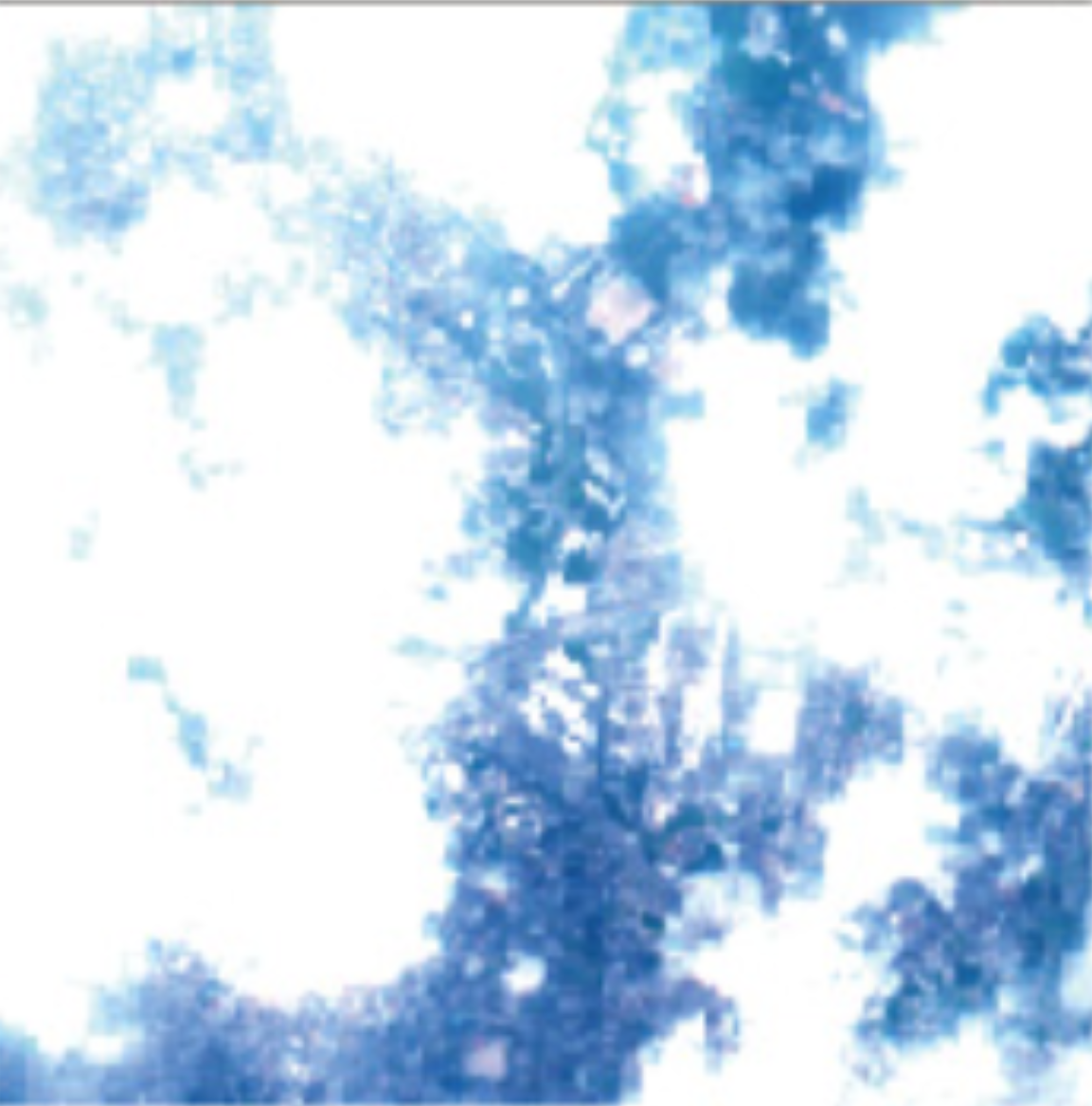}}
  {\includegraphics[height=78pt,width=0.15\textwidth]
  {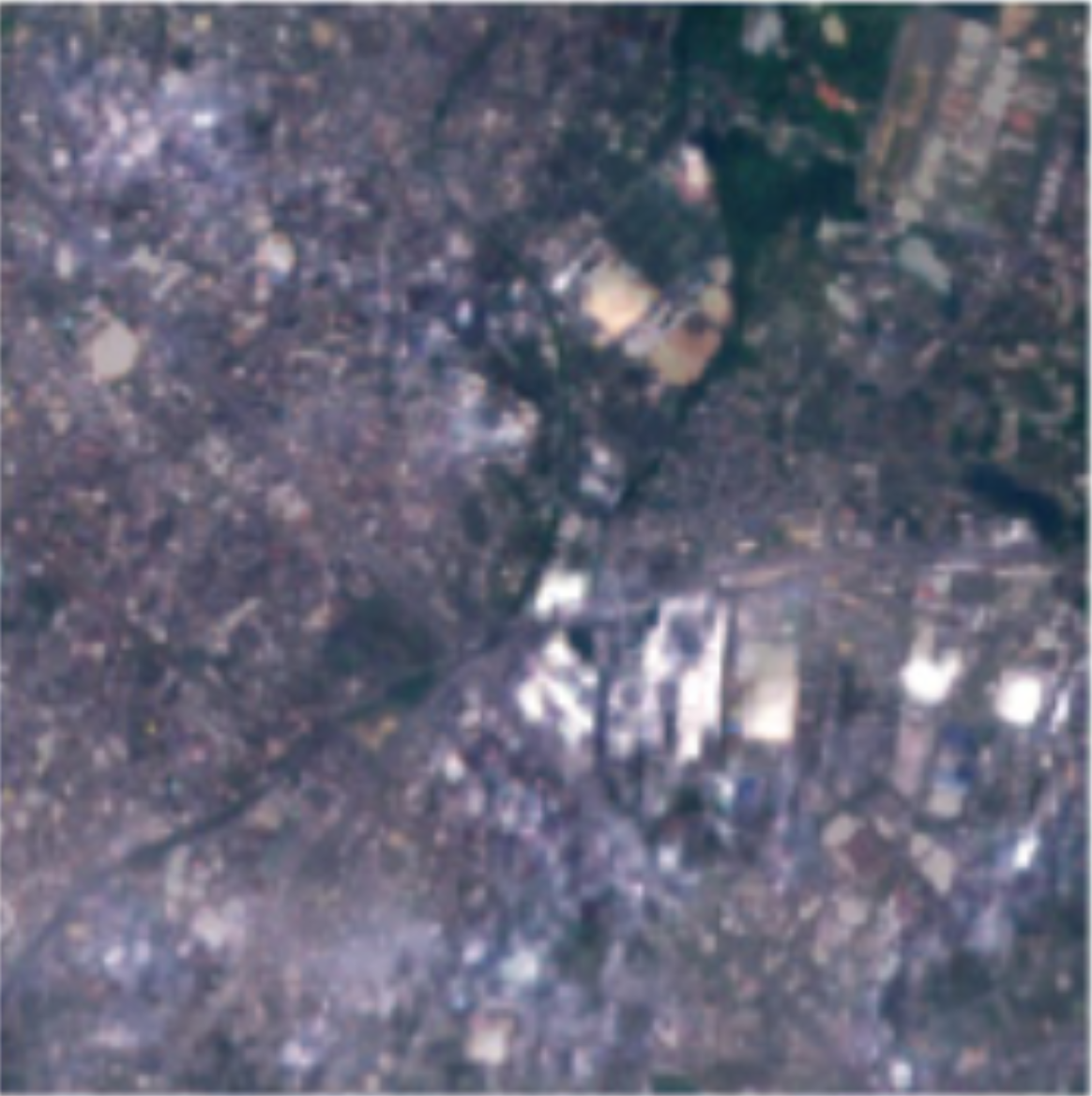}}}
  % \vspace{-0.5mm}
\caption{Examples of various low-level vision tasks with the low-quality image (left) and the enhanced high-quality image (right).
% , where the left and right images are the original low-quality and their enhanced high-quality image. 
Notice that all the enhanced results are generated with diffusion model-based algorithms, which are IDM~\cite{sp-idm} in (a), MSGD~\cite{ren2023multiscale} in (b), 
% WeatherDiffusion~\cite{sp-WeatherDiff} in (c), Refusion~\cite{sp-luo2023refusion} in (d), 
Repaint~\cite{sp-repaint} in (c), Reti-Diff~\cite{he2023reti-LLIE3} in (d), DOLCE~\cite{sp-LACT-liu2023dolce} in (e), 
% ScoreMRI~\cite{chung2022scoreMRI} in (h), 
and DDPM-CR~\cite{p-RS-cloudremoval-jing2023denoising} in (f).} \vspace{-4mm}
\label{fig:examples}
\end{figure*}

\ifCLASSOPTIONcompsoc
\IEEEraisesectionheading{\section{Introduction}\label{sec:introduction}}
\else
\section{Introduction}
\label{sec:introduction}
\fi
\IEEEPARstart{L}{ow-level} vision tasks, a fundamental aspect of computer vision, have been extensively studied for improving low-quality data degraded by complex scenarios. These tasks encompass a wide range of practical applications, including but not limited to image super-resolution~\cite{R-SR}, deblurring~\cite{R-deblur}, dehazing~\cite{R-dehaze}, inpainting~\cite{sp-diffir}, fusion~\cite{he2023degradation}, compressed sensing~\cite{sp-VDMprediction-Yang}, low-light enhancement~\cite{p-llie-wang2023lldiffusion}, and cloud removal in remote sensing~\cite{p-RS-cloudremoval-jing2023denoising}. See~\cref{fig:examples} for visual results.

% \cite{R-deblur, R-dehaze, R-denoiseDL, R-derain, R-IRDL, R-IR-DL-RS, R-SR} have been extensively explored in the field of computer vision, playing a pivotal role in enhancing image quality. These tasks encompass a diverse range of objectives, including image super-resolution, deblurring, denoising, inpainting, and compression artifact removal. 
\begin{figure*}[ht]
 {\includegraphics[width=\textwidth]{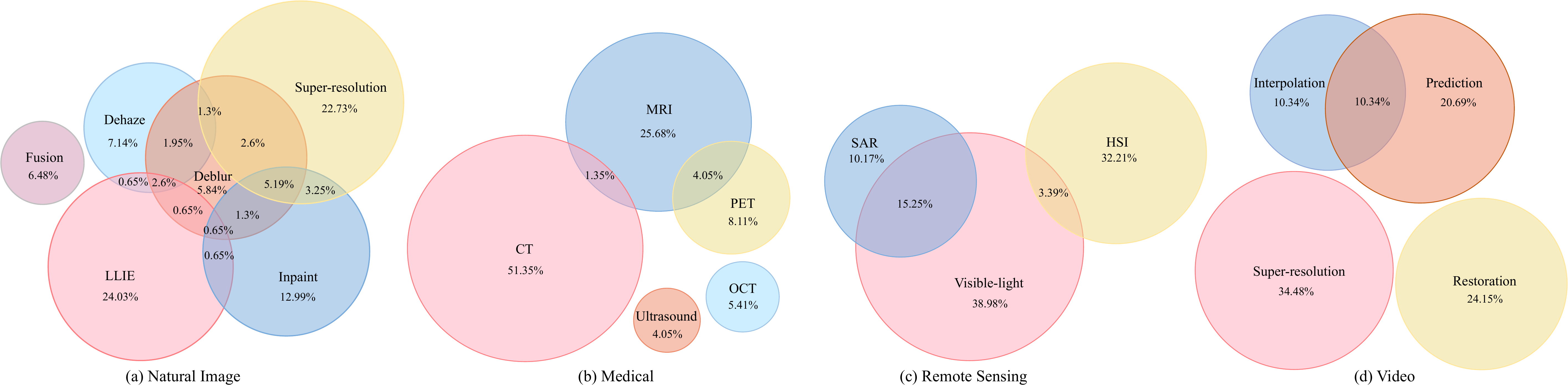}}
\caption{{Distributions of the four main low-level vision scenarios of DM-based models. In each Venn diagram, the overlapping regions between circles indicate that these models can address multiple application tasks or input modalities. }} \vspace{-4mm}
\label{fig:venn}
\end{figure*}

Traditional approaches \cite{R-denoisetraditional, he2019image} framed low-level vision problems as variational optimization challenges and utilized handcrafted algorithms to enforce proximity constraints related to specific image properties or degradation priors~\cite{he2023camouflaged,he2023weaklysupervised,xiao2023concealed,xu2022multi}. However, these methods often struggle to handle complex degradations due to their limited generalizability.
With the rise of deep learning, convolutional neural networks (CNNs) \cite{R-IRDL} and transformers \cite{R-IR-VIT} have become widely adopted in low-level vision tasks for their powerful feature extraction capabilities. Additionally, the availability of large-scale datasets, such as \textit{DIV2K} \cite{D-DIV2K} for super-resolution and \textit{Rain800} \cite{d-gopro} for deraining, has further enhanced their generalizability. While these methods have achieved promising results, particularly in distortion-based metrics like PSNR and SSIM, they still suffer from poor texture generation, limiting their applicability in complex real-world scenarios.

To address this limitation, deep generative models, particularly generative adversarial networks (GANs)~\cite{he2023hqg}, have been introduced into low-level vision tasks. Leveraging their strong generative abilities, these models aim to synthesize realistic texture details, extending their applicability to real-world scenarios. However, GAN-based methods face critical challenges: (1) the training process is prone to mode collapse and unstable optimization, requiring intricate hyperparameter tuning, and (2) the generated results often exhibit artifacts and counterfactual details, thereby undermining global coherence and limiting practical use.

Recently, diffusion models (DMs)~\cite{T-1, T-NCSN, T-DDIM, T-DDPM, T-LDM, T-nichol2021improved, T-SDE, T-watson2021learning, R-beatsGAN} have emerged as a promising alternative in computer vision due to their impressive generative capabilities and training stability. DMs operate through a forward diffusion process, which introduces noise to the data, and a reverse diffusion process that learns to remove the noise, thus generating high-quality samples. Unlike GANs, DMs fall under the category of likelihood-based models and frame their training objective as a re-weighted variational lower bound. This offers benefits such as extensive distribution coverage, a stable training objective, and straightforward scalability.

Building on these advantages, DMs have shown remarkable success across various domains, including data generation, image content comprehension, and low-level vision. In the realm of low-level vision, DMs~\cite{sp-idm,ren2023multiscale,sp-WeatherDiff,sp-luo2023refusion} primarily focus on restoring degraded data, thus enabling the reconstruction of high-quality images with detailed semantics and realistic textures, even in scenarios characterized by severe and complex degradations. As depicted in~\cref{fig:examples}, numerous DM-based algorithms have delivered promising results across diverse low-level vision tasks. However, the diversity and complexity of techniques used in different tasks pose significant challenges for understanding, improving, and developing a general-purpose reconstruction model. Therefore, there is a critical need for a well-organized and comprehensive survey on DM-based low-level vision tasks. Existing DM-based surveys~\cite{R-DM1,R-DM2,R-T2Izhang2023text,R-huang2024diffusionimageediting} generally focus on foundational theoretical models or generation-based techniques, while only a few reviews~\cite{R-DM-inpainting-parida2023survey,R-DM-SR-moser2024diffusion,R-DMIR} address specific problems or a limited range of tasks in natural image scenarios within low-level vision.

To address this gap and overcome the aforementioned limitations, we propose the first comprehensive DM-based survey tailored to low-level vision tasks (see~\cref{fig:venn,fig:MainImage}). 
This survey provides a detailed theoretical introduction, explores wide-ranging applications, offers thorough experimental analyses, and presents extensive future perspectives. Specifically, we begin with a comprehensive overview of diffusion models in \cref{chap:theorybase}, clarifying their connections to other deep generative models. We then summarize cutting-edge DM-based methods in natural low-level vision tasks in \cref{chap:DM-naturalIR}, categorizing them based on both their underlying frameworks and target tasks, covering six widely used tasks. 
{In \cref{chap:DM-extendIR}, we expand the scope to include medical imaging, remote sensing, and video scenarios, providing a broad overview of DM applications. Furthermore, \cref{chap:experiment} reviews widely used benchmarks and fundamental evaluation metrics in low-level vision tasks, and presents a comprehensive experimental evaluation of DM-based techniques across six representative tasks, both quantitatively and qualitatively.} Finally, in \cref{chap:futurework}, we identify key limitations of current DM-based methods and propose four major directions for future research, followed by a concluding summary in \cref{chap:conclusion}.

\begin{figure*}[ht]
    {\includegraphics[width=\textwidth]{images/Main\_Image/Main\_time\_new4.pdf}}
\caption{{The bar chart illustrates the continuous growth of DM-based methods in low-level vision tasks across four distinct scenarios. Representative works are categorized and marked on the line graph with colors corresponding to each scenario as indicated in the legend. The methods highlighted represent the seminal works of each period, \textit{e.g.}, StableSR~\cite{p-crossmodal-accelerate3-wang2023exploiting} has garnered 1.9k GitHub stars, SR3~\cite{sp-SR3} boasts 1.2k citations, and SUPIR~\cite{yu2024scalingsupir} is a pioneering DM-based multi-modal solution.}} \vspace{-4mm}
\label{fig:MainImage}
\end{figure*}

\noindent \textbf{Note}. 
We explored multiple databases, including DBLP, Google Scholar, and ArXiv, and focused on reputable sources such as TPAMI, IJCV, and CVPR. Preference was given to studies with available code and higher citations, reflecting broader academic recognition. We further applied a rigorous evaluation process to each paper, assessing its contribution and determining whether it was a seminal work. Hence, our survey can present a comprehensive overview of the most influential research, thus advancing the field and highlighting promising future directions.

\begin{figure}[!t]
    \centering
    \setlength{\abovecaptionskip}{0.1cm}
    {\includegraphics[width=.49\textwidth]{images/DMtheory/DMtheory\_final.pdf}}
\caption{The schematic diagram of diffusion models.}\vspace{-3.5mm} 
\label{fig:diffusion process}
\end{figure}
\section{A Walk-through of diffusion models}\label{chap:theorybase}

Diffusion models constitute a category of likelihood-based models. They are characterized by a shared principle of progressively perturbing data through a random noise process known as "diffusion" and then removing the noise to produce samples (see \cref{fig:diffusion process}). These models are typically classified into three subcategories: denoising diffusion probabilistic models (DDPMs), noise-conditional score networks (NCSNs), and stochastic differential equations (SDEs).

DDPMs and their variants have garnered significant attention owing to their straightforward algorithmic flow and the ease of integrating conditional controls.  In contrast, NCSNs and SDEs are often subject to detailed mathematical analysis, given their potential for more efficient sampling and enhancements in task generalization.
\vspace{-1mm}

\subsection{Denoising Diffusion Probabilistic Models}
A vanilla DDPM employs two Markov chains: a forward chain that perturbs data into random noise, and a reverse chain that converts the noise back to data. The initial diffusion process transforms data ${{x}_{0}} \sim q({{x}_{0}})$ from a complex distribution into a latent variable ${{x}_{T}}$ in a fixed simple prior distribution (\textit{e.g.}, standard Gaussian) over $T$ timesteps. At each diffusion step, Gaussian noise $\varepsilon$ is added to the data, following a hand-designed variance schedule $\{{{\beta }_{1}},\ldots,{{\beta }_{T}}\}$, and ${{x}_{t}} \in {{\mathbb{R}}^{d}}$, $t\in \{1,2,\ldots,T\}$, 
% represents the outcome of each timestep, 
sharing the same dimension $d$ as ${{x}_{0}}$. Hence, the forward process can be expressed as the posterior $q({{x}_{1}},\ldots,{{x}_{T}}|{{x}_{0}})$ based on the Markov chains:
\begin{equation}
\setlength{\abovedisplayskip}{0pt}
\setlength{\belowdisplayskip}{0pt}
    q({{x}_{1}},\cdots ,{{x}_{T}}|{{x}_{0}}):=\smallprod_{t=1}^{T}{q({{x}_{t}}|{{x}_{t-1}})},\label{eq:1}
\end{equation}
\begin{equation}
\setlength{\abovedisplayskip}{0pt}
\setlength{\belowdisplayskip}{0pt}
    q({{x}_{t}}|{{x}_{t-1}}):=\mathcal{N}({{x}_{t}};\sqrt{1-{{\beta }_{t}}}{{x}_{t-1}},{{\beta }_{t}}\bf I ),\label{eq:2}
\end{equation}
given the hyperparameters ${\alpha}_{t}:=1-{{\beta }_{t}}$, $~{{\bar{\alpha }}_{t}}:=\smallprod_{s=1}^{t}{{{\alpha }_{s}}}\text{ }$. The above equations can be reformulated as
\begin{equation}
    q({{x}_{t}}|{{x}_{0}})=\mathcal{N}({{x}_{t}};\sqrt{{{{\bar{\alpha }}}_{t}}}{{x}_{0}},(1-{{\bar{\alpha }}_{t}})\bf I ).\label{eq:3}
\end{equation}

% which can be further reparameterized as
By reparameterizing \cref{eq:3}, ${x}_{t}$ can be calculated as
\begin{equation}
    {{x}_{t}}({{x}_{0}},\epsilon )=\sqrt{{{{\bar{\alpha }}}_{t}}}{{x}_{0}}+\sqrt{1-{{{\bar{\alpha }}}_{t}}}\epsilon ,\epsilon \sim\mathcal{N}( 0,\bf I).\label{eq:4}
\end{equation}

While the latter process ${{p}_{\theta }}({{x}_{0}})=\int{{{p}_{\theta }}({{x}_{0:T}})d{{x}_{1:T}}}$ reverses the former from $p({{x}_{T}})=\mathcal{N}({{x}_{T}};0,\bf I )$.
\begin{equation}
    {{p}_{\theta }}({{x}_{t-1}}|{{x}_{t}})=\mathcal{N}({{x}_{t-1}};{{\mu }_{\theta }}({{x}_{t}},t),{{\smallsum }_{\theta }}({{x}_{t}},t)),\label{eq:5}
\end{equation}
where learnable Gaussian transitions kernels with $\theta$ are parameterized by deep neural networks under the training objects of minimizing the Kullback-Leibler (KL) divergence between $q({x_{0}},{{x}_{1}},\cdot \cdot \cdot ,{{x}_{T}})$ and ${{p}_{\theta }}({{x}_{0}},{{x}_{1}},\cdot \cdot \cdot ,{{x}_{T}})$.

The optimization principle is as follows: To generate \( x_0 \) in the reverse process, we sample from the noise vector $x_T \sim p(x_T)$ to obtain \( x_{T-1}, x_{T-2}, \ldots, x_1, x_0 \) using the learnable transition kernel. The key to this sampling process is training the reverse Markov chain to match the actual time reversal of the forward Markov chain. This requires adjusting \( \theta \) to align the joint distribution of the reverse Markov chain \( p_{\theta}(x_0, x_1, \ldots, x_T) \) closely with that of the forward process \( q(x_0, x_1, \ldots, x_T) \). We use the KL divergence to characterize the gap between these two distributions. \( \theta \) can be trained by minimizing the KL divergence:

\begin{equation}
\setlength{\abovedisplayskip}{-2pt}
\setlength{\belowdisplayskip}{2pt}
\begin{split}
&KL(q({{x}_{0}},{{x}_{1}},\cdot \cdot \cdot ,{{x}_{T}})||{{p}_{\theta }}({{x}_{0}},{{x}_{1}},\cdot \cdot \cdot ,{{x}_{T}}))\\&\overset{(i)}{\mathop{=}}\,-{{\mathbb{E}}_{q({{x}_{0}},{{x}_{1}},\cdot \cdot \cdot ,{{x}_{T}})}}[\log {{p}_{\theta }}({{x}_{0}},{{x}_{1}},\cdot \cdot \cdot ,{{x}_{T}})]+const\\&\overset{(ii)}{\mathop{=}}\,-{{\mathbb{E}}_{q({{x}_{0}},{{x}_{1}},\cdot \cdot \cdot ,{{x}_{T}})}}[-\log p({{x}_{T}})-\sum\limits_{t=1}^{T}{\frac{{{p}_{\theta }}({{x}_{t-1}}|{{x}_{t}})}{q({{x}_{t}}|{{x}_{t-1}})}}]\\&\ge \mathbb{E}[-\log {{p}_{\theta }}({{x}_{0}})]+const.
\end{split}
\label{eq:6}
\end{equation}

For better sample quality, a simplified form of loss function is proposed as the optimization target of the model\cite{ho2020denoising}:
\begin{equation}
    {{\mathbb{E}}_{t\sim \mathcal{U}\left[\!\left[ 1,T \right]\!\right],{{x}_{0}}\sim q({{x}_{0}}),\epsilon \sim \mathcal{N}(0,\bf I )}}\left[ \lambda (t){{\left\| \epsilon -{{\epsilon }_{\theta }}({{x}_{t}},t) \right\|}^{2}} \right],\label{eq:10}
\end{equation}
where $\lambda(t)$ is a positive weighting function. $\mathcal{U}\left[\!\left[ 1,T \right]\!\right]$ is a uniform distribution over the set $\{1, 2, \ldots, T \}$. ${{\epsilon }_{\theta }}$ is a deep network with parameters $\theta$ that predicts the noise vector $\epsilon$.

\subsection{Noise Conditioned Score Networks}\label{Sec:NCSN}

NCSNs are designed to estimate the probabilistic distribution of the target data from the score function, which guides the sampling process progressively toward the center of the data distribution. The score function for a specific data density $p(x)$ is defined as the gradient of the log-density function, ${{\nabla }{x}}\log p\left( x \right)$, which defines a vector field over the entire space that data $x$ inhabits, pointing towards the directions along which the probability density function has the largest growth rate. The Langevin dynamics algorithm uses the directions provided by these gradients \cite{T-NCSN} to iteratively shift from a random prior sample ${{x}_{0}}$ to samples ${{x}_{T}}$ in regions with high density. By learning the score function of a real data distribution, it can generate samples from any point in the same space by iteratively following the score function until a peak is reached, which is defined as 
\begin{equation}
{{x}_{t}}={{x}_{t-1}}+\frac{\gamma }{2}{{\nabla }_{x}}\log p(x)+\sqrt{\gamma }{{\epsilon }_{t}},
    \label{eq:NCSN1}
\end{equation}
where $t\sim\mathcal{U}\left[\!\left[ 1,T \right]\!\right]$. $\gamma$ controls the updating magnitude in the direction of the score, akin to the learning rate in stochastic gradient descent. The noise ${{\epsilon }_{t}}\sim \mathcal{N}\left( 0,\mathbf{I}  \right)$ represents random normal Gaussian noise at time step $t$, introducing random perturbations into the recursive process to address the issue of getting stuck in local minima. As the time step $T\to \infty$ and $\gamma \to 0$, the distribution $p\left( {{x}_{T}} \right)$ approaches the original data distribution $p(x)$. Hence, a generative model can utilize the above method to sample from $p(x)$ after estimating the score with a network ${{s}_{\theta }}\left( x,\text{ }t \right)\approx {{\nabla }_{x}}\log\text{ }{p}\left( x \right)$. This network can be trained via score matching \cite{T-scorematching1} to optimize the objective function presented as follows:
\begin{equation}
% \begin{split}
    \underset{\theta}{\mathop{\min}}\ {\mathbb{E}}_{t,{{x}_{0}},{x}_{t}}[ \lambda(t)\| {{s}_{\theta }}({x}_{t},t)-{{\nabla }_{{{x}_{t}}}}\log p({x}_{t}|{x}_{0}) \Vert_{2}^{2}],
% \end{split}
\label{eq:NCSN2}
\end{equation}
where $t\sim\mathcal{U}\left[\!\left[ 1,T \right]\!\right],{{x}_{0}}\sim p({{x}_{0}}),{x}_{t}\sim{{p}}({x}_{t}|{x}_{0})$.
In practice, because $\nabla_{x_{t}} \log p (x_t | x_0)$ is unknown, \cref{eq:NCSN2} can only be solved by those score matching-based methods rather than be directly solved, limiting the generalization to real data. According to the manifold hypothesis, conventional score function estimation methods, including denoising score matching \cite{T-scorematching1} and sliced score matching \cite{T-scorematching2}, when combined with Langevin dynamics, can lead the resulting distribution to collapse to a low-dimensional manifold and thus bring inaccurate score estimation in the low-density region. To address this issue, annealed Langevin dynamics perturbs the data with Gaussian noise at different scales and further proposes an optimization objective under a monotonically decreasing noise strategy $({{\sigma }_{t}})_{t=1}^{T}$:
\begin{equation}
% \begin{split}
 \mathcal{L}\left( \theta ,{{\sigma }_{t}} \right)\!=\!\frac{1}{T}\sum\limits_{t=1}^{T}{\lambda (}{{\sigma }_{t}}){{\mathbb{E}}_{p(x),{{x}_{t}}}} [\|{{s}_{\theta }}({{x}_{t}},{{\sigma }_{t}})\!+\!\frac{{{x}_{t}}-x}{\sigma _{t}^{2}}\|_{2}^{2}],\label{eq:NCSN3}
% \end{split}
\end{equation}   
where ${{{{x}_{t}}\sim {{p}_{{{\sigma }_{t}}}}({{x}_{t}}\left| x) \right.}}$. In inference, one can initiate with white noise and apply \cref{eq:NCSN1} for a predetermined $T$. Once ${{\theta }^{*}}$ is acquired through optimizing the objective conditioned on $T$, as shown in \cref{eq:NCSN3}, one can use the approximation ${{\nabla }_{x_{t}}}\log {p}\left( {{x}_{t}} \right)\approx {{s}_{{{\theta }^{*}}}}\left( {{x}_{t}},\text{ }t \right)$ as a plug-in estimate to replace the score function used in the stochastic differential equations~\cite{T-reverseSDE}. As iterative processes continue,
% s for subsequent time steps, 
the final sample is derived from the output obtained at $t = 0$.

\vspace{-0.5mm}
\subsection{Stochastic Differential Equations}
As an extension of NCSNs, SDE and reverse-time SDE can correspondingly model the forward diffusion process and reverse diffusion process, where the forward process is 
% Consistent with the previous notation, ${x}_{0}$ represents data that conforms to the distribution of interest, and the forward process is defined with SDE:
% The forward process modeled by SDE is
% formulated as 
\begin{equation}
  \frac{dx}{dt}=\bar {f}(x,t)+ \bar{g}(t)\omega_{t} 
  \Leftrightarrow \ 
dx=\bar{f}(x,t)dt+\bar{g}(t)d\omega ,\label{eq:11}
\end{equation}
where $\bar{f}(x,t)$ and $\bar{g}(t)$ are diffusion and drift functions of the SDE. ${\omega}\in {{\mathbb{R}}^{n}}$ denotes the standard \textit{n}-dimensional Wiener process. Based on \cref{eq:11}, the reverse process can be modeled with a reverse-time SDE~\cite{T-reverseSDE}, which is
\begin{equation}
    dx=[\bar{f}(x,t)-\bar{g}{{(t)}^{2}}{{\nabla }_{x}}\log {{p}_{t}}(x)]dt+\bar{g}(t)d\bar{\omega },\label{eq:12}
\end{equation}
where $d\bar{\omega }$ denotes the infinitesimal negative time step, defining the standard Wiener process running backward in time. Solutions to the reverse-time SDE are diffusion processes that gradually convert noise to data. Note that the reverse SDE defines the generative process through the score function ${{\nabla }_{x}}\log p(x)$, a shared concept in~\cref{Sec:NCSN}.

During both train and inference phases, SDE-based methods rely on practical numerical sampling techniques. Alongside numerical solutions discussed in~\cref{Sec:NCSN}, methodologies like Euler-Maruyama discretization and Ordinary Differential Equations (ODEs)~\cite{weinan2017proposal} are effective, with the latter offering better sample efficiency advantages.

If the score function ${{\nabla }_{x}}\log p(x)$ is known, we can solve the reverse-time SDE easily.
% Like the optimization scheme in NCSNs,
By generalizing the score-matching optimization objective in NCSNs
% in Eq.\ref{eq:NCSN3} 
to continuous time, we parameterize a time-dependent score model $s_{\theta}(x_t,t)$ to estimate the score function in reverse-time SDE, bringing the same optimization objective as \cref{eq:NCSN2}.

Comparing the expansion result of the score function that uses Bayes' rule with the noise result obtained from \cref{eq:4}, it is easy to observe that the training objectives for DDPMs and NCSNs are equivalent, as shown in \cref{eq:sde3}. Namely, the optimization learning objectives of both methods only differ by a fixed scaling factor:
\begin{equation}
    {{s}_{\theta }}({{x}_{t}},t)=-\frac{1}{\sqrt{1-{{{\bar{\alpha }}}_{t}}}}{{\epsilon }_{\theta }}({{x}_{t}},t).
    \label{eq:sde3}
\end{equation}

Moreover, when generalizing to the case of infinite time steps or noise levels, both DDPMs and NCSNs can be considered as discrete numerical solutions of SDEs in practical applications. For example, the Variance Preserving (VP) \cite{R-beatsGAN} form of the SDE can be perceived as the continuous version of DDPM \cite{T-DDPM}, and the corresponding SDE is
\begin{equation}
    dx=-\frac{1}{2}\beta (t)xdt+\sqrt{\beta (t)}d\omega, 
\end{equation}
where $\beta (\frac{t}{T})=T{{\beta }_{t}}$ as $T$ goes to infinity. NCSNs with annealed Langevin dynamics are equivalent to the discrete version of Variance Exploding (VE) SDE \cite{R-beatsGAN}, which is
\begin{equation}
    dx=\sqrt{\frac{d[\sigma {{(t)}^{2}}]}{dt}}d\omega ,
\end{equation}
where $\sigma (\frac{t}{T})={{\sigma }_{t}}$ as $T$ goes to infinity.
\begin{figure}[t]
    \centering
    \setlength{\abovecaptionskip}{0.1cm}
    \includegraphics[width=\linewidth]{images/Generative_model\_pipeline/Generative\_models3\_New10.png}
    \caption{{The flowcharts of generative models, where the HQ image $\tilde{x}$ is generated by the corresponding methods, \textit{i.e.}, LACR-VAE~\cite{wu2024light-LACR-VAE}, LLFlow~\cite{wang2022LLFlow}, Vanilla GAN~\cite{P-GAN}, PyDiff~\cite{zhou2023pyramid-LLIE2}.}}\vspace{-5mm}
\label{fig:flowchart}
\end{figure}

\subsection{Comparisons With Other Deep Generative Models}

{In this subsection, we examine the connections between DMs and other generative models, presenting a unified mathematical framework for these methods. Flowcharts in \cref{fig:flowchart} illustrate their learning objectives, advantages, and limitations. As highlighted in \cref{fig:flowchart}, a key limitation of DMs is their sampling inefficiency. To address this, approaches such as \cite{T-LDM} draw inspiration from Variational Autoencoders (VAEs), employing an encoder-decoder framework to accelerate the diffusion process within a compressed latent space.}

Both DMs and variational autoencoders (VAEs) \cite{P-VAE,deng2022pcgan} involve mapping data to a latent space, where the generative process learns to transform the latent representations back into data. In both cases, the objective function can also be derived as a lower bound of the data likelihood. However, while the latent representation in VAEs contains compressed information about the original image,  classical assumptions suggest that DMs destroy the data after the final step of the forward process.  Furthermore, the latent representations in diffusion models have the same dimensions as the original data, whereas VAEs tend to perform better with reduced dimensions. In this case,
% Drawing inspiration from these similarities, 
some existing work has explored the use of diffusion models on the latent space of a VAE to build more efficient models \cite{T-LDM,P-VAE-DM1}, or to construct hybrid models that fully leverage the advantages of both models.

{Normalizing flows (NFs) \cite{P-NF1,P-NF2} transform a simple Gaussian distribution into a complex data distribution through a series of invertible functions with easily computable Jacobian determinants. However, the learnable forward process of NFs, unlike that of DMs, imposes additional constraints on the architecture due to its requirement for invertible and differentiable properties. DiffFlow~\cite{P-DiffFlow-3}, serving as a bridge between these two generative algorithms, extends both diffusion models and normalizing flows to enable trainable stochastic forward and reverse processes.}

{Extending traditional normalizing flows, Continuous Normalizing Flows (CNFs) employ Ordinary Differential Equations (ODEs) to model transformations, learning to predict the velocity field that guides the path between distributions through iterative solving. 
Rectified Flow~\cite{liu2022flowrectified-flow} proposes straightening paths between distributions, reducing transport costs and accelerating inference. 
Leveraging this efficiency, Zhu \textit{et al.}~\cite{zhu2024flowie} propose FlowIE, which adapts to diverse degradations via flow rectification and reconstruction. 
By straightening probability transfer trajectories, FlowIE significantly speeds up inference while harnessing pretrained diffusion models. Inspired by Lagrange's Mean Value Theorem, FlowIE optimizes path estimation, achieving fast and effective task enhancement in fewer than five steps. 
Another notable extension is Flow Matching (FM)~\cite{lipman2022flow-Flow-matching}, which refines CNFs by regressing vector fields to align with fixed conditional probability paths. FM optimizes these vector fields by predicting the velocity field that efficiently maps noise to data, offering a simulation-free training alternative.}

{Flow-based models and DMs both aim to map simple distributions to complex data distributions. However, DMs use score-matching to iteratively sample from the target distribution via a stochastic process, while flow-based models transform data deterministically through invertible mappings, allowing for faster computation. Recent large-scale generative models, such as Stable Diffusion 3~\cite{SD3esser2024scaling}, have increasingly adopted FM approaches for enhanced efficiency.
In low-level vision, Martin \textit{et al.}~\cite{martin2024pnpfm} introduce the first Plug-and-Play FM-based method, which alternates between gradient descent steps, reprojections along flow trajectories, and denoising, leading to superior performance across various inverse problems. In fact, by eliminating noise perturbations from the diffusion process and utilizing ODE solvers, results similar to FM can be achieved, suggesting that FM is essentially a specialized variant of DMs. Given the limited application of FM in low-level vision, this topic is not further discussed in this paper.}

GANs \cite{P-GAN} drive the fake data distribution towards the real one through adversarial learning on the generator and the discriminator, ensuring that the sampled data resembles real data. Consequently, GANs are extensively utilized for generating photo-realistic high-resolution images (e.g., PGGAN \cite{P-PGGAN} and StyleGAN series \cite{P-styleGAN}). However, GANs are notorious for their challenging training process due to their adversarial objective~\cite{he2023strategic} and often suffer from mode collapse. In contrast, DMs exhibit a stable training process and offer greater diversity as they are likelihood-based. Despite these advantages, DMs are less efficient than GANs as they require multiple iterative steps during inference.

The distinctions between GANs and DMs also manifest in their ability to manipulate semantic properties within the latent space. GANs' latent space has been observed to contain subspaces associated with visual attributes, enabling attribute manipulation through changes in the latent space and thus facilitating more precise control over generated images. However, DMs manipulate semantic properties of the latent space in a more implicit and less controllable manner. Fortunately, Song \textit{et al.} \cite{T-SDE} demonstrate that DMs' latent space exhibits a well-defined structure. Nonetheless, the exploration of DMs' latent space has been less extensive compared to GANs, indicating the need for further research.

\vspace{-1mm}
\section{Diffusion models for natural image processing in low-level vision
}\label{chap:DM-naturalIR}

We first define "natural images", which depict common scenes and objects encountered in daily life, serving as the foundational input data in model training and evaluation, particularly for image restoration. In this section, "images" is the ordinary and general notion of natural images.

Low-level vision tasks primarily focus on various ill-posed inverse problems in the image restoration domain. These tasks aim to restore degraded and noisy low-quality (LQ) images to high-quality (HQ) images. The general form of the forward model can be stated as
\begin{equation}
    ~y=H({{x}_{0}})+n,\text{    } \ \ y,n\in {{\mathbb{R}}^{n}},{x}_{0} \in {{\mathbb{R}}^{d}},\label{eq:degrade}
\end{equation}
where $H(\cdot ):{{\mathbb{R}}^{d}}\to {{\mathbb{R}}^{n}}$ is the forward 
% measurement 
operator that maps the clean image ${x}_{0}$ to the distorted data $y$. $n$ is the 
% measurement 
noise. 
\begin{figure}[t]
    \centering
        \setlength{\abovecaptionskip}{0.1cm}
    \includegraphics[width=\linewidth]
    {images/linear\_and\_nonlinear.pdf}
    \caption{Linear and nonlinear inverse problems with DMs-based solutions. Figure adapted from~\cite{sp-dps}.}\vspace{-3mm}
\label{fig:linear_and_nonlinear}
\end{figure}

Through rapid development, DM-based models have achieved significant progress in this domain. Unlike random sample generation methods such as vanilla DDPM in~\cref{chap:theorybase}, here the degraded LQ images are used as conditional inputs to guide the latent variables during inference. The models are expected to learn a parametric approximation to the unknown conditional distribution, posterior $p\left( x|y \right)$, through a stochastic iterative refinement process. 

After conducting a comprehensive review of over {300} relevant DM-based works, we classify them from two perspectives, \textit{i.e.}, training manners and application goals.

\subsection{DM-based methods with different training manners}

\noindent\textbf{Supervised DM-based methods}. Supervised DM-based methods tend to specialize in addressing specific degradation scenarios. They employ the well-designed conditional mechanism to incorporate distorted images as guidance during the reverse process, enabling them to tackle several extreme challenges, such as dehazing and deraining, that cannot be effectively modeled using the form of~\cref{eq:degrade}. However, despite yielding promising performance, these methods need training the DM from scratch using paired clean and distorted images from a particular degradation scenario. This results in costly data acquisition and limits the algorithm's generalization to other degradation scenarios.

\noindent\textbf{Zero-shot DM-based methods}. Zero-shot DM-based techniques, leveraging the image priors extracted from pre-trained DMs, offer an appealing alternative as they are plug-and-play without retraining on a specific dataset. The underlying concept is based on the understanding that pre-trained generative models, constructed using extensive real-world datasets such as ImageNet~\cite{D-imagenet}, can serve as a repository of structure and texture. A key challenge lies in extracting the perceptual priors while preserving the underlying data structure from distorted images. Consequently, these zero-shot DM methods are often applied to degradation scenarios simplified as linear reverse problems, such as super-resolution and inpainting. 
Given the simplicity of the application process, which only requires replacing the forward measurement operator, evaluating performance on linear inverse problems has become a common practice to assess the generalization of newly proposed DMs. However, these works are frequently categorized under multi-task alongside other high-level tasks in existing surveys, without receiving systematic analysis and summary. Hence, we devote a specific subsection to introducing these DM-based solvers for general-purpose image restoration in \cref{chap:DM-AG}.

\noindent {\textbf{Discussion}. Owing to the differences in training manners, supervised and zero-shot methods exhibit significant trade-offs in scalability. Supervised methods, optimized for specific datasets, excel in task-specific performance by aligning closely with data distributions and degradations. In contrast, zero-shot methods leverage prior knowledge to model degradations and incorporate the generalizable knowledge embedded in pre-trained models, offering adaptability and competitive performance across diverse tasks.}

\subsection{DM-based methods with different application goals}\label{chap:DM-AG}

\noindent\textbf{General-purpose image restoration}. This section comprises most zero-shot methods and several supervised methods.
Notably, most methods mentioned here presuppose prior knowledge of the forward operator $H(\cdot )$ in \cref{eq:degrade}, confining their scope to non-blind inverse problems. To adhere to specific assumptions, further constraints are occasionally imposed to convert them into linear inverse problems, as shown in \cref{fig:linear_and_nonlinear}. However, the mapping $y\to x_0$ remains many-to-one, rendering it hard to precisely recover $x_0$.

Focusing on sampling from the posterior $p(x|y)$, the relationship can be formally established with the Bayes’ rule: $p(x|y)=p(y|x)p(x)/p(y)$. However, apart from $p(y|{{x}_{0}})\sim \mathcal{N}( y|A( {{x}_{0}} ),\text{ }{{\sigma }^{2}}\bf I)$, there exists no explicit dependency between $y$ and ${{x}_{t}}$, where ${{x}_{t}}$ denotes the noisy results at time step $t$. To solve the intractability of the posterior distribution, Song \textit{et al.}~\cite{T-SDE} propose conditional denoising estimator ${{s}_{\theta }}\left( x,\text{ }y,\text{ }t \right)$. The condition $y$ is added to the input of the estimator to learn an approximation to the posterior score function ${{\nabla }_{{{x}_{t}}}}\text{log }p\left( {{x}_{t}}|y \right)$ without altering the training object. %Instead of direct learning, 
The diffusive estimator jointly diffuses $x$ and $y$ and then learns the posterior approximated from the joint distribution $p\left( {{x}_{t}},\text{ }{{y}_{t}} \right)$ using denoising score matching. Batzolis \textit{et al.}\cite{sp-conditional-Batzolis} rigorously prove the effect of the above two methods theoretically and analyze the errors caused by the imperfections.

\begin{figure}[t]
    \centering
        \setlength{\abovecaptionskip}{0.1cm}
    \includegraphics[width=0.75\linewidth]
    {images/ILVR\_guidance.pdf}
    \caption{Guiding generation process in ILVR \cite{sp-ilvr}.}\vspace{-3mm}
\label{fig:guidance}
\end{figure}
To enhance consistency, \cite{sp-chungmanifoldconstraints} and \cite{sp-ilvr} guide the gradient towards high-density regions by conditioning it through projections on the subspace.
% without directly learning the posterior score function. 
Chung \textit{et al.}\cite{sp-chungmanifoldconstraints} introduce the manifold constraint after the update step, correcting deviations from the data consistency. Using pre-trained DDPM, Choi \textit{et al.}\cite{sp-ilvr} propose Iterative Latent Variable Refinement (ILVR). As shown in \cref{fig:guidance}, ILVR is a learning-free method adopting low-frequency information from $y$ to guide the generation towards a narrow data manifold. However, such methods are limited to those noiseless inverse problems.

Besides the above learning-free methods, plug-and-play posterior sampling provides a favorable choice. 
Graikos \textit{et al.}\cite{graikos2022diffusion} first showcase the viability of directly using pre-trained DDPMs as plug-and-play modules.
% that involve other differentiable constraints. 
Kawar \textit{et al.}\cite{sp-DDRM} propose the Denoising Diffusion Restoration Models (DDRM) to reconstruct the missing information in $y$ within the spectral space of $H(\cdot )$ using Singular Value Decomposition (SVD). Leveraging pre-trained DMs, DDRM demonstrates versatility across several tasks, including SR, deblurring, inpainting, and colorization.

Zhu \textit{et al.}\cite{sp-diffpir} decouple the data term and the prior term with Half-Quadratic-Splitting and propose DiffPIR, handling a wide range of degradation models with different degradation operators $H(\cdot )$. Wang \textit{et al.}\cite{sp-ddnm} propose to solve zero-shot image restoration using Denoising Diffusion Null-space Model (DDNM). The pseudo-inverse computes the low-dimensional representation, then decomposed into its range and null-space contents. By refining the null-space in the reverse process, DDNM learns missing information in image inverse problems while fitting only linear operators.

\begin{figure}[t]
    \centering
    \setlength{\abovecaptionskip}{0.1cm}
    \includegraphics[width=\linewidth]
    {images/pipelines/IDM\_pipeline.pdf}
    \caption{Outline of the IDM~\cite{sp-idm} framework. 
    % Top Section: General workflow of the inference process. Bottom Section: Elaborate depiction of a denoising stage.
    }\vspace{-3mm}
\label{fig:IDM}
\end{figure}
Methods based on Schrödinger bridges, \textit{i.e.}, InDI\cite{sp-InDI} and I2SB\cite{sp-i2sb}, revisit DMs' assumptions and depart from commencing the reverse diffusion process from Gaussian noise, ensuring efficiency. 
Chung \textit{et al.}\cite{chung2023direct} 
% demonstrate the equivalence of the above methods and 
propose the Consistent Direct Diffusion Bridge (CDDB), incorporating a novel data consistency module, to realize the generalization of Schrödinger bridges on low-level vision tasks.

To mitigate the computational overhead, DMs are shifted from the image level to the vector level. Rombach \textit{et al.}\cite{T-LDM} propose latent diffusion models (LDMs), where both the forward and reverse processes occur in the latent space obtained through an auto-encoder. To balance latent disentanglement and high-quality reconstructions, Pandey \textit{et al.}\cite{sp-pandey2022diffusevae} integrate VAEs within DM and propose DiffuseVAE, offering novel conditional parameterizations for DMs. 
% and providing a promising alternative for hybrid modeling.

Due to prevalent limitations of various presuppositions, these models are applied to relatively simple degradation scenarios that can be abstracted and simplified as linear inverse problems. Consequently, they are less effective in real-world blind tasks compared to task-specific methods.

\noindent\textbf{Super-resolution (SR)}. DMs have shown prowess in generating high-quality outputs with intricate details, addressing over-smoothing and artifacts for high-resolution SR\cite{sp-diracdiffusion}. 
SRDiff\cite{sp-srdiff} is the pioneering DM-based single-image SR model, using a pretrained low-resolution encoder and a conditional noise predictor to produce diverse and realistic SR predictions. This effectively addresses over-smoothing and large footprint issues in previous methods\cite{he2023degradation}. 
% Moreover, SRDiff introduces residual prediction and a Markov chain to expedite convergence and stabilize training.  
% Saharia \textit{et al.}\cite{sp-saharia2022SR} also leverage a conditional diffusion network, employing low-resolution images as conditional inputs to resolve SR tasks, particularly for human faces.

Cascaded Diffusion Models (CDM) \cite{sp-CDM} proposes to arrange multiple DMs. The initial model generates low-resolution images based on classes while subsequent models progressively generate images with higher resolutions, facilitating SR at arbitrary magnifications. {Leveraging the advantages of residual modeling, Yue \textit{et al.}~\cite{yue2024resshift} achieve competitive results in SR within just a few steps. The proposed ResShift establishes a Markov chain between the HR/LR image pair by shifting their residual, along with an intricately designed noise schedule for precise controlling. Wang \textit{et al.}~\cite{wang2024sinsr} achieve further breakthroughs in acceleration with SinSR, which performs SR in a single sampling step. By deriving a deterministic sampling strategy from SOTA methods like ResShift, the distilled student models with a consistency-preserving loss match or even surpass teacher methods, achieving up to a tenfold speedup in inference.}

\begin{figure}[t]
    \centering
    \setlength{\abovecaptionskip}{0.1cm}
    \includegraphics[width=\linewidth]
    {images/pipelines/Repaint\_pipeline.pdf}
    \caption{Overview of RePaint~\cite{sp-IRSDE}.}\vspace{-3mm}
\label{fig:RePaint}
\end{figure}
Gao \textit{et al.}\cite{sp-idm} propose implicit DMs for continuous SR (in \cref{fig:IDM}). They introduce a scale-adaptive mechanism to adjust the ratio of realistic data and use implicit neural representation to capture complex structures across continuous resolutions. Niu \textit{et al.}~\cite{niu2023cdpmsr} first use a pretrained SR model to generate high-resolution inputs. Besides, they propose a $n^{th}$ order sampler to perform a deterministic denoising process, reducing the iteration number. Wang \textit{et al.}~\cite{p-crossmodal-accelerate3-wang2023exploiting} propose StableSR to leverage prior knowledge contained in pretrained text-to-image DMs for blind SR. 
By utilizing a time-aware encoder, StableSR achieves promising restoration results without modifying the pretrained synthesis model.

% Similarly leveraging generative diffusion priors, 
Lin \textit{et al.}~\cite{lin2023diffbir} use generative priors to design DiffBIR for blind image SR, decoupling the restoration process into two stages. 
{Sun \textit{et al.}~\cite{sun2024coser} propose CoSeR, which leverages generative images from a pretrained LDM as implicit priors. It combines generated results with low-resolution priors and CLIP's semantic priors~\cite{radford2021learning} to control the diffusion process. Yu \textit{et al.}~\cite{yu2024scalingsupir} introduce SUPIR, further leveraging multi-modal techniques and advanced generative priors. By incorporating textual prompts into the restoration process, SUPIR guides the model to better understand and reconstruct severely degraded images. This enhances perceptual quality and enables user-defined, targeted restoration.}

\noindent\textbf{Inpainting}. 
As a probabilistic generative model, DMs exhibit robust generalization across different masks and effectively handle large missing regions. RePaint~\cite{sp-repaint} employs an enhanced denoising strategy involving resampling iterations to better condition images in \cref{fig:RePaint}. RePaint first generates a rough estimate and then refines it by a DM with a Markov random field. 
To specify a desired inpainted object, Gebre \textit{et al.}~\cite{gebre2024fill} input an extra target image to guide the generation of the masked region, providing valuable exploration in the controllable generation. 
Zhang \textit{et al.}~\cite{zhang2024mmginpainting} employ both image and text as multi-modal guidance. By integrating the inverse process with CLIP, semantic information is better encoded, thus enhancing controllability.
\begin{figure}[t]
    \centering
    \setlength{\abovecaptionskip}{0.1cm}
    \includegraphics[width=0.95\linewidth]
    {images/pipelines/Deblurring\_pipeline.pdf}
    \caption{Overview of the method proposed in~\cite{sp-whang2022deblurring}.}\vspace{-5mm}
\label{fig:Deblurring_pipeline}
\end{figure}

% To  and preserve textures, 
Spatial DM \cite{p-li2022sdm} employs a Markov random field to estimate the missing pixels, which considers surrounding contexts and thus inpaints large missing regions. 
Saharial \textit{et al.}\cite{sp-palette} introduce Palette to explore diverse optimization objectives and highlight self-attention. 
% However, the above methods are constrained to specific resolutions post-training.
BrushNet~\cite{ju2024brushnet} is a plug-and-play model and embeds pixel-level masked features into any pre-trained DMs by separating masked features and noisy latent.
% into separate branches.
Grechka \textit{et al.}~\cite{grechka2024gradpaint} propose a training-free DM, GradPaint, for gradient-guided inpainting, aiming to improve the coherence and realism of generated images.

\noindent\textbf{Deblurring}. 
DMs in realistic deblurring often rely on hand-designed networks.
Wang \textit{et al.}\cite{sp-whang2022deblurring} first introduce DMs (in \cref{fig:Deblurring_pipeline}) into deblurring, proposing a ``predict-and-refine'' conditional DM. This architecture comprises a deterministic data-adaptive predictor and a stochastic sampler, refining the output through residual modeling.
Ren \textit{et al.}\cite{ren2023multiscale} introduce multiscale structure guidance in image-conditioned DPMs for deblurring. Their guidance module projects the input into a multiscale representation and the guidance is integrated into intermediate layers as an implicit bias, thus enhancing robustness.
Hierarchical Integration Diffusion Model (HI-Diff)~\cite{sp-chen2023hierarchical} leverages LDM to generate priors and fuse these priors through a cross-attention mechanism, enabling generalization in complex scenarios.

Laroche \textit{et al.}~\cite{p-laroche2024fast} propose a DM-based blind image deblurring method. This method integrates DMs with the Expectation-Minimization (EM) estimation to jointly estimate restored images and the unknown blur kernel. Spetlik \textit{et al.}~\cite{p-spetlik2024single} propose a DDPM-based method for single-image deblurring and trajectory recovery of fast-moving objects, getting competitive results to multi-frame methods. DiffEvent~\cite{p-wang2024diffevent} firstly introduces DMs into event deblurring. To adapt to real-world scenes, DiffEvent builds an Event-Blur Residual Degradation (EBRD) to provide pseudo-inverse guidance, enhancing subtle details and handling unknown degradation. Luo \textit{et al.}\cite{sp-IRSDE} propose the Image Restoration Stochastic Differential Equation (IR-SDE), whose core is a mean-reverting SDE with a maximum likelihood objective. This ensures that the entire SDE will diffuse towards the mean $\mu$ (low-quality image) with specific Gaussian noise. Owing to its ability to simulate the degradation process, IR-SDE also excels in super-resolution and inpainting.

 \begin{figure}[t]
    \centering
    \setlength{\abovecaptionskip}{0.1cm}
    \includegraphics[width=\linewidth]
    {images/pipelines/PyramidLLIE\_pipeline.pdf}
    \caption{Overview of PyDiff~\cite{zhou2023pyramid-LLIE2}.}\vspace{-5mm}
\label{fig:PyDiff}
\end{figure}
\noindent\textbf{Dehazing, deraining, and desnowing}. 
As aforementioned, real-world degradations like dehazing and deraining are complex and cannot be effectively modeled by a prior operator $H(\cdot )$. Consequently, they pose challenges for incorporation into general-purpose image restoration frameworks.

Özdenizci \textit{et al.}~\cite{sp-WeatherDiff} present a patch-based image restoration algorithm termed WeatherDiffusion. This approach facilitates size-agnostic image restoration by employing a guided denoising process with smoothed noise estimates across overlapping patches during inference, mitigating the drawbacks of merging artifacts from independently restored intermediate results. WeatherDiffusion achieves superior performance on both weather-specific and multi-weather image restoration tasks, including dehazing, desnowing, deraining~\cite{jin2024raindrop}, and raindrop removal. 

Building upon IR-SDE, Luo \textit{et al.}~\cite{sp-luo2023refusion} further enhance it to perform restoration in a low-resolution latent space, which constitutes a resolution-agnostic architecture. This enhancement offers another viable option for handling large-size images. Wang \textit{et al.}~\cite{wang2024frequency} propose a Frequency Compensation block, equipped with a bank of filters that collectively amplify the mid-to-high frequencies of an input signal, enhancing the reconstruction of image details and improving generalization to real haze scenarios.

\noindent\textbf{Low-light image enhancement}. 
Compared to the black-box design in other tasks, a plethora of research related to DMs has emerged in low-light image enhancement (LLIE).
Zhu \textit{et al.}~\cite{p-llie-zhu2023diffusion} first introduce DMs into LLIE within space-based visible cameras. This method effectively reduces computational complexity by diffusing processes on grayscale images and supplementing features with RGB images. Wu \textit{et al.} \cite{p-llie-wu2023difflie} focus on restoring pure black images, providing a robust generative network for enhancing low-light images with diverse outputs. Zhou \textit{et al.} \cite{zhou2023pyramid-LLIE2} propose the Pyramid Diffusion model named PyDiff (illustrated in \cref{fig:PyDiff}) for LLIE, which increases the resolution during the reverse process, reducing computational burden. Jiang \textit{et al.} \cite{jiang2023low-LLIE} introduce a wavelet-based conditional diffusion model, which proposes a high-frequency restoration branch module to provide extra vertical and horizontal details. Wang \textit{et al.} \cite{p-llie-wang2023exposurediffusion} integrate DMs with a physics-based exposure model in the raw image space, where the reverse process can start from a noisy image, boasting fast inference speed.

Some methods that integrate DMs with other advanced techniques have yielded superior results. Hou \textit{et al.}~\cite{p-llie-hou2024global} introduce a global structure-aware regularization to constrain the intrinsic structures, along with an uncertainty-guided regularization to relax constraints on extreme situations. Diff-Retinex\cite{p-llie-yi2023diff} decomposes the image into illumination and reflectance maps and then uses multi-path DMs to estimate the clean image. Adopting the opposite strategy, He \textit{et al.}~\cite{he2023reti-LLIE3}  propose a Retinex-based LDM to extract reflectance and illumination priors, and then perform decomposition and enhancement using a Retinex-guided transformer, achieving superior results.
Yin \textit{et al.}~\cite{p-llie-yin2023cle} achieve an interactive and controllable LLIE model based on a conditional DM. Users can customize the brightness level and enhance specific target regions with the Segment Anything Model~\cite{kirillov2023segment}. {To fully utilize the CLIP-based model prior, Xue \textit{et al.}~\cite{p-llie-xue2024low} introduce multi-modal visual-language information and propose a novel approach named CLIP-Fourier Guided Wavelet Diffusion (CFWD). CFWD combines the strengths of wavelet transform, Fourier transform, and CLIP to guide the DM-based enhancement process in a multiscale visual-language manner, demonstrating the immense potential of integrating semantic features from CLIP and high-frequency detail recovery from the Fourier transform.}

\noindent\textbf{Image fusion}.
Image fusion can elevate the overall visual quality and facilitate diverse downstream applications.
Yue \textit{et al.} \cite{p-if-yue2023dif} propose the first DM-based method, Dif-Fusion, for image fusion (see in \cref{fig:Dif-Fusion}). By creating a multi-channel data distribution, Dif-Fusion enhances color fidelity in infrared-visible image fusion (IVF). {Guo \textit{et al.}~\cite{guo2024glad} propose GLAD, which leverages DMs to capture the joint distribution of multi-channel data, addressing texture and edge blurring. Li \textit{et al.} \cite{p-if-li2024fusiondiff} apply the DDPM model to the multi-focus image fusion task, showcasing excellent performance in terms of noise resistance.}
 \begin{figure}[t]
    \centering
    \setlength{\abovecaptionskip}{0.1cm}
    \includegraphics[width=\linewidth]
    {images/pipelines/Dif\_fusion\_pipeline.pdf}
    \caption{The overall framework of Dif-Fusion~\cite{p-if-yue2023dif}.}\vspace{-4mm}
\label{fig:Dif-Fusion}
\end{figure}

Zhao \textit{et al.}\cite{p-if-zhao2023ddfm} propose DDFM for IVF and divide the problem into an unconditional DDPM for utilizing image generation priors and a maximum likelihood sub-problem for preserving cross-modal information of source images, generating visually fidelity results. {Diff-IF~\cite{yi2024diff-if} breaks down the diffusion process into a conditional DM and multi-modal fusion knowledge prior, which is used to guide the forward diffusion process. Cao \textit{et al.}\cite{p-RS-imagefusion-cao2023ddrf} devise two injection modulation modules to introduce coarse-grained style information and fine-grained frequency information, achieving state-of-the-art results. Yang \textit{et al.}~\cite{yang2025lfdt} introduce LFDT-Fusion for general image fusion, which compresses inputs into a low-resolution latent space and employs a transformer-based denoiser to achieve the diffusion process.}

\noindent {\textbf{Discussion}. Various task-specific DM modifications mentioned in \cref{chap:DM-AG} impact interpretability and generalizability. For instance, latent space compression~\cite{T-LDM} facilitates the acquisition of generalized latent representations, while such representations are inherently compact, thus reducing interpretability. Hybrid models~\cite{he2023reti-LLIE3,sp-diffir}, leverage DM priors to guide and improve other methods, enhancing controllability and validating interpretability through explicit prior usage. Integrating the strengths of different frameworks, hybrid models also achieve superior generalizability.}

\vspace{-2mm}
\section{Extended diffusion models}
\label{chap:DM-extendIR}

\subsection{Diffusion models for medical image processing}
Compared with natural data, medical data acquisition typically involves more intricate and precise physical imaging processes~\cite{R-dataeffcientMIA}, resulting in poor image quality due to equipment and usage limitations (\textit{e.g.}, hospital throughput requirements, patient examination time constraints, and radiation dosage limits). Leveraging the robust learning capacity of DMs, these models can implicitly capture knowledge related to imaging physics from dataset distributions. Hence, DM-based methods have been introduced to address low-quality medical images degraded by imaging limitations, \textit{e.g.}, limited-angle computed tomography (CT) and accelerated magnetic resonance imaging (MRI).

In addition to enhancing low-quality data, another key application of DM-based methods is the generation of missing modalities. In disease diagnosis, the combination of multi-modal data assists doctors in making more accurate diagnoses. 
However, certain rarer medical images (\textit{e.g.}, Positron Emission Computed Tomography (PET) and Optical Coherence Tomography (OCT)) unavoidably contain speckle noise that traditional methods fail to eliminate. Due to the nature of generative models in detail reconstruction, diffusion models are well-suited for addressing such issues.

To provide a multi-perspective categorization, we will classify methods according to their imaging modalities, covering MRI, CT, multi-modal, and other modalities.
\begin{figure}[t]
    \centering
    \setlength{\abovecaptionskip}{0.1cm}
    \includegraphics[width=\linewidth]
    {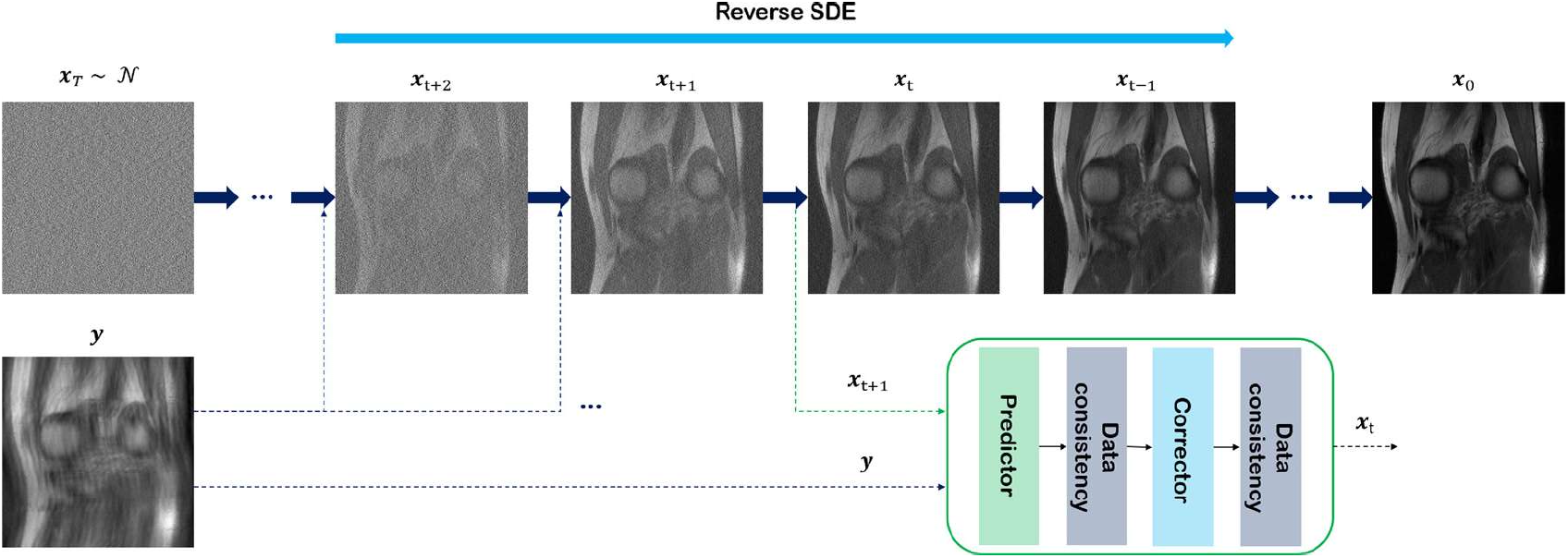}
    \caption{{Overview of DiffAMRI \cite{chung2022scoreMRI}. 
    }}\vspace{-4mm}
\label{fig:MRI}
\end{figure}

\noindent\textbf{MRI}. 
MRI involves a time-consuming imaging process, where patient movement can lead to various artifacts. Hence, medical image reconstruction is necessary to achieve faster acquisition speed. Chung \textit{et al.}\cite{chung2022scoreMRI} design a score-based framework for accelerated MRI reconstruction, shown in \cref{fig:MRI}. 
They train a time-dependent score function using score matching on magnitude images and employ the VE SDE for sampling distribution from the pre-trained score model. By applying data consistency mapping, this approach effectively handles multi-coil images and exhibits robust generalization to different subsampling patterns.

{Ozturkler \textit{et al.}~\cite{SMRD-ozturkler2023smrd} propose SMRD, integrating Stein's Unbiased Risk Estimator into the sampling stage of DMs for automatic hyperparameter tuning. SMRD addresses the reliance on validation-based hyperparameter tuning, offering a more automated solution.
Güngör \textit{et al.}~\cite{AdaDiff-gungor2023adaptive} present AdaDiff
% , the first adaptive prior-based diffusion method 
for MRI reconstruction. AdaDiff uses an adaptive diffusion prior trained via adversarial mapping over a two-phase process: a rapid-diffusion phase for initial reconstruction, followed by an adaptation phase for prior refinements. 
Similarly, DiffuseRecon~\cite{DiffuseRecon-peng2022towards} leverages a pre-trained diffusion model with under-sampled signals gradually guiding the reverse diffusion process. This shows robustness to varying acceleration factors without requiring retraining. 
% It also employs a Monte Carlo sampling scheme to achieve significant speed-ups.
Korkmaz \textit{et al.}~\cite{SSDiffRecon-korkmaz2023self} propose SSDiffRecon, a self-supervised method that constructs training pairs by randomly masking under-sampled k-space data. By further combining 
% cross-attention transformers and 
data consistency blocks, SSDiffRecon can accurately model complex data distributions, improving reconstruction reliability.}

\begin{figure}[t]
    \centering
    \setlength{\abovecaptionskip}{0.1cm}
    \includegraphics[width=\linewidth]
    {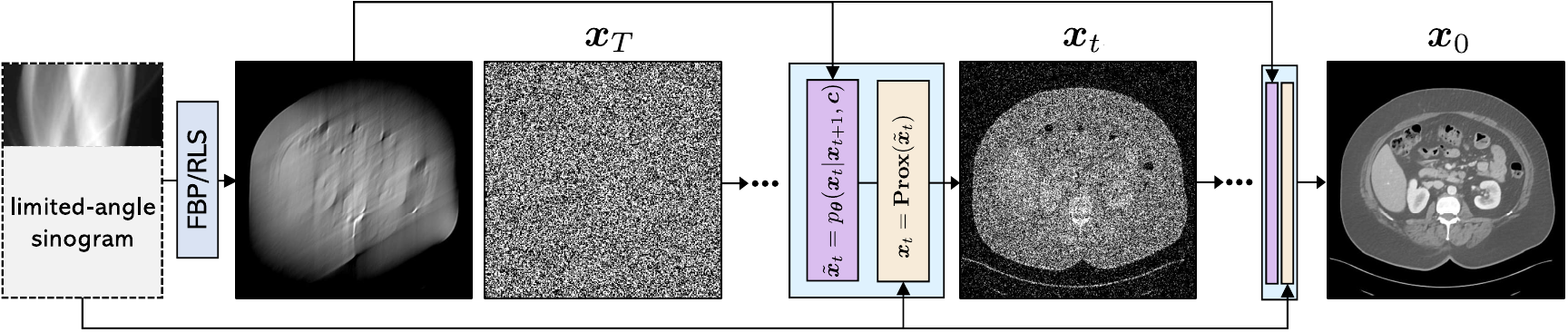}
    \caption{{Overview of DOLCE \cite{sp-LACT-liu2023dolce}. 
    % Starting from the Gaussian noise $x_{T}$ , then sample an image $x_{0}$ from the posterior by solving the reverse process of conditional denoising diffusion model (cDM), alternating between the denoising-update step and the data-consistency step.
    }}\vspace{-4mm}
\label{fig:LACT}
\end{figure}
% \begin{figure}[t]
%     \centering
%     \setlength{\abovecaptionskip}{0.1cm}
%     \includegraphics[width=\linewidth]
%     {images/mmDM.pdf}
%     \caption{Schematics of UMM-CSGM~\cite{meng2022multi-modal}. The upper panel shows the cross-modal diffusion and reverse generation process. The lower panel shows the structure of mm-CSN.}\vspace{-3mm}
% \label{fig:mmDM}
% \end{figure}
\noindent\textbf{CT}. Similar to MRI, limited-angle CT reconstruction has been a primary focus in CT research, aiming to reduce patient radiation exposure and enhance examination throughput. DM-based methods have shown remarkable performance in this reconstruction task. For example, Liu \textit{et al.}~\cite{sp-LACT-liu2023dolce} introduce DOLCE, a method specifically designed for limited-angle CT reconstruction within a DDPM framework. Conventionally, the Filtered Back Projection (FBP) algorithm~\cite{FBP} is employed to map CT images from sinograms, leveraging the Fourier slice theorem. However, limited-angle measurements lead to Fourier measurement loss and subsequently degraded reconstruction outcomes. 

Due to the ill-posed nature, directly using DDPM presents challenges. Following the design in inpainting tasks, DOLCE~\cite{sp-LACT-liu2023dolce} integrates the FBP output on limited sinograms as prior information to condition the diffusion model (\cref{fig:LACT}). Besides, DOLCE enforces a consistency term in the denoising iteration to ensure reconstruction consistency through iterative refinement using proximal mapping in the inference step to meet the consistency conditions presented by sinograms. Evaluation on \textit{C4KC-KiTS} verifies DOLCE's effectiveness in generating high-quality CT images.
% Moreover, the reconstruction performance is further evaluated in downstream tasks such as 3D Segmentation.

\noindent\textbf{Multi-modal medical data}. MRI and CT are the two most widely used medical imaging modalities. MRI shows soft tissues such as vessels and organs in rich contrast while CT is preferred for imaging hard tissues such as bones and interfaces. 
Due to their complementary characteristics, multi-modality imaging with MRI and CT is often used in clinical practice. Therefore, the development of a simultaneous CT-MRI device is currently a hot research topic, and various studies have been carried out to propose advanced designs for such a device\cite{xu2023dm,device2,ju2022ivf}. To translate MR to CT images, Lyu \textit{et al.}\cite{sp-CT2MRI} examine conditional DDPM and SDE models, employing three different sampling methods.

% As shown in \cref{fig:mmDM}, 
Meng \textit{et al.}\cite{meng2022multi-modal} introduce a Unified Multi-Modal Conditional Score-based Generative Model (UMM-CSGM) to complete missing modality images. This model is presented in a conditional SDE, using a multi-in multi-out conditional score network (mm-CSN) module, to learn cross-modal conditional distributions. 
% Experiments on \textit{BraTS19}~\cite{bakas2018identifying} indicate that the method can generate missing-modality images with clear brain structural details. 
Due to inter-modality differences,
% and lacking medical data, 
training DM-based models in a zero-shot manner is not feasible for image translation and can only be applied to certain tasks with low difficulties, \textit{e.g.}, CBCT-to-CT image translation and cross-institutional MRI image translation. For example, Li \textit{et al.}~\cite{sp-FGDM} propose the Frequency-Guided Diffusion Model (FGDM), which uses frequency-domain filters to preserve structure during translation. FGDM enables zero-shot learning and exclusive training on target domain data, allowing direct deployment for source-to-target domain translation. 
% This verifies significant advantages of zero-shot medical image translation over existing methods.

% Due to the vast inter-modality differences coupled with the difficulty in obtaining medical data, it's not feasible to train powerful pre-trained DM-based models in a zero-shot manner for image-to-image translation on a sufficiently large multi-modal medical dataset. However, this approach remains feasible in lower-difficulty translation tasks, such as in the CBCT-to-CT image translation task and cross-institutional MRI image translation task. For instance, Li \textit{et al.}\cite{sp-FGDM} propose a novel strategy known as the Frequency-Guided Diffusion Model (FGDM), employing frequency-domain filters to preserve structure during image translation. FGDM allows zero-shot learning and exclusive training on target domain data. Moreover, it enables direct deployment for source-to-target domain translation. This approach demonstrates significant advantages in zero-shot medical image translation over existing methods.

\noindent\textbf{Other modalities}. PET, crucial for cancer screening, faces challenges related to low SNR and resolution due to the limited beam count radiation during scans. To mitigate the oversmoothing in previous PET denoising methods, Gong \textit{et al.}~\cite{sp-pet} introduce a DDPM-based framework for PET denoising, termed PET-DDPM.
% They utilize the \textit{18F-FDG} and \textit{18F-MK-6240} datasets for PET and MR modalities, respectively. 
PET-DDPM explores the collaboration of diverse modalities to learn noise distribution through PET images. The MR image, serving as the prior, is seamlessly integrated as the input for the denoising network.
% , which is proven to yield good results.
Experiments reveal that employing MR prior as the input while embedding PET images as a data-consistency constraint during inference achieves the best performance.

\begin{figure}[t]
    \centering
    \setlength{\abovecaptionskip}{0.1cm}
    \includegraphics[width=\linewidth]
    {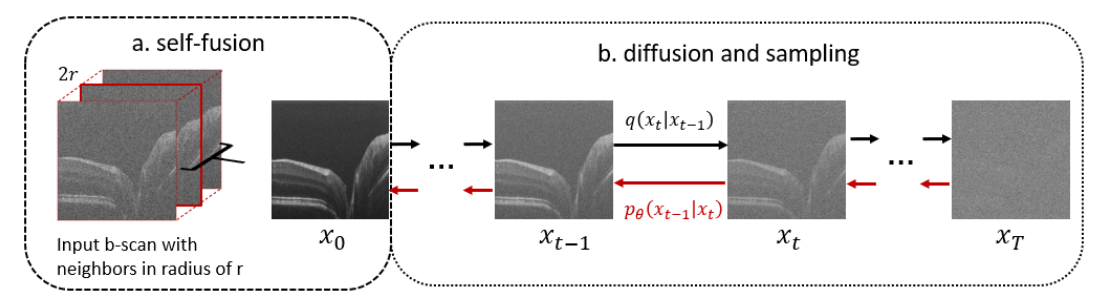}
    \caption{General pipeline of DenoOCT-DDPM~\cite{sp-OCT}.}\vspace{-3mm}
\label{fig:OCT}
\end{figure}

Hu \textit{et al.}\cite{sp-OCT} apply a DDPM to address speckle noise in OCT volumetric retina data in an unsupervised manner called DenoOCT-DDPM, aiming to address the intrinsic challenges of OCT imaging due to restricted spatial-frequency bandwidth.
% OCT imaging encounters challenges due to restricted spatial-frequency bandwidth, resulting in images with speckle noise that hampers ophthalmologist diagnosis and tissue visibility, and conventional methods like averaging multiple B-scans struggle to handle this.
% suffer from prolonged acquisition time and registration artifacts, often amplifying the noise due to its multiplicative nature. 
DenoOCT-DDPM exploits DDPM’s adaptability to noise patterns and
% instead of real-data patterns. It 
incorporates self-fusion as a preprocessing step, feeding the DDPM with a clear reference image for training the parameterized Markov chain (refer to \cref{fig:OCT}), thus eliminating speckle noise while preserving detailed features like small vessels.

\subsection{Diffusion models for remote sensing data }\label{chap:DM-remotesensing}

The versatility of diffusion models makes them well-suited for remote sensing data processing. Their applications span a spectrum of challenges encountered in the analysis of diverse remote sensing modalities, including visible-light images, hyperspectral imaging (HSI), and Synthetic Aperture Radar (SAR). These tasks encompass but are not limited to super-resolution~\cite{p-RS-SR-HSI-wu2023hsr,p-RS-SR-HSI-shi2023hyperspectral,p-RS-SR-liu2022diffusion}, despeckling\cite{p-SAR-despecking-perera2023sar,p-SAR-despecking-xiao2023unsupervised}, cloud removal\cite{p-RS-cloudremoval-zhao2023cloud,p-RS-cloudremoval-jing2023denoising,p-RS-cloudremoval-badheimplementation}, multi-modal fusion\cite{p-RS-imagefusion-cao2023ddrf}, and cross-modal image translation\cite{p-RS-imagetransition-seo2023improved}.

We continue categorizing these works based on the imaging modality, examining the significant impact of DMs.
% and advancements brought about by diffusion models.
% in addressing low-level vision tasks on remote sensing data.

\noindent\textbf{Visible-light remote sensing data}. 
% Compared with natural images, DM-based methods in remote sensing remain underdeveloped. 
Visible-light Remote Sensing Images share a high similarity with natural images. In this case, Sebaq \textit{et al.}\cite{p-RS-sebaq2023rsdiff} employ techniques similar to Imagen~\cite{P-saharia2022photorealistic-Imagen} for low-resolution generation and reference the SR pipeline of CDM~\cite{sp-CDM}, constructing a powerful framework for high-resolution satellite imagery generation.

\begin{figure}[t]
    \centering
        \setlength{\abovecaptionskip}{0.1cm}
    \includegraphics[width=\linewidth]
    {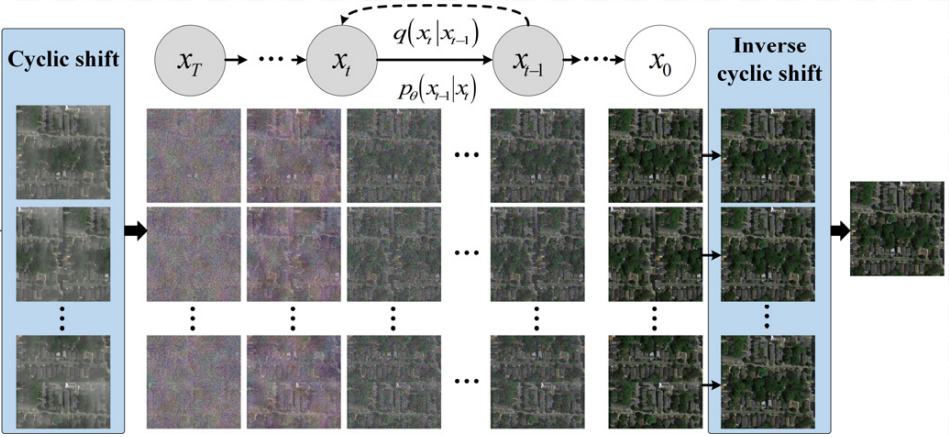}
    \caption{{Architecture of RSDDM \cite{P-RS_dehaze-huang2023remote} for RS dehazing.}}\vspace{-5mm}
\label{fig:RS-dehaze}
\end{figure}

Given that RS images suffer from detail loss,  Liu \textit{et al.}\cite{p-RS-SR-liu2022diffusion} propose the first DM for Remote Sensing Super-Resolution and introduce a supplement inpainting task through random masking, aiming to enhance the recovery ability for specific small objects and complex scenes. 
% Besides, they introduce a joint loss to suppress the undesirable excessive diversity.
% , which is undesirable in the context of RSSR tasks. 
Considering that RS images often have higher resolution and exhibit unusual sizes, Huang \textit{et al.}\cite{P-RS_dehaze-huang2023remote} introduce an Adaptive Region-Based DM (in \cref{fig:RS-dehaze}) to address arbitrary RS image dehazing tasks. They employ the cyclic shift strategy\cite{cyclic-coifman1995translation} to eliminate inconsistent color and artifacts.
% resulting from region cropping.

\noindent\textbf{Hyperspectral imaging}. HSI is a crucial modality in remote sensing with widespread applications. However, due to the limitations of imaging devices, HSIs suffer from data-hungry, noise corruption, and low spatial resolution. 
Zhang \textit{et al.}\cite{P-RS-HSI-zhang2023r2h} propose the first DM
% Diffusion Density Power Model 
% conditioned on RGB natural images 
for HSI generation. The authors employ a spectral folding technique to achieve spectral-to-spatial mapping, 
% effectively 
addressing the convergence challenges 
% associated with the large spatial sampling space of HSI images 
due to their high channel count. 
Deng \textit{et al.}\cite{deng2023noise} propose a DM-based model for HSI denoising, utilizing random masking, resembling the one in \cite{p-RS-SR-liu2022diffusion}, to balance spatial and spectral information for performance improvement.

\begin{figure}[t]
    \centering
        \setlength{\abovecaptionskip}{0.1cm}
    \includegraphics[width=\linewidth]
    {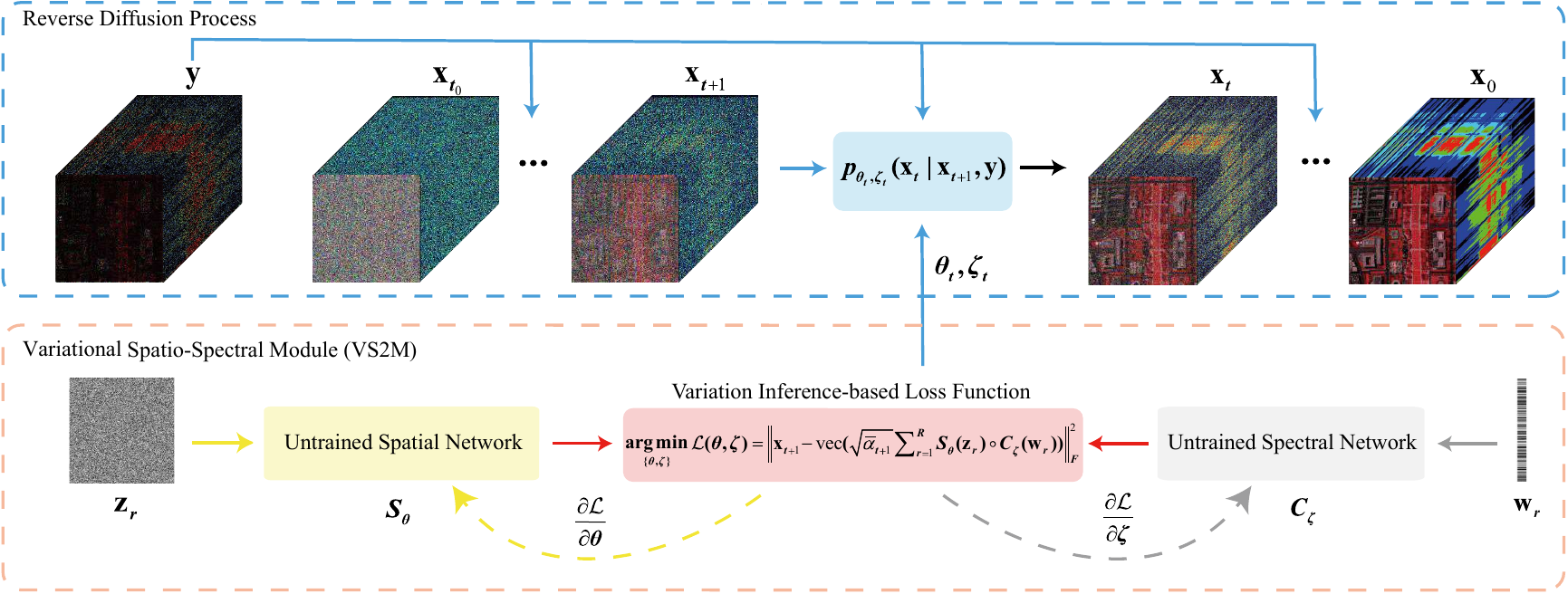}
    \caption{An overview of the self-supervised DDS2M in\cite{miao2023dds2m}.}\vspace{-5mm}
\label{fig:DDS2M}
\end{figure}

As shown in \cref{fig:DDS2M}, Miao \textit{et al.}\cite{miao2023dds2m} introduce an innovative self-supervised DM, DDS2M, for HSI restoration, addressing the data-hungry issue. DDS2M leverages the variational spatio-spectral module, comprising two untrained networks, each focusing on the spatial and spectral dimensions, to exploit the intrinsic structural information of the underlying HSIs. 
By introducing prior information, DDS2M can learn the posterior distribution solely using the degraded HSI.
% without extra training data. 
Experiments on HSI denoising and noisy HSI completion
% , and super-resolution 
verify the superiority of DDS2M. 
% DDS2M provides new insight into integrating existing DMs with untrained networks and offers a promising solution for HSI restoration.

% The spatial and spectral resolutions of spectral images are often challenging to balance due to limitations in imaging technology. 
To balance the spatial and spectral resolutions of spectral images,
Wu \textit{et al.}\cite{p-RS-SR-HSI-wu2023hsr} propose HSR-Diff, the first diffusion model for HSI Super-resolution. The model fuses high-resolution multispectral images with low-resolution hyperspectral images (LR-HSI) to obtain HR-HSI. 
% The conditional DDPM uses the Conditional Denoising Transformer, which replaces the time embedding with noise embedding, designing spatio-spectral transformer layers for HSI characteristics. 
Shi \textit{et al.}\cite{p-RS-SR-HSI-shi2023hyperspectral} employ a similar approach and demonstrate the effect of DM-based models on multiple remote sensing datasets.

% \begin{figure}[t]
%     \centering
%         \setlength{\abovecaptionskip}{0.1cm}
%     \includegraphics[width=\linewidth]
%     {images/cloudremoval.png}
%     \caption{A summary of CRRS \cite{p-RS-cloudremoval-zhao2023cloud} in training and inferential strategies based on sequences.}\vspace{-3mm}
% \label{fig:cloudremoval}
% \end{figure}

\noindent\textbf{Synthetic Aperture Radar}. Tuel \textit{et al.}\cite{p-SAR-tuel2023diffusion} pioneer the use of diffusion models for radar remote sensing imagery.
% , achieving tasks related to SAR image generation and denoising. 
This method highlights, due to limited data, the lack of powerful feature extractors specific to remote sensing data as a major bottleneck for high-quality generation. Speckle, a type of signal-dependent multiplicative noise affecting coherent imaging modalities including SAR images, is addressed by Perera \textit{et al.}\cite{p-SAR-despecking-perera2023sar}, who introduce DDPM to SAR despeckling. Besides, a new inference strategy based on cycle spinning is proposed to further improve performance. 
Xiao \textit{et al.}\cite{p-SAR-despecking-xiao2023unsupervised} transform multiplicative noise into traditional additive noise through operations in the logarithmic domain for DM-based denoising. 
This method introduces a patch-shifting and averaging-based algorithm to adapt to inputs of arbitrary resolutions, further enhancing performance.

% \begin{figure}[t]
%   \centering
%       \setlength{\abovecaptionskip}{0.1cm}
%   \includegraphics[width=\linewidth]{images/pipelines/SATeCo\_pipeline.pdf}
% \caption{The architecture of SATeCo proposed in~\cite{p-chen2024learning-videoSR}.}\vspace{-5mm}
% \label{fig:SATeCo}
% \end{figure}
\noindent\textbf{Muti-modal remote sensing data}. SAR images are robust to weather conditions but are hard to interpret, lacking intuitive visual clarity. Hence, SAR often collaborates with other modalities for cloud removal. Similarly, in DM-based models, compared to simply modeling cloud removal tasks as inpainting tasks, results with SAR as auxiliary input often exhibit higher credibility. 
Jing \textit{et al.}\cite{p-RS-cloudremoval-jing2023denoising} introduce an innovative approach in optical satellite images with DDPM Feature-Based Network for Cloud Removal (DDPM-CR). This model incorporates auxiliary SAR data and multilevel features from DDPM to recover missing information across various scales. 
% The cloud removal head, equipped with an attention mechanism, recovers missing information, and 
A cloud loss is proposed to balance information recovery in the cloud and no cloud regions. 
Zhao \textit{et al.}\cite{p-RS-cloudremoval-zhao2023cloud}
% , building upon the foundation of multi-modal cloud removal, 
propose CRRS that integrates multi-temporal sequence information into DMs
% in \cref{fig:cloudremoval}
, combining two mainstream cloud removal concepts in a single framework.

{Rui \textit{et al.}~\cite{rui2024unsupervised} propose the first unsupervised hyperspectral pansharpening method leveraging a pre-trained diffusion model. By projecting hyperspectral images into a low-dimensional subspace, the approach exploits their low-rank properties to learn distributions efficiently. This method addresses the complexities of merging low-resolution hyperspectral data with high-resolution panchromatic images, yielding superior quality and improved generalization compared to traditional Bayesian and deep learning methods.
Seo \textit{et al.}\cite{p-RS-imagetransition-seo2023improved}, employing a self-supervised denoiser in the latent space, train the Brownian-Bridge diffusion model to achieve SAR to Electro-Optical image translation tasks, thereby achieving visual-fidelity performance.}
% and downstream-friendly results. 
% Experiments conducted around flood datasets verify enhanced visual information and downstream segmentation task performance for the translated images.
\begin{table*}[htbp]
  \centering
\setlength{\abovecaptionskip}{0cm}
  \caption{{Datasets for low-level vision. In the column of scales, we present detailed separation information if the dataset is separated as the training and testing sets. Due to space constraints, only three representative datasets are listed. For a comprehensive collection, please refer to our \href{https://github.com/ChunmingHe/awesome-diffusion-models-in-low-level-vision}{\color{blue}repository}. Clicking on the dataset will redirect you to its download link.
  }}
  \resizebox{\linewidth}{!}{
    \begin{tabular}{l|c|c|c|c|l}
    \toprule
    Tasks  & Datasets & Scales & Sources & Modalities  & \multicolumn{1}{c}{Remarks} \bigstrut\\
    \hline
        \multirow{3}[6]{*}{SR} 
%         &\href{https://www2.eecs.berkeley.edu/Research/Projects/CS/vision/bsds/}{\textit{BSD500}}~\cite{D-BSD500} & 500   & TPAMI 2010 & Syn   & A synthetic benchmark that is initially designed for object contour detection. \bigstrut\\
% \cline{2-6}          
% & \href{https://sites.google.com/site/romanzeyde/research-interests}{\textit{Set14}}~\cite{D-set14} & 14    & TPAMI 2015 & Syn    & Commonly utilized for testing performance of super-resolution algorithms. \bigstrut\\

%\cline{2-6}         
% &\href{http://www.manga109.org/en/}{\textit{Manga109}}~\cite{d-Manga109} & 109   & MTAP 2015 & Syn    & Compiled mainly for academic research on Japanese manga media processing. \bigstrut\\
% \cline{2-6}          & \href{https://www.kaggle.com/datasets/msahebi/super-resolution}{\textit{General100}}~\cite{D-General100} & 100   & ECCV 2016 & Syn   & Synthesized images in uncompressed BMP format covering various scales.
% \bigstrut\\
%\cline{2-6}         
& \href{https://data.vision.ee.ethz.ch/cvl/ntire17//}{\textit{DIV2K}}~\cite{D-DIV2K}  & 900/100 & NTIRE 2018 & Syn   & A commonly-used dataset with diverse scenarios and realistic degradations. \bigstrut\\
% \cline{2-6}          & \href{https://github.com/YingqianWang/Flickr1024}{\textit{Flickr1024}}~\cite{wang2019flickr1024} & 1024  & ICCVW 2019 & Syn   &  A large-scale stereo image dataset with high-quality pairs and diverse scenarios. \bigstrut\\
\cline{2-6}          & \href{https://github.com/jbhuang0604/SelfExSR}{\textit{Urban100}}~\cite{d-Urban100} & 100   & CVPR 2019 & Syn   & Sourced from urban environments: city streets, buildings, and urban landscapes. \bigstrut\\
\cline{2-6}          & \href{https://github.com/xiezw5/Component-Divide-and-Conquer-for-Real-World-Image-Super-Resolution}{\textit{DRealSR}}~\cite{d-DRealSR} & 31970  & ECCV 2020& Real   &  Benchmarks captured by DSLR cameras, circumventing simulated degradation.
\bigstrut\\
% \cline{2-6}         
% & \href{https://github.com/xiezw5/Component-Divide-and-Conquer-for-Real-World-Image-Super-Resolution}{\textit{DRealSR}}~\cite{d-DRealSR} & 31970 & ECCV 2020 & Real  & Benchmarks captured by DSLR cameras, circumventing simulated degradation. \bigstrut\\
    \hline
     \multirow{3}[6]{*}{Deblur} & \href{https://seungjunnah.github.io/Datasets/gopro}{\textit{GoPro}}~\cite{d-gopro} & 2103/1111  & CVPR 2017 & Syn   & Acquired by high-speed cameras for video quality assessment and restoration.\bigstrut\\
\cline{2-6}          & \href{https://github.com/joanshen0508/HA\_deblur}{\textit{HIDE}}~\cite{HIDE}  & 8422  & ICCV 2019 & Syn  & Cover long-distance and short-distance scenarios degraded by motion blur. \bigstrut\\
\cline{2-6}          
% & \href{https://seungjunnah.github.io/Datasets/reds}{\textit{REDS}}~\cite{REDS}  & 270/30 & NTIRE 2019 & Real  & Contain 300 video sequences with dynamic duration and varied resolutions.\bigstrut\\
% \cline{2-6}         
% & \href{https://seungjunnah.github.io/Datasets/reds}{\textit{BSD}}~\cite{D-BSD}   & 80/20 & ECCV 2020 & Real  & Comprise more scenes and use the proposed beam-splitter acquisition system. \bigstrut\\
% \cline{2-6}       
& \href{https://github.com/rimchang/RealBlur}{\textit{RealBlur}}~\cite{d-RealBlur-J} & 3758/980 & ECCV 2020 & Real  & Cover common instances of motion blur, captured in raw and JPEG formats. \bigstrut\\
    \hline
    \multirow{3}[6]{*}{Dehaze}
%     & \href{https://data.vision.ee.ethz.ch/cvl/ntire18//i-haze/}{\textit{I-Haze}}~\cite{D-I-HAZE} & 35    & NTIRE 2018 & Real   & Indoor dataset with real haze for objective image dehazing and evaluation. \bigstrut\\
% \cline{2-6}          & \href{https://data.vision.ee.ethz.ch/cvl/ntire18//o-haze/}{\textit{O-Haze}}~\cite{D-O-haze} & 45    & NTIRE 2018 & Real   & Outdoor dataset with real haze for objective image dehazing and evaluation. \bigstrut\\
% \cline{2-6}          & \href{https://data.vision.ee.ethz.ch/cvl/ntire19//dense-haze/}{\textit{Dense-Haze}}~\cite{D-DenseHaze} & 33    & ICIP 2019 & Real  & Real-world dataset with dense haze for robust single image dehazing methods. \bigstrut\\
% \cline{2-6}          
& \href{https://github.com/Boyiliee/RESIDE-dataset-link}{\textit{RESIDE}}~\cite{D-RESIDE} & 13000/990 & TIP 2019 & Syn+Real  & Divided into five subsets to highlight diverse sources and heterogeneous contents.\bigstrut\\
\cline{2-6}          & \href{https://data.vision.ee.ethz.ch/cvl/ntire20/nh-haze/}{\textit{NH-Haze}}~\cite{D-NH-Haze} & 55    & CVRPW 2020 & Real   & The first non-homogeneous dehazing dataset with realistic haze distribution. \bigstrut\\
\cline{2-6}          & \href{https://github.com/liuye123321/DMT-Net}{\textit{Haze-4K}}~\cite{D-Haze-4k} & 4000  & MM 2021 & Syn    & A large-scale synthetic dataset for image dehazing with varing distributions. \bigstrut\\
    \hline
    \multirow{3}[6]{*}{Derain} 
    &\href{https://www.icst.pku.edu.cn/struct/Projects/joint\_rain\_removal.html}{\textit{Rain100H}}~\cite{D-Rain100}  & 1800/100 & CVPR 2017 & Syn  &  Comprise synthetic datasets with five types of rain streaks for rain removal. \bigstrut\\
\cline{2-6}          & \href{https://github.com/rui1996/DeRaindrop}{\textit{RainDrop}}~\cite{d-raindrop} & 861/239 & CVPR 2018 & Syn & \textcolor[rgb]{ .051,  .051,  .051}{Image pairs with raindrop degradation, captured using the setup of dual glasses.} \bigstrut\\
\cline{2-6}         
% & \href{https://github.com/stevewongv/SPANet}{\textit{SPA-Data}}~\cite{D-SPA-data} & 638492/1000 & CVPR 2019 & Real & Design a semi-automatic method to generate clean images from real rain streaks. \bigstrut\\
% \cline{2-6}          & \href{https://github.com/panda-lab/Single-Image-Deraining}{\textit{MPID}}~\cite{D-MPID}  & 3961/419 & CVPR 2019 & Syn+Real &A large-scale benchmark that focuses on driving and surveillance scenarios. \bigstrut\\
% \cline{2-6}          & \href{https://github.com/xw-hu/DAF-Net}{\textit{RainCityscapes}}~\cite{D-RainCityscapes} & 9432/1188 & CVPR 2019 & Syn  &A famous rain removal dataset with paired depth maps for outdoor scenarios. \bigstrut\\
% \cline{2-6}          & \href{https://github.com/Songforrr/RainDS\_CCN}{\textit{RainDS}}~\cite{D-RainDS} & 3450/900 & CVPR 2021 & Syn+Real  & A hybrid dataset with both real and synthesized data under diverse scenarios.
% % and lighting conditions. 
% \bigstrut\\
% \cline{2-6}          & \href{https://github.com/Yueziyu/RainDirection-and-Real3000-Dataset}{\textit{RainDirection}}~\cite{D-RainDirection} & 2920/430 & ICCV 2021 & Syn & A large-scale synthetic rainy dataset with directional labels in the training phase. \bigstrut\\
% \cline{2-6}       
& \href{https://github.com/UCLA-VMG/GT-RAIN}{\textit{GT-RAIN}}~\cite{D-GTrain} & 28217/2100 & ECCV 2022 & Real  &  The first paired deraining dataset with real data by controlling non-rain variations. \bigstrut\\
\hline
    \multirow{3}[6]{*}{LLIE}
%     & \href{https://data.csail.mit.edu/graphics/fivek/}{\textit{MIT-Fivek}}~\cite{D-MIT-FiveK} & 4500/500 & CVPR 2011 & Syn    & A curated dataset of RAW photos adjusted by skilled retouchers for visual appeal. \bigstrut\\
%     \cline{2-6}          & \href{https://github.com/cs-chan/Exclusively-Dark-Image-Dataset}{\textit{LOLv1}}~\cite{D-LOL} & 485/15 & BMVC 2018 & Real  & The first dataset with image pairs from real scenarios for low-light enhancement.  \bigstrut\\
% \cline{2-6}         
% & \href{https://github.com/cchen156/Learning-to-See-in-the-Dark}{\textit{SID}}~\cite{D-SID}   & 5094  & CVPR 2018 & Real  & A dataset of raw short-exposure images with their long-exposure reference images.  \bigstrut\\
& \href{https://daooshee.github.io/BMVC2018website/}{\textit{LOLv1}}~\cite{D-LOL}   & 485/15  & BMVC 2018 & Real  & The first dataset with image pairs from real scenarios for low-light enhancement.  \bigstrut\\
\cline{2-6}         
% & \href{https://github.com/csjcai/SICE}{\textit{SICE}}~\cite{D-SICE}  & 589& TIP 2018 & Syn    & A large-scale multi-exposure image dataset with complex illumination conditions. \bigstrut\\
% \cline{2-6}          & \href{https://github.com/cs-chan/Exclusively-Dark-Image-Dataset}{\textit{ExDark}}~\cite{D-Exdark} & 7363 & CVIU 2019 & Real  & Collected in low-light scenarios with 12 classes and instance-level annotations.\bigstrut\\
% \cline{2-6}         
& \href{https://github.com/flyywh/SGM-Low-Light}{\textit{LOLv2-Real}}~\cite{D-lolv2} & 689/100 & TIP 2021 & Real  & A three-step shooting strategy is used to eliminate intra-pair image misalignments.  \bigstrut\\
\cline{2-6}          & \href{https://github.com/flyywh/SGM-Low-Light}{\textit{LOLv2-Syn}}~\cite{D-lolv2} & 900/100 & TIP 2021 & Syn    & Synthetic dark images mimic real low-light photography via histogram analysis. \bigstrut\\
\hline
    \multirow{3}[6]{*}{IVF}
 & \href{https://github.com/hanna-xu/RoadScene}{\textit{RoadScene}}~\cite{xu2020u2fusion-Roadscene} & 221 & TPAMI 2020 & Real    & Aligned Vis-IR image pairs from diverse road scenes with noise-removed IR images. \bigstrut\\       
 \cline{2-6} 
& \href{https://github.com/Linfeng-Tang/MSRS}{\textit{MSRS}}~\cite{D-MSRS-tang2022piafusion}   & 1444  & Inf. Fusion 2022 & Real  & High-quality dataset optimized for contrast and noise in day and night road scenarios.  \bigstrut\\
\cline{2-6}               
& \href{https://github.com/dlut-dimt/TarDAL}{\textit{M3FD}}~\cite{D-M3FD-liu2022target} & 4177 & CVPR 2022 & Real  & A dataset of aligned pairs, featuring various environments, illumination conditions.  \bigstrut\\

\hline
    \multirow{3}[6]{*}{
    \parbox{1.25cm}
    {
    % Accelerated MRI\\ reconstruction
    MRI Data \\ Processing
    }}
    % & \href{https://brain-development.org/ixi-dataset/}{\textit{IXI}} & 600 & --- & Real  & A public MRI dataset from healthy subjects with diverse modalities.  \bigstrut\\
    % \cline{2-6}   
    % & \href{https://www.med.upenn.edu/cbica/brats/}
    % {\textit{BraTS}}~\cite{menze2014multimodalBRATS} & --- & TMI 2014 & Real    & A MRI dataset for annually Brain Tumor Segmentation Challenge. \bigstrut\\
    % \cline{2-6}   
& \href{https://fastmri.med.nyu.edu/}{\textit{FastMRI}}~\cite{D-fastMRI}   & 8400  & arXiv 2018 & Real  & Raw data and DICOM images for knee and brain MRIs with diverse contrasts.  \bigstrut\\ \cline{2-6}  
& \href{https://github.com/StanfordMIMI/skm-tea/}
    {\textit{SKM-TEA}}~\cite{desai2skm} & 19200/5800 & NeurIPS 2021 & Real    & Raw data, DICOM images, and masks for double echo steady state MRI knee scans. \bigstrut\\
    \cline{2-6}   
& \href{https://github.com/microsof/fastmri-plus/}
    {\textit{FastMRI+}}~\cite{zhao2022fastmri+} & 8400 & Sci. Data 2022 & Real    & Add clinical pathology annotations for FastMRI, facilitating disease diagnosis. \bigstrut\\
%     \multirow{3}[6]{*}{Desnow} & \href{https://sites.google.com/view/yunfuliu/desnownet}{\textit{Snow100k}}~\cite{D-Snow100k} & 100000 & TIP 2018 & Syn+Real  & A large-scale dataset with over 1k real-world images degraded by heavy snow. \bigstrut\\
% \cline{2-6}          & \href{https://github.com/weitingchen83/JSTASR-DesnowNet-ECCV-2020}{\textit{SRRS}}~\cite{D-SRRS}  & 16000 & ECCV 2020 & Syn+Real  & A hybrid snow dataset with 15k synthesized images and 1k real-world images.
% % collected from the Internet. 
% \bigstrut\\
% \cline{2-6}          & \href{https://github.com/weitingchen83/ICCV2021-Single-Image-Desnowing-HDCWNet}{\textit{CSD}}~\cite{D-CSD}   & 10000 & ICCV 2021 & Syn  &A large-scale desnowing dataset to comprehensively simulate snow scenarios. \bigstrut\\
    % \hline
%      \multirow{5}[10]{*}{Shadow Removal} & \href{http://visual.cs.ucl.ac.uk/pubs/softshadows/}{\textit{LRSS}}~\cite{D-LRSS}  & 37    & TOG 2015 & Syn  & Dataset of real soft shadow test photographs. \bigstrut\\
% \cline{2-6}          & \href{https://www3.cs.stonybrook.edu/~minhhoai/projects/shadow.html}{\textit{SBU}}~\cite{D-SBU}   & 4727  & ECCV 2016 & Syn    & Dataset with diverse shadows, annotated with binary masks. \bigstrut\\
% \cline{2-6}          & \href{https://github.com/Liangqiong/DeShadowNet}{\textit{SRD}}~\cite{D-SRD}   & 2680/408 & CVPR 2017 & Syn    & The first large scale benchmark. \bigstrut\\
% \cline{2-6}          & \href{https://github.com/IsHYuhi/ST-CGAN\_Stacked\_Conditional\_Generative\_Adversarial\_Networks}{\textit{ISTD}}~\cite{D-ISTD}  & 1870  & CVPR 2018  & Syn  & For shadow understanding that consists of image triplets of shadow image, mask, and shadow-free image. \bigstrut\\
% \cline{2-6}          & \href{https://github.com/CXH-Research/DocShadow-SD7K}{\textit{SD7K}}~\cite{D-SD7K}  & 7620  & ICCV 2023 & Syn    & The only large-scale high-resolution dataset. \bigstrut\\
    \bottomrule
    \end{tabular}}
  \label{tab:dataset collection}%
  \vspace{-3mm}
\end{table*}%

 \subsection{Diffusion models for video processing}
The latest research endeavors aim to extend the exploration of DMs into higher-dimensional data, particularly in video tasks~\cite{sp-VDM-LDM-Blattmann,P-VDM1,P-VDM2,sp-VDM-LAVIE,sp-VDMzeroshotprior}. However, compared with image, video processing requires temporal consistency across video frames. Currently, the number of DM-based video models is relatively few, only applied in several fundamental tasks.
% , including video frame prediction~\cite{sp-VDMprediction-Yang}, interpolation~\cite{sp-VDM-interpolation-MCVD,sp-VDM-LDMVFInterpolation,sp-VDMinfilling}, super-resolution~\cite{p-chen2024learning-videoSR,p-yuan2024inflation-videoSR}, and restoration~\cite{p-yang2024genuine-videoIR}.

% \begin{figure}[t]
%   \centering
%       \setlength{\abovecaptionskip}{0.1cm}
%   \includegraphics[width=\linewidth]{images/pipelines/Yang2.png}
% \caption{The architecture of the two-stage hybrid models \cite{sp-VDMprediction-Yang}.}\vspace{-3mm}
% \label{fig:sp-VDMprediction-Yang}
% \end{figure}

\noindent\textbf{Video frame prediction and interpolation}. Renowned for remarkable generative capacities, DM-based models are especially suitable for video prediction and interpolation.
% , regarding temporal generation. 
Yang \textit{et al.}\cite{sp-VDMprediction-Yang} first use DMs in autoregressive video prediction. 
% Aiming to model residuals in video compression, 
The two-stage hybrid model %(see in \cref{fig:sp-VDMprediction-Yang}) 
initially utilizes RNNs to obtain deterministic predictions for the next frame, providing sequential priors for the DM. Then the DM focuses on modeling residuals, whose effect is verified with various metrics perceptually and probabilistically.

By employing different mask manners for time series, masked conditional DMs can be trained for prediction and interpolation. 
Höppe \textit{et al.} \cite{sp-VDMinfilling} introduce conditions through a randomized masking schedule, allowing the model to be trained conditionally with only slight modifications to the unconditionally trained models. 
Voleti \textit{et al.}\cite{sp-VDM-interpolation-MCVD} employ a similar masking concept but further propose a blockwise autoregressive conditioning procedure to facilitate coherent long-term generation. 
% They also incorporate temporal information into the U-Net architecture. 
In contrast to direct modifications of DDPM, Danier \textit{et al.}\cite{sp-VDM-LDMVFInterpolation} first use LDM in video frame interpolation. They design a vector-quantized autoencoding model for LDM, better recovering high-frequency details and achieving perceptual superiority.

\noindent\textbf{Video super-resolution}.
Early DM-based video works \cite{P-VDM1, P-VDM2} merely tailor the classical framework to meet data dimensionality of input-output sequences
% for temporal tasks 
and train the models from scratch, resulting in an undeniable computational burden. 
Given the tremendous success of DMs~\cite{T-LDM}, one approach is to leverage off-the-shelf pre-trained models and endow them with temporal modeling capacities by integrating temporal layers into the U-Net architecture. 
Inspired by~\cite{sp-VDM-LDM-Blattmann, sp-VDM-LAVIE, sp-VDM-interpolation-MCVD}, Yuan \textit{et al.}~\cite{p-yuan2024inflation-videoSR} propose an efficient DM for text-to-video super-resolution. By inflating text-to-image model weights into the video generation framework
% and incorporating 
with an attention-based temporal adapter,
% for coherence across frames, 
this method achieves high-quality and temporally consistent results.

Striving for Spatial Adaptation and Temporal Coherence (SATeCo), Chen \textit{et al.}~\cite{p-chen2024learning-videoSR} propose a novel video SR approach SATeCo,
% (see in \cref{fig:SATeCo}), 
which freezes pre-trained parameters and optimizes spatial feature adaptation (SFA) and temporal feature alignment (TFA) modules. 
% Specifically, SFA modulates the pixel-level high-resolution features for spatial adaptation through learning affine parameters guided by low-resolution videos. TFA conducts self-attention to enhance feature interaction and performs cross-attention between counterparts of different resolutions to guide temporal feature alignment learning. 
Experiments validate the effect of the modules in preserving spatial fidelity and enhancing temporal feature alignment.

\noindent\textbf{Video restoration}.
Limited DM-based algorithms focus on video restoration, showing a promising future direction.
Yang \textit{et al.} \cite{p-yang2024genuine-videoIR} propose a novel Diffusion Test-Time Adaptation (Diff-TTA) method for all-in-one adverse weather removal in videos. At the training stage, a novel temporal noise model is introduced to exploit frame-correlated information in degraded video clips. During inference, the authors first introduce test-time adaptation to DM-based methods by proposing a novel proxy task named Diffusion Tubelet Self-Calibration (Diff-TSC). This allows the model to adapt in real-time without modifying the training process and achieve restoration under unseen weather conditions.

\section{EXPERIMENTS}\label{chap:experiment}
% In this section, we initially collect the prevalent datasets and evaluation metrics used in low-level vision. Then select four significant tasks to conduct comprehensive comparisons of those diffusion model-based methods. 

% To ensure an effective and comprehensive comparison among different diffusion model-based methods for the aforementioned low-level vision tasks in four major fields, we initially collated the prevalent datasets, evaluation metrics, and experimental configurations for each specific task. Subsequently, under as fair a model parameter setting as possible, we conducted a comprehensive comparison between several typical low-level vision tasks' DM-based methods, existing benchmark methods and state-of-the-art methods based on other models such as CNN-based, Transformer-based, GAN-based, and more. This comparison delved into algorithmic performance across evaluation metrics, model scale, and computational complexity. Due to space constraints, the experimental section and its analysis are exclusively focused on the natural Image aspect presented in \cref{chap:DM-naturalIR}.

\subsection{Datasets}

\noindent \textbf{{Large-scale datasets for model pre-training}}{.
% Although pre-trained diffusion models may pose a risk of potential data leakage\cite{leakage1,leakage2,tang2023consistency},
% Constructing DMs from scratch has a rigorous training process, requiring great computation capacities and high-memory hardware that are beyond the reach of many research labs~\cite{R-SOTA-DM2023}. Hence, 
% there is a preference for 
% Using pre-trained models is a common practice in generative modeling tasks \cite{tang2023consistency}. 
Several large-scale datasets, \textit{e.g.}, \textit{ImageNet} \cite{D-imagenet} and \textit{CelebA} \cite{D-CelebA}, are commonly used for generative model pre-training \cite{tang2023consistency,tang2024mind}. 
\textit{ImageNet} \cite{D-imagenet} is a large-scale dataset with over 14 million natural images spanning over 21k classes, termed \textit{ImageNet21K}. \textit{ImageNet1k}, serving as a subset of \textit{ImageNet21K}, has 1k classes with about 1k images per class. Besides, \textit{CelebA} has 200k facial images, each annotated with 40 attributes, where \textit{CelebA-HQ}\cite{CelebA-HQ} is a subset having 30k high-resolution facial images.
Please see our \href{https://github.com/ChunmingHe/awesome-diffusion-models-in-low-level-vision}{\color{blue}repository} for more datasets.}

\noindent \textbf{{Low-level vision datasets for model training}}{.
Various datasets are tailored to 
% address diverse low-level vision tasks, aiming to 
accommodate various degradation modes. For space limitations, we summarize commonly used datasets for several classical low-level vision tasks in \cref{tab:dataset collection}.
% , including their scales, sources, modalities, and remarks. 
Please refer to our \href{https://github.com/ChunmingHe/awesome-diffusion-models-in-low-level-vision}{\color{blue}repository} for more information.
% about datasets in different scenarios, please refer to our \href{https://github.com/ChunmingHe/awesome-diffusion-models-in-low-level-vision}{\color{blue}repository}. 
In practice, DM-based models are typically pre-trained on large-scale datasets to learn general features and structures, before being fine-tuned on specific low-level vision datasets to address the specific degradation issues.}

\subsection{Evaluation metrics}

\noindent\textbf{{Distortion-based metrics}}{.
Several commonly used metrics are introduced here.
% , \textit{i.e.}, Peak Signal-to-Noise Ratio (PSNR), Structural Similarity (SSIM). 
Peak Signal-to-Noise Ratio (PSNR) quantifies the pixel-wise disparity between a corrupted image and its clean image by computing their mean squared error, while Structural Similarity (SSIM) 
assesses the likeness between distorted and clean images across three aspects, including contrast, brightness, and structure. Mutual Information (MI)~\cite{qu2002information} and Qabf~\cite{xydeas2000objective} are two important fusion metrics, where MI evaluates the amount of information transferred from source images to the fused image and Qabf focuses on the preservation of edge information.
}
% \noindent\textbf{Distortion-based metrics}.
% \begin{itemize}[label=\textcolor{black}{$\bullet$}]
%   \item {\bfseries PSNR} \cite{he2023reti-LLIE3} (Peak Signal to Noise Ratio) quantifies the pixel-wise disparity between a corrupted image and its clean image by computing their mean squared error.
%   \item {\bfseries SSIM} (Structural Similarity \cite{SSIM}), aiming to accommodate human visual perception, assesses the likeness between distorted and clean images across three aspects, including contrast, brightness, and structure. 
% \end{itemize}

\noindent\textbf{{Inception-based metrics}}{. Learned Perceptual Image Patch Similarity (LPIPS) \cite{LPIPS} and Fréchet inception distance (FID) \cite{FID} are two representative metrics. LPISP uses the pre-trained AlexNet as a feature extractor and adjusts linear layers to emulate human perception. Besides, FID assesses the fidelity and diversity of generated images by computing the Fréchet distance of their reference images. }
\begin{table*}[tp]
\begin{minipage}[c]{0.5\textwidth}
\centering
\setlength{\abovecaptionskip}{0cm}
\renewcommand\arraystretch{1.5}
\caption{{Results of DM-based 4$\times$ SR methods.}}
\resizebox{1\textwidth}{!}{
\begin{tabular}{l|ccc|ccc|c|c}
\toprule
\multicolumn{1}{l|}{\multirow{2}{*}{Methods}} & \multicolumn{3}{c|}{\textit{DIV2K}~\cite{D-DIV2K}} & \multicolumn{3}{c|}{\textit{Urban100}~\cite{d-Urban100}} & \multicolumn{1}{c|}{Time} & \multicolumn{1}{c}{Parameters}  \\  
& \cellcolor{gray!40}PSNR$\uparrow$ & \cellcolor{gray!40}SSIM$\uparrow$ & \cellcolor{gray!40}LPIPS$\downarrow$ & \cellcolor{gray!40}PSNR$\uparrow$ & \cellcolor{gray!40}SSIM$\uparrow$ & \cellcolor{gray!40}LPIPS$\downarrow$ &\cellcolor{gray!40}[s/image] &\cellcolor{gray!40}[M]  \\ 
\midrule
Bicubic & 25.36 & 0.643 & 0.31 & 24.26 & 0.628 & 0.34 & - & - \\
IR-SDE~\cite{sp-IRSDE} & 25.90 & 0.657 & 0.23 & 26.63 & 0.786 & 0.18 & 63.9 & 137.2  \\ 
CDPMSR~\cite{sp-cdpmsr} & 27.43 & 0.712 & 0.19 & 26.98 & 0.801 & 0.16 & - & -  \\
IDM~\cite{sp-idm} & 27.13 & 0.703 & {{0.18}} & 26.76 & 0.657 & {{0.13}} & 59.5 & 116.6 \\
DiffIR~\cite{sp-diffir} & \textbf{29.13} & \textbf{0.730} & \textbf{0.09} & 26.05 & 0.776 & \textbf{0.10} & 0.3 & 26.5  \\
ResDiff~\cite{sp-resdiff} & 27.94 & 0.723 & 0.23 & \textbf{27.43} & \textbf{0.824} & 0.14 & 51.79 & 98.91 \\
\bottomrule
\end{tabular}}
\label{tab:sm-sr}
\end{minipage}
\begin{minipage}[c]{0.5\textwidth}
\centering
\setlength{\abovecaptionskip}{0cm}
\renewcommand\arraystretch{1.5}
\caption{{Results of DM-based motion deblurring methods.}}
\resizebox{1\textwidth}{!}{

\begin{tabular}{l|ccc|ccc|c|c}
\toprule
 \multicolumn{1}{l|}{\multirow{2}{*}{Methods}}& \multicolumn{3}{c|}{\textit{Gopro}~\cite{d-gopro}}& \multicolumn{3}{c|}{\textit{HIDE}~\cite{HIDE}}& \multicolumn{1}{c|}{Time} & \multicolumn{1}{c}{Parameters} \\
& \cellcolor{gray!40}PSNR$\uparrow$ & \cellcolor{gray!40}SSIM$\uparrow$ & \cellcolor{gray!40}LPIPS$\downarrow$ & \cellcolor{gray!40}PSNR$\uparrow$ & \cellcolor{gray!40}SSIM$\uparrow$ & \cellcolor{gray!40}LPIPS$\downarrow$ &\cellcolor{gray!40}[s/image] &\cellcolor{gray!40}[M]  \\ 
\midrule
Blurred image& 25.64& 0.793& 0.29& 23.95& 0.763& 0.33& - & - \\
DvSR~\cite{sp-whang2022deblurring}& 33.23 & 0.963 & 0.08 & 30.07 & 0.928 & 0.09 & - & - \\
IR-SDE~\cite{sp-IRSDE}&30.70&0.901&\textbf{0.06}&28.34 &0.914 &0.10 & 4.3 & 137.2 \\ 
MSGD~\cite{ren2023multiscale} & 31.19 & 0.943 & \textbf{0.06} & 29.14 & 0.910 & \textbf{0.09} & - & - \\
DiffIR~\cite{sp-diffir} &33.20 &0.963 &0.08 &\textbf{31.55} &\textbf{0.947} &0.10 & 0.436 & 26.91  \\
{HI-Diff~\cite{sp-chen2023hierarchical}}  & {\textbf{33.33}}  &{ \textbf{0.964}}   & {0.08}& {31.46}  & {0.945}  & {0.11} & 2.280 & 28.5 \\
\bottomrule
\end{tabular}}
\label{tab:deblur-quantitative}%\vspace{1mm}
\end{minipage} 
\vspace{-3mm}
\end{table*}

\begin{table*}[tp]
\begin{minipage}[c]{0.405\textwidth}
\centering
\setlength{\abovecaptionskip}{0cm}
\renewcommand\arraystretch{1.5}
\caption{{Results of zero-shot DM-based inpainting methods using the same pre-trained model with 552.8M parameters (LPIPS $\downarrow$).}}
\resizebox{1\textwidth}{!}{\begin{tabular}{l|ccc|ccc|c}
\toprule
\multicolumn{1}{l|}{\multirow{2}{*}{Methods}}& \multicolumn{3}{c|}{\textit{ImageNet 1K}~\cite{D-imagenet}}& \multicolumn{3}{c|}{\textit{CelebA-HQ}~\cite{CelebA-HQ}}& \multicolumn{1}{c}{Time}  \\
& \cellcolor{gray!40}Half & \cellcolor{gray!40}Narrow & \cellcolor{gray!40}Wide & \cellcolor{gray!40}Half & \cellcolor{gray!40}Narrow  & \cellcolor{gray!40}Wide    &\cellcolor{gray!40}[s/image] \\ \hline
Masked image &0.502 &0.347 &0.297 &0.474 &0.389 &0.279& -  \\
RePaint~\cite{sp-repaint} & 0.323 &\textbf{0.072} &0.156  &0.199 &0.039 &0.072& 176.7    \\
DDRM~\cite{sp-DDRM}  &0.385 &0.211 &0.231 &0.273 &0.140 &0.125& 4.9  \\  
DDNM~\cite{sp-ddnm}  &0.408 &0.101 &0.185 &0.257 &0.071 &0.111& 8.2   \\
CoPaint~\cite{sp-copaint}  &0.307 &0.078 &0.138 &0.188 &0.040 &0.071& 146.9  \\
Tiramisu~\cite{liu2023imageinpaintingtractablesteering}  &\textbf{0.286} &0.079 &\textbf{0.125} &\textbf{0.189} &\textbf{0.033} &\textbf{0.069} & 186.5  \\
 \bottomrule
\end{tabular}}
\label{tab:inpainting-quantitative}
\end{minipage}
% \vspace{-1mm}
\begin{minipage}[c]{0.595\textwidth}
\centering
\setlength{\abovecaptionskip}{0cm}
\renewcommand\arraystretch{1.5}
\caption{{Results of DM-based low-light enhancement methods (*: using the gt mean strategy, $^\dagger$: a multi-modal method, $\rightarrow$: cross-dataset transfer learning tests from \textit{LOLv2-Real (v2R)}, \textit{LOLv2-Syn (v2S)} to \textit{LOLv1 (v1)}.).}}
\resizebox{1\textwidth}{!}{
\begin{tabular}{l|cc|cc|cc|cc|c|c}
\toprule
\multicolumn{1}{l|}{\multirow{2}{*}{Methods}} & \multicolumn{2}{c|}{\textit{v2R}~\cite{D-lolv2}}
& \multicolumn{2}{c|}{\textit{v2S}~\cite{D-lolv2}}
&\multicolumn{2}{c|}{\textit{v2R $\rightarrow$ v1}~\cite{D-LOL}}
& \multicolumn{2}{c|}{\textit{v2S $\rightarrow$ v1}~\cite{D-LOL}}
& \multicolumn{1}{c|}{Time}
& \multicolumn{1}{c}{Parameters} \\  
   & \cellcolor{gray!40}PSNR$\uparrow$                 &\cellcolor{gray!40}LPIPS$\downarrow$               
   &\cellcolor{gray!40}PSNR$\uparrow$         &\cellcolor{gray!40}LPIPS$\downarrow$ 
   & \cellcolor{gray!40}PSNR$\uparrow$        
   & \cellcolor{gray!40}LPIPS$\downarrow$  
   & \cellcolor{gray!40}PSNR$\uparrow$        
   & \cellcolor{gray!40}LPIPS$\downarrow$  &\cellcolor{gray!40}[s/image] 
   &\cellcolor{gray!40}[M]             \\ \midrule
Low-Light Image & 9.71& 0.52& 11.22& 0.38& 7.77 & 0.56& 7.77 & 0.56& - & - \\
PyDiff*\cite{zhou2023pyramid-LLIE2}& 24.01& 0.23& 19.60& 0.22 & 24.25 & 0.14 & 18.13 & 0.34& 0.28 & 97.9 \\
Diff-Retinex\cite{p-llie-yi2023diff} & 20.17 &0.10  &24.30   &0.06& 18.83 &0.13& 16.66 & 0.39& 0.24 & 56.9 \\
GSAD*\cite{p-llie-hou2024global}& 28.82& 0.09& \textbf{28.67}& 0.04& 27.29 & 0.09& 20.48 & 0.36& 0.43 & 17.2 \\
Reti-Diff\cite{he2023reti-LLIE3}& 22.97& \textbf{0.08}& 27.53& \textbf{0.03}& 20.25 & 0.11& 17.84 & 0.34& 0.08 & 26.1 \\
CFWD$^\dagger$\cite{p-llie-xue2024low}& \textbf{29.86}& 0.19& 24.42  & 0.10& - & - & - & -& 0.81 & 22.1 \\
 \bottomrule
\end{tabular}}
\label{tab:lowlight-quantitative}
\end{minipage} \\
\vspace{1mm}
\begin{minipage}[c]{0.5\textwidth}
\centering
\setlength{\abovecaptionskip}{0cm}
\renewcommand\arraystretch{1.5}
\caption{{Results of DM-based infrared and visible image fusion methods.}}
\resizebox{1\textwidth}{!}{\begin{tabular}{l|ccc|ccc|c|c}
\toprule
 \multicolumn{1}{l|}{\multirow{2}{*}{Methods}}& \multicolumn{3}{c|}{\textit{MSRS}~\cite{D-MSRS-tang2022piafusion}}& \multicolumn{3}{c|}{\textit{M3FD}~\cite{D-M3FD-liu2022target}}& \multicolumn{1}{c|}{Time} & \multicolumn{1}{c}{Parameters} \\
& \cellcolor{gray!40}MI$\uparrow$                         & \cellcolor{gray!40}Qabf$\uparrow$                    & \cellcolor{gray!40}SSIM$\uparrow$                      & \cellcolor{gray!40}MI$\uparrow$                         & \cellcolor{gray!40}Qabf$\uparrow$                     & \cellcolor{gray!40}SSIM$\uparrow$        &\cellcolor{gray!40}[s/image] &\cellcolor{gray!40}[M]             \\ \hline
DDFM~\cite{p-if-zhao2023ddfm}  &2.35 & 0.58 & \textbf{0.94} & 2.52 & 0.49 & \textbf{0.95} & 37.28 & 552.8 \\
Dif-Fusion~\cite{p-if-yue2023dif}  &3.34 &0.58 &0.81 &2.96 &0.58 &0.85 & 0.3763 & 416.5 \\  
GLAD~\cite{guo2024glad}  &3.34 &0.63 &0.92 &3.24 &\textbf{0.63} &0.86& - & - \\
Diff-IF~\cite{yi2024diff-if} &3.45 &\textbf{0.69} & 0.88 &3.19 &0.59 &0.90& 0.4588 & 23.7 \\
LFDT-Fusion~\cite{yang2025lfdt} &\textbf{3.58} &0.68 & 0.90 &\textbf{3.27}  &\textbf{0.63} & 0.93 & 0.5469 & 21.3 \\
                                     \bottomrule
\end{tabular}}\vspace{-5mm}
\label{tab:imagefusion-quantitative}
\end{minipage}
\begin{minipage}[c]{0.5\textwidth}
\centering
\setlength{\abovecaptionskip}{0cm}
\renewcommand\arraystretch{1.28}
\caption{{Results of DM-based accelerated MRI reconstruction methods (single coil).}}
\resizebox{1\textwidth}{!}{
\begin{tabular}{l|cc|cc|c|c}
\toprule
\multicolumn{1}{l|}{\multirow{2}{*}{Methods}}& \multicolumn{2}{c|}{\textit{FastMRI}~\cite{D-fastMRI} R=4x}
&\multicolumn{2}{c|}{\textit{FastMRI}~\cite{D-fastMRI} R=8x}& \multicolumn{1}{c|}{Time} & \multicolumn{1}{c}{Parameters}\\  
   & \cellcolor{gray!40}PSNR$\uparrow$                         & \cellcolor{gray!40}SSIM$\uparrow$                         & \cellcolor{gray!40}PSNR$\uparrow$                         & \cellcolor{gray!40}SSIM$\uparrow$              &\cellcolor{gray!40}[s/image] &\cellcolor{gray!40}[M]          \\ \midrule
Score-MRI\cite{chung2022scoreMRI} & 31.1   & 0.76  &28.4  &0.77& 40.8628 & 61.43 \\
DiffuseRecon\cite{DiffuseRecon-peng2022towards}& 31.7& 0.71 & 29.9& 0.61& 8.654 & 149.59 \\
SMRD\cite{SMRD-ozturkler2023smrd}& 36.5& 0.89 & 32.4& 0.80& 463.4216 & 94.13 \\
SSDiffRecon\cite{SSDiffRecon-korkmaz2023self}&40.1 &\textbf{0.97}& 35.1& 0.93& 278.61 & 78.58 \\
AdaDiff\cite{AdaDiff-gungor2023adaptive}& \textbf{40.2}& 0.96 &\textbf{37.2} &\textbf{0.94}& 10.2157 & 39.72 \\
\bottomrule
\end{tabular}}\vspace{-5mm}
\label{tab:accelerated MRI reconstruction-quantitative}
\end{minipage}
\end{table*}

\noindent\textbf{{Human-centric evaluations}}{.
Human-centric evaluation is a subjective assessment method, where participants select the image verifying the most effective performance from a set of images. For fairness, anonymizing the method and randomizing the order 
% within each set 
is essential. Human assessment scores are calculated using the Mean Opinion Score (MOS) derived from a pool of participants. A higher MOS indicates superior perceptual quality as perceived by humans. }
% However, evaluating via MOS can be costly, and the results may be biased due to subjective perceptual differences. Besides, the time-consuming nature of the procedure makes it suitable for small-scale assessments, such as user studies, but challenging to employ for evaluation during training and broader comparisons.

\noindent\textbf{Downstream application-based evaluations}.
% \subsubsection{Downstream application-based evaluations}
Apart from improving visual quality, generating those enhanced images that can facilitate high-level vision tasks, such as image segmentation\cite{sp-LACT-liu2023dolce, xiao2024survey}, is also a significant object. Hence, the evaluation of various methods extends to examining the impact 
% of low-level vision methods 
on real-world vision-based applications.

% \subsubsection{Model efficiency}
% Apart from addressing how to comprehensively evaluate algorithm performance more objectively, DM-based models face constraints in metric selection due to computational burdens. Vanilla DMs, for instance, require hundreds or even thousands of iterations of denoising. Consequently, employing metrics like FID, which necessitate substantial sample collections to assess the proximity between restored results and ground truth distributions, brings about significant computational expenses and time consumption, posing challenges for the evaluation process. Such challenges have been overlooked in previous review papers. Recent research efforts have largely concentrated on efficient design strategies\cite{R-efficient} to tackle these drawbacks. Nonetheless, the computational demands of DMs persist and have become a primary hindrance to their widespread adoption.

\subsection{Experimental results}

\begin{figure*}[ht]
    {\includegraphics[width=\textwidth]{images/Main\_Image/visualization.pdf}}\vspace{-1mm}
\caption{{Qualitative comparisons for DM-based methods on six commonly investigated tasks.}} \vspace{-4mm}
\label{fig:Result}
\end{figure*}

% To demonstrate the superiority of different diffusion models, 
% We provide quantitative comparisons for DM-based methods on six commonly investigated tasks. 
% dcommonly investigated 
% low-level vision 

% , including super-resolution, motion deblurring, and low-light image enhancement. 
% The evaluation metrics are composed of PSNR, SSIM\cite{SSIM}, and LPIPS\cite{LPIPS}. The qualitative results of some diffusion models are shown in \cref{tab:sm-sr}, \cref{tab:deblur-quantitative}, \cref{tab:lowlight-quantitative}.
{The runtime of all algorithms was measured at a resolution of $256 \times 256$ using an RTX 4090 GPU. For methods that are not publicly available, their cells are marked with ``-''.}

\noindent\textbf{{Results on super-resolution}}{. The results for DM-based models on 4$\times$ image SR, tested on \textit{DIV2k}~\cite{D-DIV2K} and \textit{Urban100} \cite{d-Urban100}, are listed in \cref{tab:sm-sr}. We find that IDM~\cite{sp-idm} and DiffIR~\cite{sp-diffir} perform well on LPIPS. They leverage preprocessed features as conditional input, enhancing perceptual quality. Resdiff~\cite{sp-resdiff} performs well on PSNR and SSIM. This is because Resdiff focuses on residual information, ensuring salient consistency.
% of the restored image with the high-resolution image. 
Visualization is presented in~\cref{fig:Result}.}
% at the pixel level.

\noindent\textbf{{Results on deblurring}}{.
We evaluate five DM-based methods on the motion deblurring task using the \textit{Gopro}~\cite{d-gopro} and \textit{HIDE}~\cite{HIDE} datasets.
As shown in \cref{tab:deblur-quantitative}, DiffEvent~\cite{p-wang2024diffevent} and HI-Diff~\cite{sp-chen2023hierarchical} achieve competitive performance on PSNRs and SSIMs. DiffEvent is enabled to achieve both low-light recovery and image deblurring by introducing a learnable decomposer.
% and input reconstruction task, combined with Retinex theory and contrast regularization. 
% HI-Diff achieves good generalization performance in complex fuzzy scenarios by using LDM to generate a highly-compressed prior.
% features in a highly compressed latent space
% , while fusing a prior from multiple scales using a hierarchical integration module. 
In contrast, MSGD~\cite{ren2023multiscale} introduces a multi-scale structural bootstrap to better sample from the target condition distribution, hence the best performance on perceptual metrics. The qualitative analysis is presented in~\cref{fig:Result}.}

\noindent\textbf{{Results on zero-shot inpainting}}{. As shown in \cref{tab:inpainting-quantitative,fig:Result}, the experimental results
% on both the \textit{ImageNet 1K} and \textit{CelebA-HQ} datasets 
demonstrate that Tiramisu \cite{liu2023imageinpaintingtractablesteering} consistently outperforms others in most scenarios, particularly excelling in cases with large masks.
% , achieving the best LPIPS scores. 
This is because Tiramisu uses TPMs to constrain the generation process of natural images.
In contrast, the Repaint \cite{sp-repaint} stands out in narrower regions 
% by conditioning the generation process 
by sampling from the given pixels during the reverse iterations. }
% By addressing the incoherence problem without violating the inpainting constraints and introducing a new solution to address the challenges in computing and sampling from the posterior distribution, Copaint \cite{sp-copaint} demonstrated the balanced results in both datasets with different types of masks.

\noindent\textbf{{Results on low-light image enhancement}}{. {Basic experiments are conducted on \textit{LOLv2-Real (v2R)}\cite{D-lolv2} and \textit{LOLv2-Syn (v2S)}\cite{D-lolv2}, with the results presented in \cref{tab:lowlight-quantitative,fig:Result}. GSAD \cite{p-llie-hou2024global} shows superior performance in PSNR, while 
Reti-Diff \cite{he2023reti-LLIE3} achieves competitive performance in LPIPS \cite{LPIPS}. CFWD \cite{p-llie-xue2024low} first introduces multi-modal into diffusion-based low-light enhancement, reaching the best real-world performance. To explore how datasets, such as synthetic versus real-world data, shape performance trends, we conduct further cross-dataset transfer tests. Considering that the ultimate goal of low-level vision methods is practical application under real-world degradation, we tested models trained from the real-world dataset (\textit{v2R}) and the synthetic dataset (\textit{v2S}) on the real-world dataset \textit{LOLv1 (v1)}~\cite{D-LOL} respectively. Evidently, models trained on real-world data consistently outperform those trained on synthetic data in practical scenarios. Noting that GSAD \cite{p-llie-hou2024global} and PyDiff \cite{zhou2023pyramid-LLIE2} employ the ``gt mean'' strategy, which involves fine-tuning the brightness of the generated results using the ground truth, thus producing much more impressive results than others in PSNR.}
}

\noindent\textbf{{Results on infrared and visible image fusion}}{. The results are reported in \cref{tab:imagefusion-quantitative,fig:Result}.
DDFM~\cite{p-if-zhao2023ddfm} designs a likelihood rectification module and achieves impressive SSIM, indicating strong structural fidelity. 
Diff-IF~\cite{yi2024diff-if} stands out with a strong Qabf~\cite{xydeas2000objective}, hinting at its effect enhancing image quality. 
LFDT-Fusion~\cite{yang2025lfdt}, combining LDM and transformer, achieves the highest MI~\cite{qu2002information} on \textit{MSRS} and gets competitive scores in Qabf and SSIM on \textit{M3FD}.}

\noindent\textbf{{Results on accelerated MRI reconstruction}}. As presented in~\cref{tab:accelerated MRI reconstruction-quantitative,fig:Result}, AdaDiff \cite{AdaDiff-gungor2023adaptive} achieves the best overall performance, particularly in the R=8x scenario. SSDiffRecon \cite{SSDiffRecon-korkmaz2023self} combines a conditional DM with data-consistency projections, showing strong performance, particularly in R=4x, where it closes to AdaDiff \cite{AdaDiff-gungor2023adaptive} in both PSNR and SSIM. The visualizations presented in~\cref{fig:Result} further confirm that both methods generate high-quality reconstruction results.

\noindent {\textbf{Discussion of model scalability}. The analysis indicates that computational costs and parameter counts are not necessarily correlated with model performance. Notably, IR-SDE~\cite{sp-IRSDE}, a supervised method, achieves outstanding results in both super-resolution and motion deblurring tasks, demonstrating exceptional multi-task scalability. This observation suggests that integrating an optimal amount of learnable parameters can enhance a model's adaptability to complex real-world degradations, thereby improving its scalability. Furthermore, these findings provide valuable insights for addressing the limitations of current zero-shot methods, which, despite their strong scalability, remain confined to linear degradation scenarios.
% The above analysis reveals that computational costs and parameter counts are not directly correlated with model performance. Moreover, IR-SDE~\cite{sp-IRSDE}, a supervised method, demonstrates impressive performance in both SR and motion deblurring, showcasing its multi-task scalability. This suggests that incorporating an appropriate amount of learnable parameters could enable better adaptation to complex real-world degradations, thus enhancing scalability. This provides insights into addressing the drawbacks of existing zero-shot methods, which excel in scalability while limited to linear degradation scenarios.
}

\section{Future directions}\label{chap:futurework}
% Like other deep generative models such as VAE\cite{P-VAE} and GANs\cite{P-GAN}, 
% DMs showcase superiority by generating a more accurate target distribution without needing large latent code hierarchies like VAEs, and without encountering optimization instability or mode collapse as seen in GAN-based methods. 

% Compared to other generative models, DMs exhibit the capability to generate high-fidelity images with complex details, rendering DMs widely applied in low-level vision tasks. However, considerable room for advancement remains in both DMs and low-level vision tasks.

\subsection{Mitigating the limitations of DMs}
Due to the high computational overhead, DMs encounter barriers to be applied in low-level vision tasks. Two viable ways are listed and discussed to mitigate this challenge.
% include reducing sampling steps and compressing model consumption.

% Reducing sample steps and compressing model consumption are two reasonable ways to address this problem.

\noindent \textbf{Reducing sample steps}.
% 改成 reducing sampling steps
Various efforts, extending beyond low-level vision, have been undertaken to enhance the sampling efficiency of DM:
(1) Modeling the diffusion process with a non-Markov Chain, such as DDIM\cite{T-DDIM}.
(2) Designing efficient ODE solvers~\cite{weinan2017proposal}.
(3) Using knowledge distillation to reduce sampling steps\cite{p-distillation2-meng2023distillation}.
(4) Performing DMs on compressed latent spaces~\cite{T-LDM}.
(5) Introducing cross-modality priors with conditional mechanisms\cite{p-crossmodal-accelerate1-liu2023improved,p-crossmodal-accelerate2-abu2022adir
% ,p-crossmodal-accelerate3-wang2023exploiting
}.
{(6) Rethinking diffusion process modeling with more efficient latent variable transitions (e.g., residual-based methods in Resshift~\cite{yue2024resshift}) and optimized noise design~\cite{shi2023resfusion}.}

% With these efforts, the number of sampling steps has been notably reduced to 10-20 steps, ensuring faster reconstruction. 
{These efforts reduce sampling steps to 10-20, with some studies, \textit{e.g.}, SinSR~\cite{wang2024sinsr}, even getting results in a single step, ensuring faster reconstruction. DDRM~\cite{sp-DDRM} achieves an inference time reduction to 5 seconds for a single $256 \times 256$ image by using the sampling strategy of DDIM~\cite{T-DDIM}.} Besides, 
% acknowledging the theoretical impracticality of the diffusion process starting from pure noise, 
some studies\cite{chung2022come,sp-zhao2023partdiff}
% , as shown in \cref{fig:revise_assumption}, revise assumptions and 
initialize networks by sampling from low-quality images or one-step reconstruction results of baseline networks,
% This approach capitalizes on the structural and textual information contained in low-quality images for enhanced restoration. Similarly, residual-space diffusion models focus on the disparity between low- and high-quality images
streamlining the learning target. 
% Furthermore, Zhao \textit{et al.}\cite{sp-zhao2023partdiff} verify the indistinguishability of intermediate latent states between low- and high-resolution images in the super-resolution task. Leveraging this insight, they employ an approximate substitution approach, reducing the need for numerous denoising iterations and significantly expediting both training and inference processes.
However, despite notable progress, the overall computational cost remains high, particularly for high-resolution images, presenting a substantial gap from real-time applications. 
% Addressing this challenge remains a longstanding and crucial direction for accelerating diffusion models.

% \begin{figure}[t]
%     \centering
%     \includegraphics[width=\linewidth]
%     {images/revise_assumption.pdf}
%     \caption{Plot of average error $\bar{\varepsilon }$ vs. time $t$. (a) vanilla reverse process. (b) different approaches in \cite{chung2022come} that revise the assumptions.}
% \label{fig:revise_assumption}
% \end{figure}

\begin{figure}[t]
  % \subfloat[Image Super-resolution]{\includegraphics[width=0.3\textwidth]{images/inpaint.jpg}}
  \subfloat[ Shift in pareto-frontier\cite{chung2023direct}.]{
    {\includegraphics[width=0.25\textwidth]{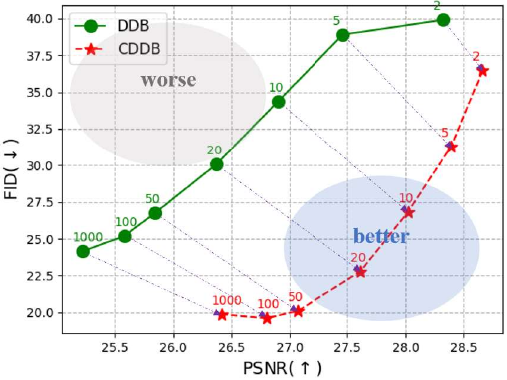}}}
    \label{fig:tradeoff}
    \subfloat[ Bi-level optimization\cite{he2023hqg}.]{
    {\includegraphics[width=0.223\textwidth]{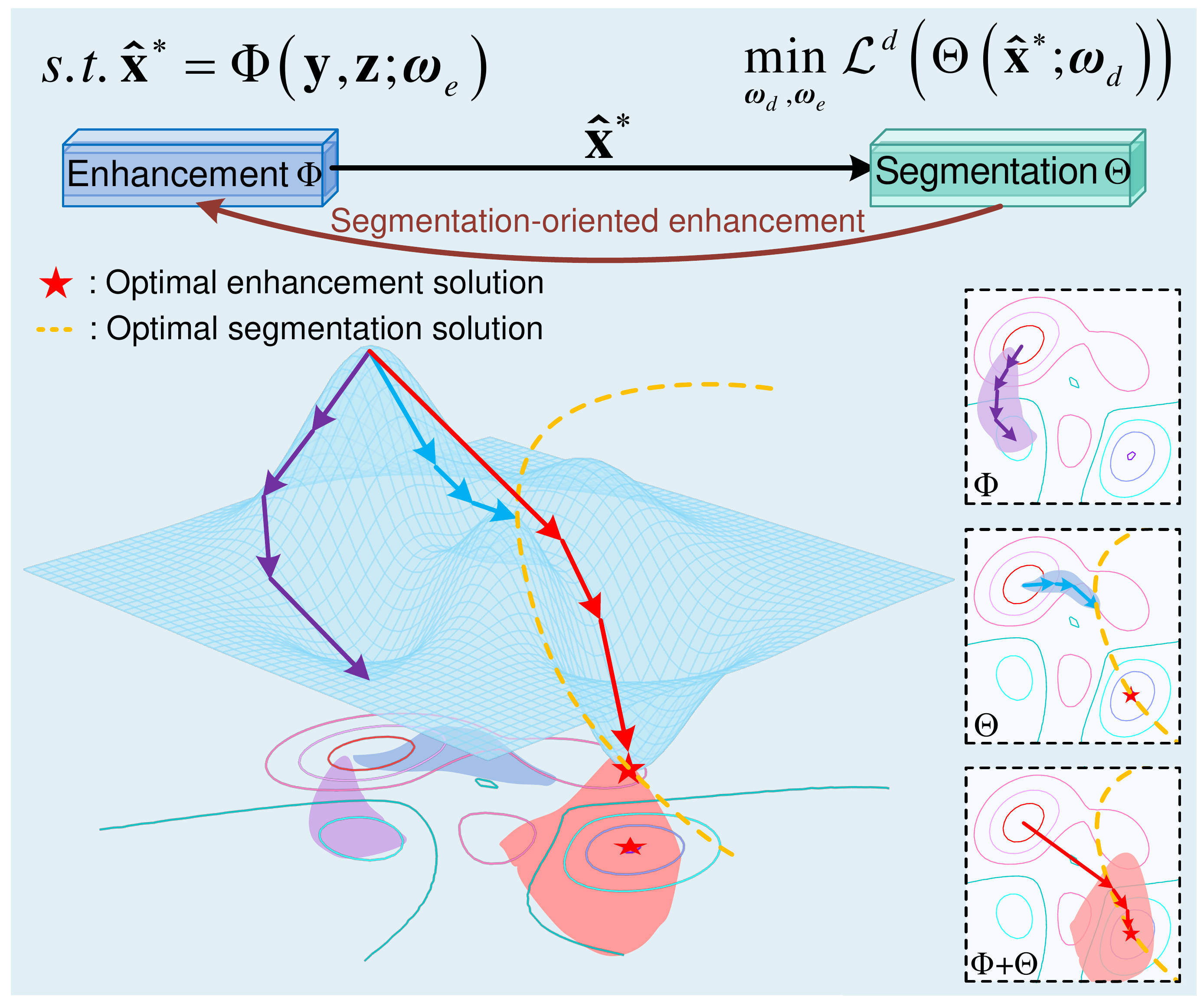}}}
    \label{fig:BLO}
  \vspace{-1mm}
\caption{Two strategies to amalgamate the strengths of DMs with the traits of low-level vision in \cref{Sec:future2}.} \vspace{-3mm}
\label{fig:Sec6.2}
\end{figure}
\noindent \textbf{Compressing model consumption}. 
The deployment of DM-based models in low-resource environments, such as edge devices, faces challenges due to their immense parameter size and computational complexity. Apart from employing fewer-step inference, researchers can explore architectural optimizations to address this issue, including model quantization, pruning, and knowledge distillation. 
Zhang \textit{et al.}~\cite{zhang2024laptop} combine automated layer pruning with normalized feature distillation to compress models. Castells \textit{et al.}~\cite{castells2024edgefusion} propose EdgeFusion, an optimized model for deploying SDMs on Neural Processing Units, which leverages advanced distillation techniques and model-level tiling to facilitate rapid inference. However, current methods primarily focus on generation tasks.
% popular tasks namely T2I models especially SDMs. 
In the future, these techniques are expected to be extended to low-level vision tasks, leveraging specific properties of each task for model compression.

\subsection{Amalgamating the strengths of DMs with the traits of low-level vision}\label{Sec:future2}
The greatest trait of low-level vision lies in the diversity of evaluation criteria, including visual fidelity, content invariance, and downstream task-based evaluations. DM-based methods, generating visual fidelity results, also should ensure the content invariance of the original one and the generated result and facilitate downstream tasks.
% Many perspectives can be used to evaluate the performance of low-level vision, including visual perception.

\noindent\textbf{Perception-distortion trade-off}. DM-based methods generate visually appealing results and excel in inception-based metrics, such as LPIPS\cite{LPIPS} and FID\cite{FID}. However, their high diversity often leads to challenges in maintaining content consistency, resulting in suboptimal performance in those distortion-based metrics such as PSNR and SSIM.

One potential solution involves designing hybrid models that integrate DMs with CNN-based or Transformer-based frameworks~\cite{he2023reti-LLIE3,sp-diffir}. These hybrid models have shown promising results, particularly in improving distortion-based metrics.
% One solution tries to design hybrid models that combine DMs with CNN-based or Transformer-based frameworks\cite{he2023reti-LLIE3,sp-diffir}, which achieve success in those distortion-based metrics. 
% Although there have been many attempts to achieve competitive performance on both types of metrics using hybrid models, 
% breakthrough progress has not yet been made. This raises the issue of the Perception-Distortion trade-off, where 
Besides, Pareto-frontiers are introduced as a comprehensive indicator to evaluate both perception and distortion and have proven the positive shift of the multiscale guidance mechanism  \cite{chung2023direct} that enhances coarse sharp image structures (in \cref{fig:Sec6.2} (a)). However, breakthrough progress has not yet been made and further explorations about novel mixed structure and new metrics are expected.
% , have been proven effective in achieving a positive shift in the Pareto-frontier ( \cref{fig:tradeoff}). 

% \begin{figure}[t]
%     \centering
%     \setlength{\abovecaptionskip}{0.1cm}
%     \includegraphics[width=0.7\linewidth]{images/Pareto/tradeoff.pdf}
%     \caption{Shift in pareto-frontier\cite{chung2023direct}.}
% \label{fig:tradeoff}
% \vspace{-3mm}
% \end{figure}

\noindent \textbf{Downstream task-friendly designs}.
Enabling reconstructed images to better serve downstream tasks is a continuous endeavor in low-level vision research\cite{sp-SR3,jin2024des3,tang2023source}. This pursuit manifests in three primary approaches with DMs. 

First, as shown in \cref{fig:Sec6.2} (b), several strategies~\cite{he2023hqg,liu2021investigating} adopt bi-level optimization to jointly optimize the networks of both the low-level vision task and the downstream task, such as image segmentation and object detection. By jointly optimizing the enhancement network with constraints from both itself and the downstream task, these methods aim to produce visually appealing results while enhancing downstream performance.
Besides, He \textit{et. al}~\cite{he2023hqg} propose feature-level information aggregation between low-level vision tasks and downstream tasks instead of the previous image-level manner, improving performance with deep constraints. Inspired by the adversarial attacks, which introduce slight perturbations to cause original methods to fail, Sun \textit{et. al}\cite{sun2022rethinking} propose adding slight noise to dehazed images. This strategy enhances downstream detection performance without altering the visual outcome.
However, these methods are often tailored to specific downstream tasks. There remains a need for a unified strategy, especially DM-based solutions that can generate visually friend results, to optimize generated images for a wide range of downstream tasks, which awaits further exploration.

\subsection{Tackling the inherent challenges of low-level vision}

Low-level vision tasks have several inherent challenges, including generalizability, data volume, and controllability.

\noindent\textbf{Real-world image restoration}. 
% Real-world image restoration is a challenging yet significant task that aims to address the unknown and complex degradations encountered in real-world scenarios. 
% Currently, most DM-based general-purpose image restoration methods utilize pre-trained models and exploit priors that are independent of the degradation process, achieving general-purpose image restoration performance on certain tasks. However, this approach requires the identification of the degradation mode and even simplifying the degradation process into a linear reverse problem, which is necessary for consistency constraints in diffusion models. 
% Therefore, most methods focus on synthetic distortions since identifying the distortion mode in the real world is difficult, and the complexity of the degradation mode often makes it challenging to estimate them as specific linear reverse problems.
% To achieve this,
% successful generalization and robustness of diffusion models in real-world/blind image restoration tasks, 
Two ways help DM-based methods to address real-world scenarios~\cite{fang2024real}, \textit{i.e.}, distortion invariant learning (DIL) and distortion estimation (DE).

% for DM-based models that have already achieved excellent performance on specific real-world image restoration tasks, 
DIL, renowned for its degradation-invariant representation and structural information preservation\cite{wang2024distortion}, can enhance DM-based methods by incorporating a distortion-invariant noise predictor and condition. This enables these methods to generalize effectively to diverse and even unknown degradations. 
Pioneering efforts have focused on redesigning the condition module to achieve distortion-invariant conditions, as demonstrated in works such as DifFace \cite{yue2022difface} and DR2 \cite{p-DR2wang2023dr2}. Notably, the effectiveness of such conditions also relies on DIL, warranting further research.

Moreover, DE techniques, extracting prior knowledge of degradation processes, are also urgently needed to extend the zero-shot diffusion models to real-world applications. Even though explicit results cannot be obtained, the powerful image synthesis capability of DMs can be utilized to convert synthetic datasets into real-world paired datasets, which will be discussed in detail in the following subsection.

\noindent \textbf{Data generation for data-hungry fields}. Data hungry is a prevalent challenge in low-level tasks, often stemming from limitations inherent in imaging devices and scenarios. 

While the unsupervised training is one avenue, many existing approaches \cite{he2023hqg} resort to data generation strategies to create pseudo image pairs. These pairs typically consist of generated degraded low-quality images paired with their corresponding original high-quality counterparts. This is a promising way for DM-based methods, although with limited explorations, for their powerful generation capacity. 
Moreover, certain extreme tasks suffer from severely limited data availability due to the difficulty or costliness of data acquisition, as seen in Photoacoustic data\cite{wang2012photoacoustic} and Cryo-electron microscopy data\cite{zeng2023high}. He \textit{et. al}~\cite{he2023strategic} propose leveraging existing data to generate more training data with GAN and thus enhance the generalizability of the method. This strategy aligns well with the DM-based methods, offering stable training conditions. Furthermore, controllable data generation, facilitated by user interaction, presents a promising approach to filtering out negative data that could otherwise affect stable performance.

% In specific, two directions worth DM-based methods to explore. The first direction is to generate pseudo image pairs, that is generated degraded low-quality images and those high-quality ones.

% Based on a deeper theoretical understanding of SDE, IR-SDE has achieved a certain level of generalization in complex degradation scenarios by establishing diffusion processes that simulate actual degradation processes. The underlying mathematical designs of targeted diffusion models for low-level vision tasks greatly enhance the algorithm's performance. However, collecting paired real-world distorted/clean image pairs is challenging, which hinders the widespread adoption of supervised learning-based algorithms.

% Limited-angle 2D to 3D reconstruction techniques are also applicable in cryo-electron microscopy reconstruction. Constrained by instrument limitations, these techniques enable the reconstruction of 3D voxel-level subtomogram samples from a specific angular range of 2D in situ electron microscopy images.

\noindent \textbf{Controllable and interactive low-level vision}.
Enhancing the controllability of low-level vision methods, enabling them to discern what and where users desire recovery, is of paramount importance. This focus has persisted over time, with efforts including the integration of human perception-related loss functions \cite{liang2023iterative} and interactive guidance priors \cite{he2023hqg,he2025run}. Recently, the utilization of vision prompts facilitated by Vision-Language models \cite{luo2023controlling} has provided a means for existing low-level vision methods to explicitly incorporate and interact with prompts within their networks, thereby achieving improved control and restoration effects \cite{li2023promptPIP
% ,lin2023improvingcontrol
% ,chen2023imagecontrol
}. Given that these vision prompts can act as interactive priors to curb the excessive diversity inherent in DM-based methods, leveraging Vision-Language models to develop controllable and interactive DM-based methods shows promise.

Moreover, future efforts should address real-world scenarios that involve multiple degradations. Zheng \textit{et al.} \cite{zheng2024selectiveDiffUIR} introduce a novel DM-based method named DiffUIR, employing a selective hourglass mapping technique. DiffUIR combines shared distribution mapping and robust conditional guidance based on Residual Denoising Diffusion Models~\cite{liu2023residualRDDM} to improve image restoration performance. Improving the internal mechanisms of deep learning to better learn the distribution of multi-task degradations represents a promising direction for future DM-based explorations.

\subsection{{Empowering low-level vision through multi-modal advances}}
% \noindent \textbf{Current multi-modal approaches for DM-based image generation.} 
{Multi-modal technology has advanced rapidly in image generation, revolutionizing the integration of images, text, and other relevant data. This section seeks to draw inspiration from advancements in generation to foster the development of low-level vision using multi-modal techniques.}
% , aiming to improve accuracy in image restoration and reconstruction.

% The rise of text-prompt CLIP has revolutionized the integration of text and image generation, facilitating the development of multi-modality-based strategies. 
% Models like Stable Diffusion can produce highly detailed images from diverse text prompts by leveraging latent space representations learned by CLIP to bridge the semantic gap between text and visual outputs. ControlNet~\cite{zhang2023adding} further enhances the precision of generated images by conditioning the diffusion process on additional multimodal inputs, such as sketches or depth maps, while still adhering to the semantic fidelity of text prompts. This section aims to 

\noindent \textbf{{Text prompt for low-level vision}}. Leveraging multi-modal condition control, recent low-level vision methods combine text-based inputs to harness the potential of CLIP in pre-trained DMs. This integration has led to notable performance improvements across various tasks~\cite{p-llie-xue2024low,sun2024coser}, enabling user-centered, customized image restoration~\cite{yu2024scalingsupir}, and even achieving all-in-one restoration~\cite{jiang2023autodir,ai2024multimodal}.

By using pre-trained DMs and multi-modal prompt engineering, these models demonstrate superiority over task-specific methods, showcasing robustness and adaptability in zero-shot settings. Ai \textit{et al.}~\cite{ai2024multimodal} introduce MPerceiver, the first multi-modal prompt framework that leverages Stable Diffusion's generative priors for all-in-one image restoration. MPerceiver employs a dual-branch architecture with a cross-modal adapter to convert CLIP image embeddings into degradation-aware text prompts. AutoDIR\cite{jiang2023autodir} leverages text prompts to enable customizable image restoration for multiple degradation types, using a CLIP model finetuned with semantic-agnostic constraints to detect dominant degradations and generate text prompts for DM-based image restoration, supplemented by user inputs. 
 
\noindent \textbf{{Extending to additional modalities beyond text}}{.
Multi-modal approaches extending beyond text and images show great potential for low-level vision tasks. Incorporating audio as an additional modality could further boost performance, particularly in video-related tasks where audio cues serve as valuable contextual information. The temporal and auditory alignment can provide insights into motion patterns or environmental conditions, aiding model understanding. Moreover, integrating audio could enable more fluid user interactions in real time, allowing for dynamic refinements during the restoration process. For example, models like Mini-Omni2~\cite{xie2024mini} illustrate the potential of combining audio, vision, and text within a unified framework, fostering more interactive and adaptive systems.}

\noindent {\textbf{Embodied Intelligence for low-level vision}. 
Recently, Embodied Intelligence~\cite{duan2022survey-embodied1} has gained significant traction, promoting the integration of multisensory methods into AI systems. This paradigm emphasizes interaction with the physical world through various sensory inputs, \textit{e.g.}, vision, touch, audio, and environmental data. It provides a foundation for low-level vision to incorporate diverse multi-modal information for improved performance
% and enhanced robustness
\cite{wang2024embodiedscan-embodied2}.}

{Leveraging multisensory inputs offers a transformative opportunity to tackle real-world challenges~\cite{gupta2021embodied}. For instance, humidity and temperature sensors can optimize dehazing methods by providing real-time environmental context. Tactile sensors, on the other hand, can enhance fine-grained texture restoration by using touch-based feedback to inform surface detail reconstruction in medical imaging and material analysis. Besides, integrating motion sensors, such as accelerometers and gyroscopes, can improve deblurring, strengthening robustness in dynamic environments.}

{The integration of these technologies within Embodied Intelligence suggests a future where low-level vision models become more adaptable, closely mimicking human sensory perception and interaction with the physical world.}

% and capable of perceiving and interacting with the world in ways that closely mimic human sensory experiences.

\section{Conclusions}\label{chap:conclusion}
This survey offers an extensive examination of diffusion models applied in low-level vision tasks, a gap overlooked in previous surveys. Our review covers both advances and practical implementations. Firstly, we identify and discuss various generic diffusion modeling frameworks. We then propose a detailed categorization of diffusion models used in low-level vision from multiple angles. Lastly, we highlight limitations of existing diffusion models and propose future research directions. Advances in low-level vision tasks using these models are emerging in more complex and higher-dimensional areas, including 3D objects, locomotion, and 4D scenes, highlighting the need for continued research.

\bibliographystyle{ieeetr}
\bibliography{reference}
\vspace{-1.6cm}
\begin{IEEEbiography}[{{\includegraphics[width=1in,height=1.25in,clip,keepaspectratio]{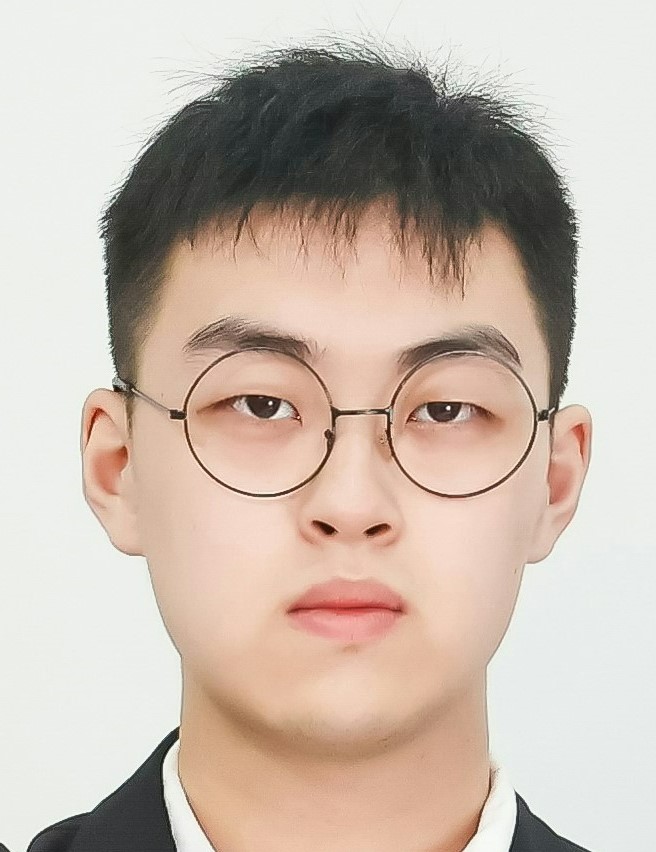}}}]{Chunming He} received the B.S. degree in communication engineering from Nanjing University of Posts and Telecommunications, Nanjing, China, in 2021, and the M.E. degree in computer science from Tsinghua University, Beijing, China, in 2024. He is currently a Ph.D. student with the Department of Biomedical Engineering, Duke University, Durham, USA. His research interests include computer vision, image processing, and biomedical image analysis.
\end{IEEEbiography}
\vspace{-1.6cm}
\begin{IEEEbiography}[{{\includegraphics[width=1in,height=1.25in,clip,keepaspectratio]{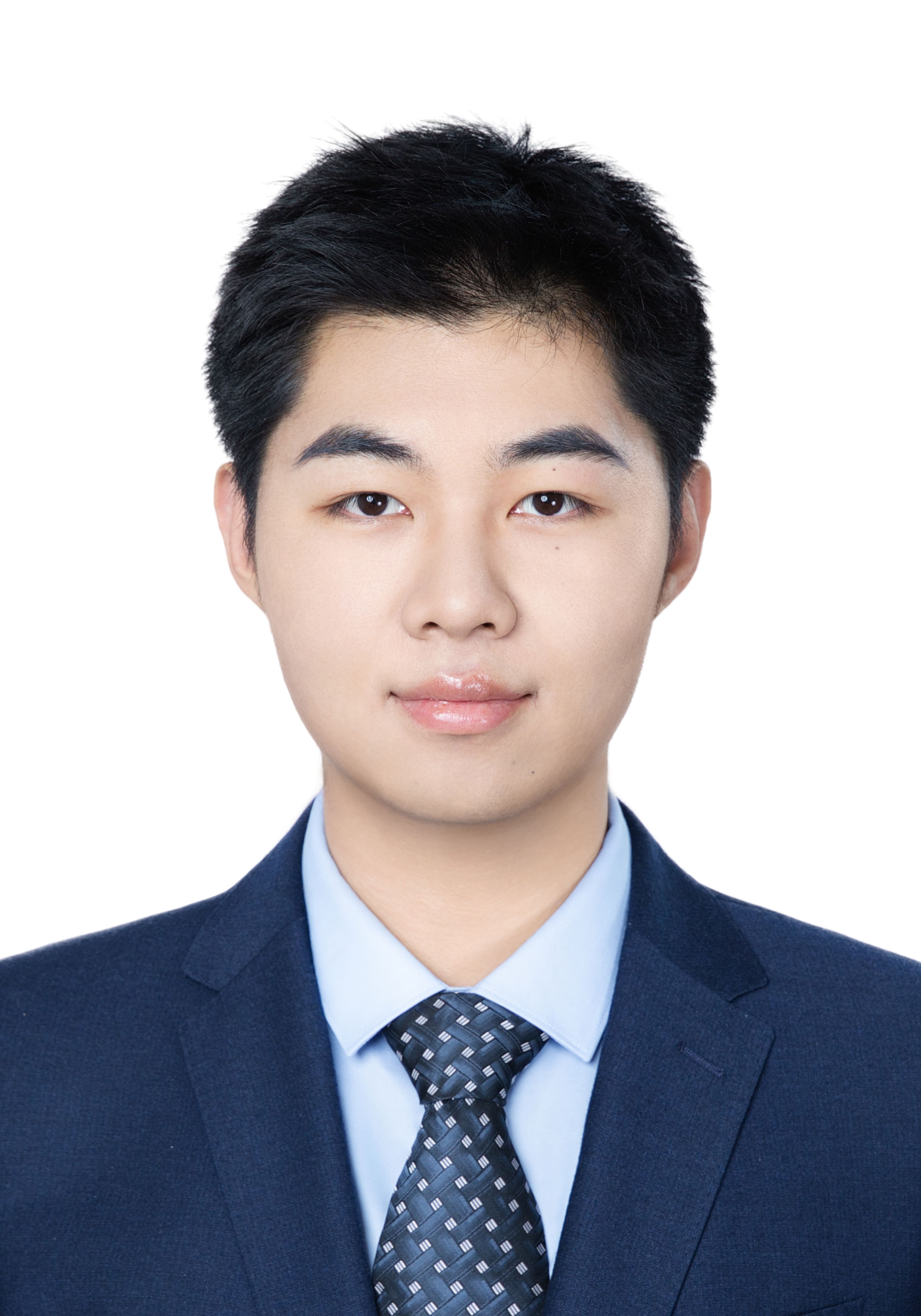}}}]{Yuqi Shen} received the B.S. degree in aircraft control and information engineering with the School of Astronautics, Beihang University, Beijing, China in 2024. Now, he is pursuing his M.S. degree in artificial intelligence, Tsinghua Shenzhen International Graduate School, Tsinghua University. His research interests include machine learning and computer vision.
\end{IEEEbiography}
\vspace{-1.6cm}
\begin{IEEEbiography}[{{\includegraphics[width=1in,height=1.25in,clip,keepaspectratio]{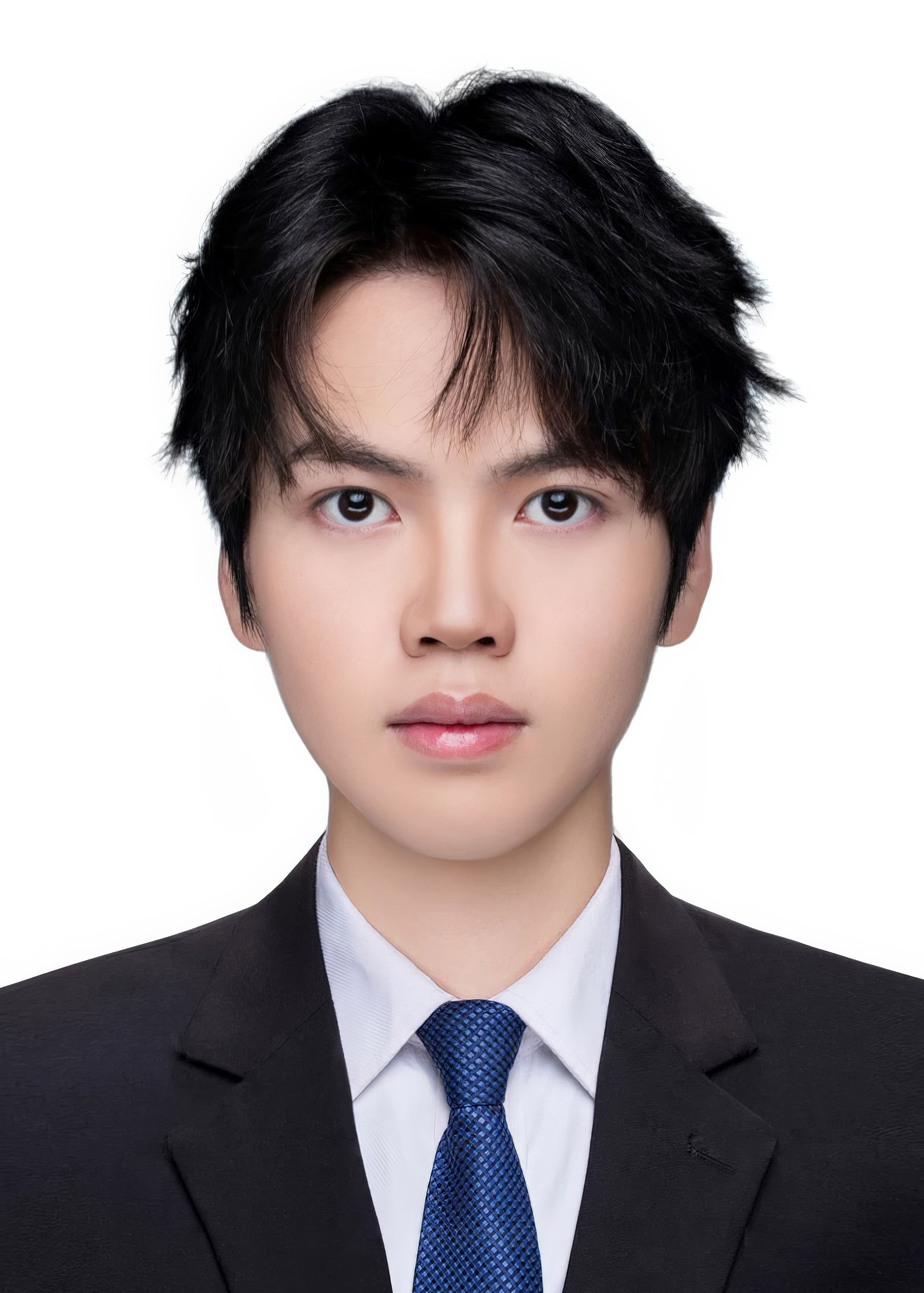}}}]{Chengyu Fang} received the B.S. degree in software engineering from Southwest University, Chongqing, China in 2024. Now, he is pursuing his M.S. degree in artificial intelligence, Tsinghua Shenzhen International Graduate School, Tsinghua University. His research interests include computer vision and image processing.
\end{IEEEbiography}
\vspace{-1.3cm}
\begin{IEEEbiography}[{{\includegraphics[width=1in,height=1.25in,clip,keepaspectratio]{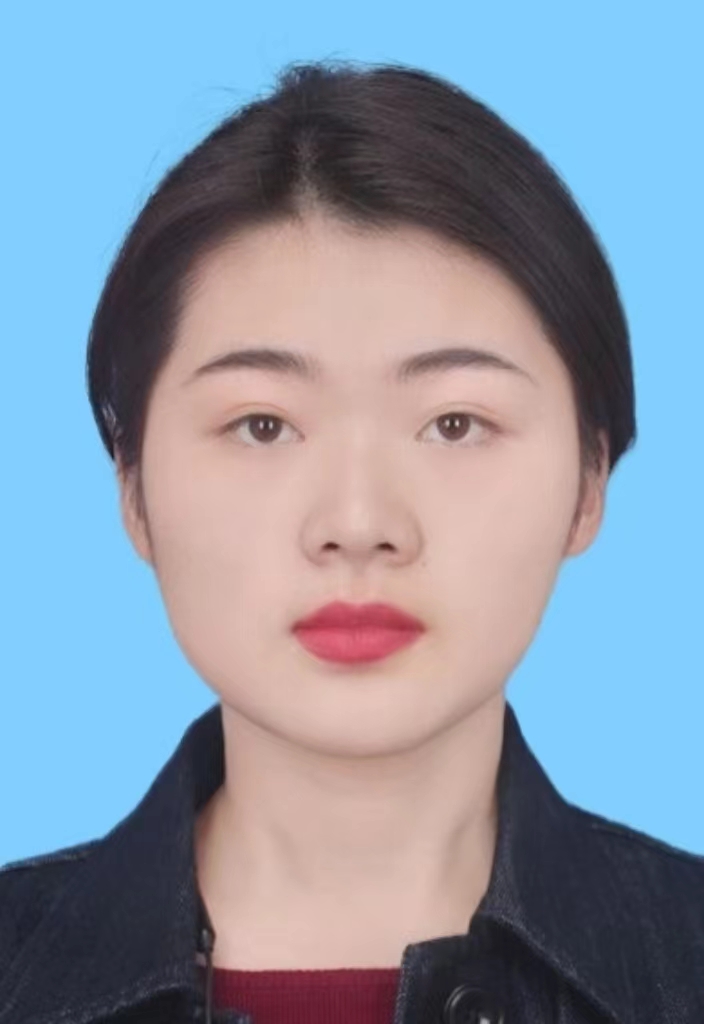}}}]{Fengyang Xiao} received the B.S. degree in information and computational science from Nanjing University of Posts and Telecommunicationa, Jiangsu Nanjing, China in 2021. Now, she is pursuing her M.S. degree in mathematics, School of Mathematics (Zhuhai), Sun Yat-sen University. She will be a Ph.D. student with the Department of Biomedical Engineering, Duke University, Durham, USA. Her research interests include differential equations and numerical solutions, image processing and computer vision.
\end{IEEEbiography}
\vspace{-1.3cm}
\begin{IEEEbiography}[{\resizebox{1in}{1.25in}{\includegraphics*{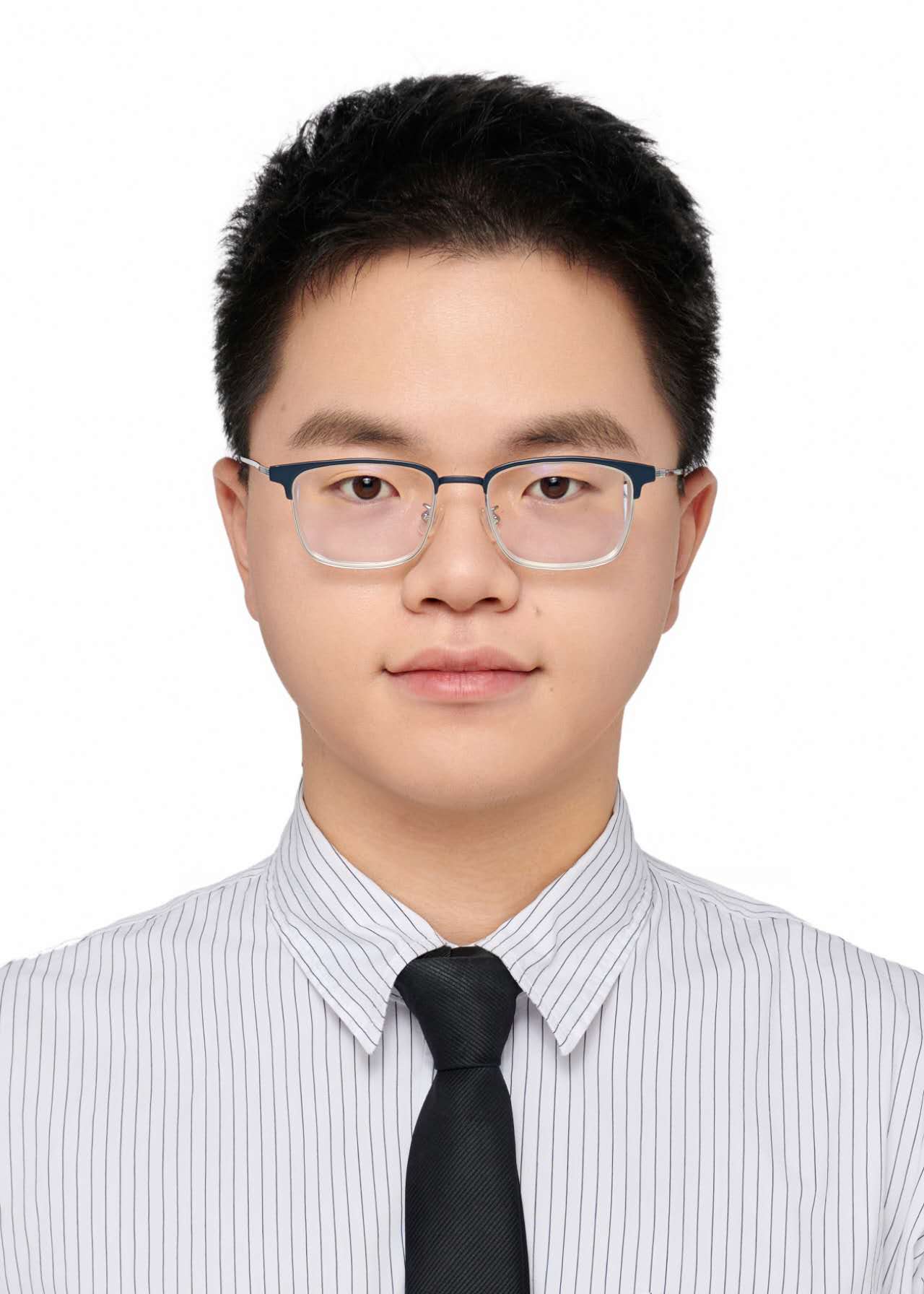}}}]{Longxiang Tang} is currently a master's student at Tsinghua Shenzhen International Graduate School, Tsinghua University. Before it, he received his B.S. degree in software engineering from University of Electronic Science and Technology of China. His research interests include multi-modal large language model and representation learning.       
\end{IEEEbiography}
\vspace{-1.3cm}
\begin{IEEEbiography}[{\includegraphics[width=1in,height=1.25in,clip,keepaspectratio]{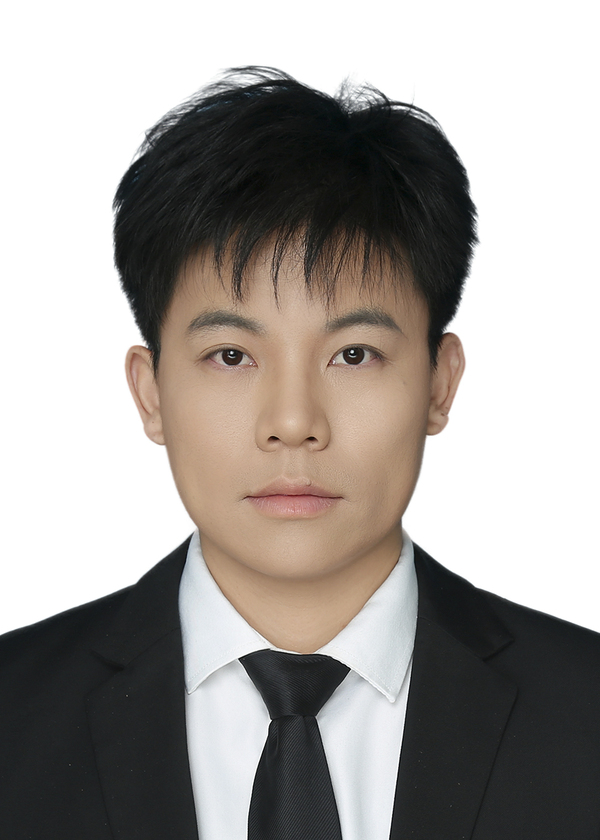}}]{Yulun Zhang} received a B.E. degree from the School of Electronic Engineering, Xidian University, China, in 2013, an M.E. degree from the Department of Automation, Tsinghua University, China, in 2017, and a Ph.D. degree from the Department of ECE, Northeastern University, USA, in 2021. He is an associate professor at Shanghai Jiao Tong University, Shanghai, China. He was a postdoctoral researcher at Computer Vision Lab, ETH Zürich, Switzerland. His research interests include image/video restoration and synthesis, biomedical image analysis, model compression, multimodal computing, large language model. He is/was an Area Chair for CVPR, ICCV, ECCV, NeurIPS, ICML, ICLR, IJCAI, ACM MM, and a Senior Program Committee (SPC) member for IJCAI and AAAI.
\end{IEEEbiography}
\vspace{-1.3cm}
\begin{IEEEbiography}[{\includegraphics[width=1in,height=1.25in,clip,keepaspectratio]{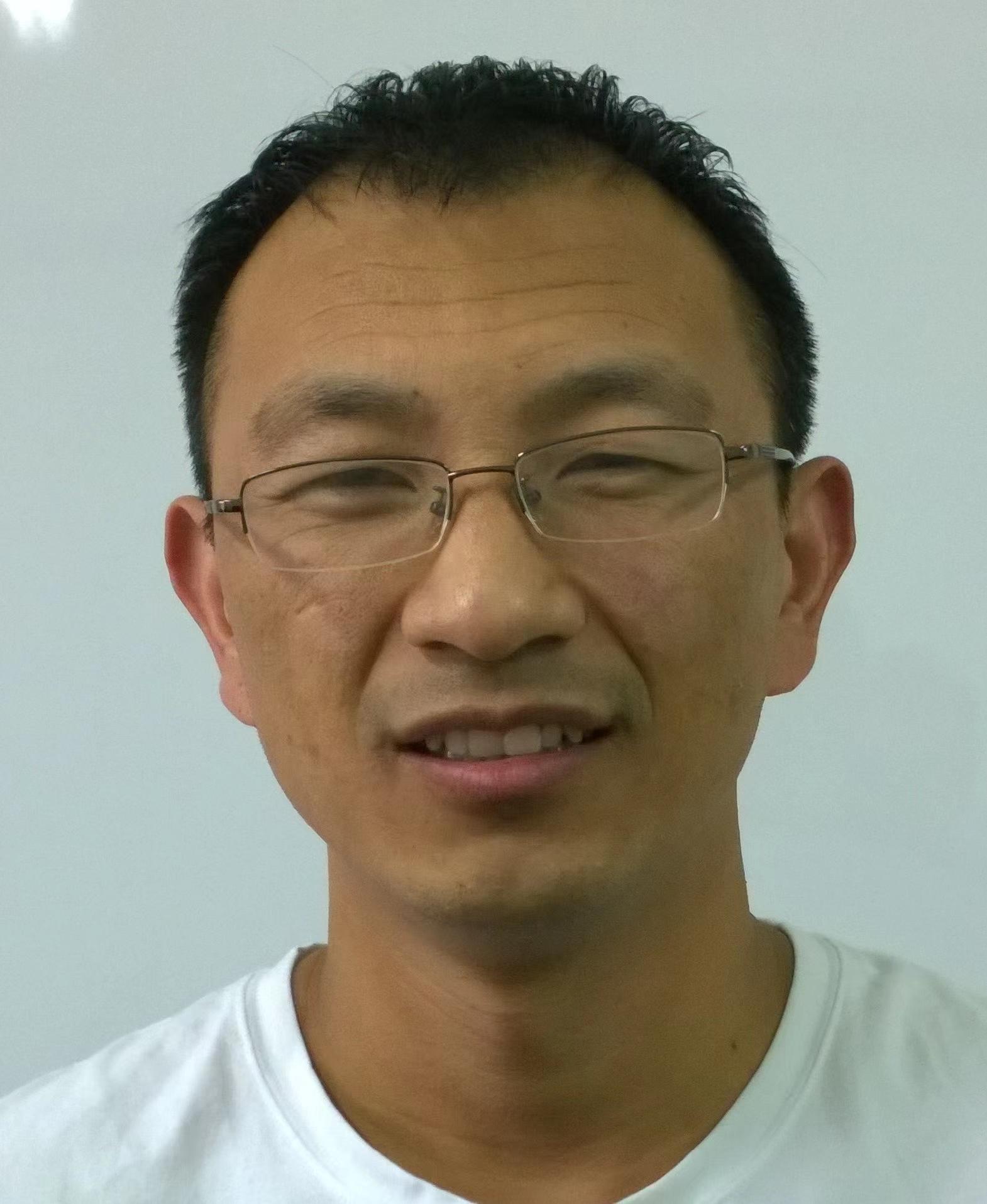}}]{Wangmeng Zuo} (M'09, SM'14) received the Ph.D. degree in computer application technology from the Harbin Institute of Technology, China, in 2007. He is currently a Professor with the School of Computer Science and Technology, Harbin Institute of Technology. He has published over 90 papers in top-tier academic journals and conferences. His current research interests include image enhancement and restoration, image generation and editing, visual tracking, object detection, and image classification. He has served as a Tutorial Organizer in ECCV 2016, an Associate Editor of the IET Biometrics, The Visual Computers, Journal of Electronic Imaging, and the Guest Editor of Neurocomputing, Pattern Recognition, IEEE Transactions on Circuits and Systems for Video Technology, and IEEE Transactions on Neural Networks and Learning Systems.
\end{IEEEbiography}
\vspace{-1.3cm}
\begin{IEEEbiography}[{\includegraphics[width=1in,height=1.25in,clip,keepaspectratio]{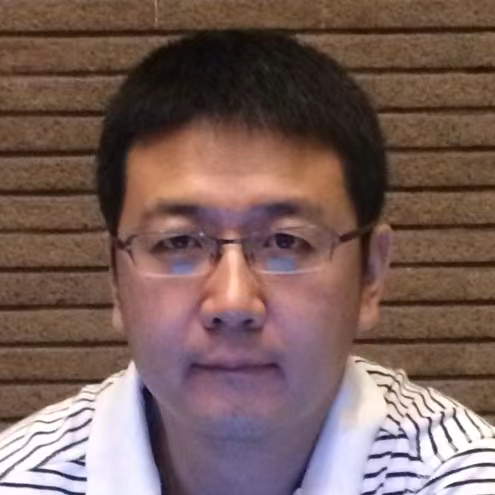}}]{Zhenhua Guo} received the Ph.D. degree in computer science from The Hong Kong Polytechnic University, Hong Kong, in 2010. He was a Visiting Scholar of electrical and computer engineering (ECE) with Carnegie Mellon University, Pittsburgh, PA, USA, from 2018 to 2019. Since September 2022, he has been working with the Tianyijiaotong Technology Ltd., China. His research interests include computer vision, deep learning and object detection.
\end{IEEEbiography}
\vspace{-1.3cm}
\begin{IEEEbiography}[{{\includegraphics[width=1in,height=1.25in,clip,keepaspectratio]{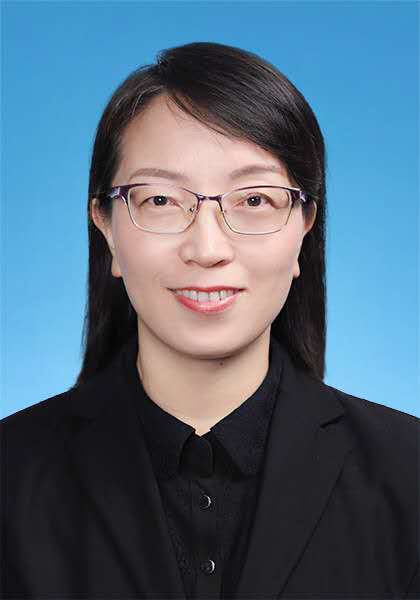}}}]{Xiu Li} received the Ph.D. degree in computer integrated manufacturing from Nanjing University of Aeronautics and Astronautics in 2000. She was a Postdoctoral Fellow with the Department of Automation, Tsinghua University, Beijing, China. 
From 2003 to 2010, she was an Associate Professor with the Department of Automation, Tsinghua University, Beijing, China. Since 2016, She has been a Full Professor at Shenzhen International Graduate School, Tsinghua University. 
% She has authored more than 100 papers in peer-reviewed journals and conferences. 
Her research interests include computer vision and pattern recognition. 
\end{IEEEbiography}

\end{document}